\documentclass{article}

\usepackage[nonatbib,preprint]{neurips_2020}
\usepackage{times}
\usepackage{graphicx} 
\usepackage[numbers]{natbib}
\usepackage{algorithm}
\usepackage{algorithmic}
\usepackage{hyperref}

\usepackage{graphicx}
\usepackage[space]{grffile}

\usepackage{url}            
\usepackage{nicefrac}       
\usepackage{amssymb}
\usepackage{amsmath}
\usepackage{amsthm}
\usepackage{paralist}
\usepackage{subcaption}
\usepackage{setspace}
\usepackage{booktabs}
\usepackage{microtype}
\usepackage{wasysym}
\usepackage{balance}
\usepackage{enumitem}
\usepackage{bbm}

\usepackage{amssymb}
\usepackage{amsfonts}
\usepackage{mathrsfs}
\usepackage{xspace}
\usepackage{bm}
\usepackage{upgreek}

\newcommand{\safemath}[2]{\newcommand{#1}{\ensuremath{#2}\xspace}}

\safemath{\bma}{\mathbf{a}}
\safemath{\bmb}{\mathbf{b}}
\safemath{\bmc}{\mathbf{c}}
\safemath{\bmd}{\mathbf{d}}
\safemath{\bme}{\mathbf{e}}
\safemath{\bmf}{\mathbf{f}}
\safemath{\bmg}{\mathbf{g}}
\safemath{\bmh}{\mathbf{h}}
\safemath{\bmi}{\mathbf{i}}
\safemath{\bmj}{\mathbf{j}}
\safemath{\bmk}{\mathbf{k}}
\safemath{\bml}{\mathbf{l}}
\safemath{\bmm}{\mathbf{m}}
\safemath{\bmn}{\mathbf{n}}
\safemath{\bmo}{\mathbf{o}}
\safemath{\bmp}{\mathbf{p}}
\safemath{\bmq}{\mathbf{q}}
\safemath{\bmr}{\mathbf{r}}
\safemath{\bms}{\mathbf{s}}
\safemath{\bmt}{\mathbf{t}}
\safemath{\bmu}{\mathbf{u}}
\safemath{\bmv}{\mathbf{v}}
\safemath{\bmw}{\mathbf{w}}
\safemath{\bmx}{\mathbf{x}}
\safemath{\bmy}{\mathbf{y}}
\safemath{\bmz}{\mathbf{z}}
\safemath{\bmzero}{\mathbf{0}}
\safemath{\bmone}{\mathbf{1}}

\bmdefine{\biad}{a}
\bmdefine{\bibd}{b}
\bmdefine{\bicd}{c}
\bmdefine{\bidd}{d}
\bmdefine{\bied}{e}
\bmdefine{\bifd}{f}
\bmdefine{\bigd}{g}
\bmdefine{\bihd}{h}
\bmdefine{\biid}{i}
\bmdefine{\bijd}{j}
\bmdefine{\bikd}{k}
\bmdefine{\bild}{l}
\bmdefine{\bimd}{m}
\bmdefine{\bind}{n}
\bmdefine{\biod}{o}
\bmdefine{\bipd}{p}
\bmdefine{\biqd}{q}
\bmdefine{\bird}{r}
\bmdefine{\bisd}{s}
\bmdefine{\bitd}{t}
\bmdefine{\biud}{u}
\bmdefine{\bivd}{v}
\bmdefine{\biwd}{w}
\bmdefine{\bixd}{x}
\bmdefine{\biyd}{y}
\bmdefine{\bizd}{z}

\bmdefine{\bixid}{\xi}
\bmdefine{\bilambdad}{\lambda}
\bmdefine{\bimud}{\mu}
\bmdefine{\bithetad}{\theta}
\bmdefine{\biphid}{\phi}
\bmdefine{\bideltad}{\delta}

\safemath{\bmia}{\biad}
\safemath{\bmib}{\bibd}
\safemath{\bmic}{\bicd}
\safemath{\bmid}{\bidd}
\safemath{\bmie}{\bied}
\safemath{\bmif}{\bifd}
\safemath{\bmig}{\bigd}
\safemath{\bmih}{\bihd}
\safemath{\bmii}{\biid}
\safemath{\bmij}{\bijd}
\safemath{\bmik}{\bikd}
\safemath{\bmil}{\bild}
\safemath{\bmim}{\bimd}
\safemath{\bmin}{\bind}
\safemath{\bmio}{\biod}
\safemath{\bmip}{\bipd}
\safemath{\bmiq}{\biqd}
\safemath{\bmir}{\bird}
\safemath{\bmis}{\bisd}
\safemath{\bmit}{\bitd}
\safemath{\bmiu}{\biud}
\safemath{\bmiv}{\bivd}
\safemath{\bmiw}{\biwd}
\safemath{\bmix}{\bixd}
\safemath{\bmiy}{\biyd}
\safemath{\bmiz}{\bizd}

\safemath{\bmxi}{\bixid}
\safemath{\bmlambda}{\bilambdad}
\safemath{\bmmu}{\bimud}
\safemath{\bmtheta}{\bithetad}
\safemath{\bmphi}{\biphid}
\safemath{\bmdelta}{\bideltad}

\safemath{\bA}{\mathbf{A}}
\safemath{\bB}{\mathbf{B}}
\safemath{\bC}{\mathbf{C}}
\safemath{\bD}{\mathbf{D}}
\safemath{\bE}{\mathbf{E}}
\safemath{\bF}{\mathbf{F}}
\safemath{\bG}{\mathbf{G}}
\safemath{\bH}{\mathbf{H}}
\safemath{\bI}{\mathbf{I}}
\safemath{\bJ}{\mathbf{J}}
\safemath{\bK}{\mathbf{K}}
\safemath{\bL}{\mathbf{L}}
\safemath{\bM}{\mathbf{M}}
\safemath{\bN}{\mathbf{N}}
\safemath{\bO}{\mathbf{O}}
\safemath{\bP}{\mathbf{P}}
\safemath{\bQ}{\mathbf{Q}}
\safemath{\bR}{\mathbf{R}}
\safemath{\bS}{\mathbf{S}}
\safemath{\bT}{\mathbf{T}}
\safemath{\bU}{\mathbf{U}}
\safemath{\bV}{\mathbf{V}}
\safemath{\bW}{\mathbf{W}}
\safemath{\bX}{\mathbf{X}}
\safemath{\bY}{\mathbf{Y}}
\safemath{\bZ}{\mathbf{Z}}

\safemath{\bZero}{\mathbf{0}}
\safemath{\bOne}{\mathbf{1}}
\safemath{\bDelta}{\mathbf{\Delta}}
\safemath{\bLambda}{\boldsymbol\Lambda}
\safemath{\bPhi}{\mathbf{\Upphi}}
\safemath{\bSigma}{\mathbf{\Upsigma}}
\safemath{\bOmega}{\mathbf{\Upomega}}
\safemath{\bTheta}{\mathbf{\Uptheta}}

\bmdefine{\biAd}{A}
\bmdefine{\biBd}{B}
\bmdefine{\biCd}{C}
\bmdefine{\biDd}{D}
\bmdefine{\biEd}{E}
\bmdefine{\biFd}{F}
\bmdefine{\biGd}{G}
\bmdefine{\biHd}{H}
\bmdefine{\biId}{I}
\bmdefine{\biJd}{J}
\bmdefine{\biKd}{K}
\bmdefine{\biLd}{L}
\bmdefine{\biMd}{M}
\bmdefine{\biOd}{N}
\bmdefine{\biPd}{O}
\bmdefine{\biQd}{P}
\bmdefine{\biRd}{R}
\bmdefine{\biSd}{S}
\bmdefine{\biTd}{T}
\bmdefine{\biUd}{U}
\bmdefine{\biVd}{V}
\bmdefine{\biWd}{W}
\bmdefine{\biXd}{X}
\bmdefine{\biYd}{Y}
\bmdefine{\biZd}{Z}

\bmdefine{\biDelta}{\Delta}
\bmdefine{\biLambda}{\Lambda}
\bmdefine{\biPhi}{\Phi}
\bmdefine{\biSigma}{\Sigma}
\bmdefine{\biOmega}{\Omega}
\bmdefine{\biTheta}{\Theta}

\safemath{\bimA}{\biAd}
\safemath{\bimB}{\biBd}
\safemath{\bimC}{\biCd}
\safemath{\bimD}{\biDd}
\safemath{\bimE}{\biEd}
\safemath{\bimF}{\biFd}
\safemath{\bimG}{\biGd}
\safemath{\bimH}{\biHd}
\safemath{\bimI}{\biId}
\safemath{\bimJ}{\biJd}
\safemath{\bimK}{\biKd}
\safemath{\bimL}{\biLd}
\safemath{\bimM}{\biMd}
\safemath{\bimN}{\biNd}
\safemath{\bimO}{\biOd}
\safemath{\bimP}{\biPd}
\safemath{\bimQ}{\biQd}
\safemath{\bimR}{\biRd}
\safemath{\bimS}{\biSd}
\safemath{\bimT}{\biTd}
\safemath{\bimU}{\biUd}
\safemath{\bimV}{\biVd}
\safemath{\bimW}{\biWd}
\safemath{\bimX}{\biXd}
\safemath{\bimY}{\biYd}
\safemath{\bimZ}{\biZd}

\safemath{\bimDelta}{\biDelta}
\safemath{\bimLambda}{\biLambda}
\safemath{\bimPhi}{\biPhi}
\safemath{\bimSigma}{\biSigma}
\safemath{\bimOmega}{\biOmega}
\safemath{\bimTheta}{\biTheta}

\safemath{\setA}{\mathcal{A}}
\safemath{\setB}{\mathcal{B}}
\safemath{\setC}{\mathcal{C}}
\safemath{\setD}{\mathcal{D}}
\safemath{\setE}{\mathcal{E}}
\safemath{\setF}{\mathcal{F}}
\safemath{\setG}{\mathcal{G}}
\safemath{\setH}{\mathcal{H}}
\safemath{\setI}{\mathcal{I}}
\safemath{\setJ}{\mathcal{J}}
\safemath{\setK}{\mathcal{K}}
\safemath{\setL}{\mathcal{L}}
\safemath{\setM}{\mathcal{M}}
\safemath{\setN}{\mathcal{N}}
\safemath{\setO}{\mathcal{O}}
\safemath{\setP}{\mathcal{P}}
\safemath{\setQ}{\mathcal{Q}}
\safemath{\setR}{\mathcal{R}}
\safemath{\setS}{\mathcal{S}}
\safemath{\setT}{\mathcal{T}}
\safemath{\setU}{\mathcal{U}}
\safemath{\setV}{\mathcal{V}}
\safemath{\setW}{\mathcal{W}}
\safemath{\setX}{\mathcal{X}}
\safemath{\setY}{\mathcal{Y}}
\safemath{\setZ}{\mathcal{Z}}
\safemath{\emptySet}{\varnothing}

\safemath{\colA}{\mathscr{A}}
\safemath{\colB}{\mathscr{B}}
\safemath{\colC}{\mathscr{C}}
\safemath{\colD}{\mathscr{D}}
\safemath{\colE}{\mathscr{E}}
\safemath{\colF}{\mathscr{F}}
\safemath{\colG}{\mathscr{G}}
\safemath{\colH}{\mathscr{H}}
\safemath{\colI}{\mathscr{I}}
\safemath{\colJ}{\mathscr{J}}
\safemath{\colK}{\mathscr{K}}
\safemath{\colL}{\mathscr{L}}
\safemath{\colM}{\mathscr{M}}
\safemath{\colN}{\mathscr{N}}
\safemath{\colO}{\mathscr{O}}
\safemath{\colP}{\mathscr{P}}
\safemath{\colQ}{\mathscr{Q}}
\safemath{\colR}{\mathscr{R}}
\safemath{\colS}{\mathscr{S}}
\safemath{\colT}{\mathscr{T}}
\safemath{\colU}{\mathscr{U}}
\safemath{\colV}{\mathscr{V}}
\safemath{\colW}{\mathscr{W}}
\safemath{\colX}{\mathscr{X}}
\safemath{\colY}{\mathscr{Y}}
\safemath{\colZ}{\mathscr{Z}}

\safemath{\opA}{\mathbb{A}}
\safemath{\opB}{\mathbb{B}}
\safemath{\opC}{\mathbb{C}}
\safemath{\opD}{\mathbb{D}}
\safemath{\opE}{\mathbb{E}}
\safemath{\opF}{\mathbb{F}}
\safemath{\opG}{\mathbb{G}}
\safemath{\opH}{\mathbb{H}}
\safemath{\opI}{\mathbb{I}}
\safemath{\opJ}{\mathbb{J}}
\safemath{\opK}{\mathbb{K}}
\safemath{\opL}{\mathbb{L}}
\safemath{\opM}{\mathbb{M}}
\safemath{\opN}{\mathbb{N}}
\safemath{\opO}{\mathbb{O}}
\safemath{\opP}{\mathbb{P}}
\safemath{\opQ}{\mathbb{Q}}
\safemath{\opR}{\mathbb{R}}
\safemath{\opS}{\mathbb{S}}
\safemath{\opT}{\mathbb{T}}
\safemath{\opU}{\mathbb{U}}
\safemath{\opV}{\mathbb{V}}
\safemath{\opW}{\mathbb{W}}
\safemath{\opX}{\mathbb{X}}
\safemath{\opY}{\mathbb{Y}}
\safemath{\opZ}{\mathbb{Z}}
\safemath{\opZero}{\mathbb{O}}
\safemath{\identityop}{\opI}

\safemath{\veca}{\bma}
\safemath{\vecb}{\bmb}
\safemath{\vecc}{\bmc}
\safemath{\vecd}{\bmd}
\safemath{\vece}{\bme}
\safemath{\vecf}{\bmf}
\safemath{\vecg}{\bmg}
\safemath{\vech}{\bmh}
\safemath{\veci}{\bmi}
\safemath{\vecj}{\bmj}
\safemath{\veck}{\bmk}
\safemath{\vecl}{\bml}
\safemath{\vecm}{\bmm}
\safemath{\vecn}{\bmn}
\safemath{\veco}{\bmo}
\safemath{\vecp}{\bmp}
\safemath{\vecq}{\bmq}
\safemath{\vecr}{\bmr}
\safemath{\vecs}{\bms}
\safemath{\vect}{\bmt}
\safemath{\vecu}{\bmu}
\safemath{\vecv}{\bmv}
\safemath{\vecw}{\bmw}
\safemath{\vecx}{\bmx}
\safemath{\vecy}{\bmy}
\safemath{\vecz}{\bmz}

\safemath{\veczero}{\bmzero}
\safemath{\vecone}{\bmone}
\safemath{\vecxi}{\bmxi}
\safemath{\veclambda}{\bmlambda}
\safemath{\vecmu}{\bmmu}
\safemath{\vectheta}{\bmtheta}
\safemath{\vecphi}{\bmphi}
\safemath{\vecdelta}{\bmdelta}

\safemath{\matA}{\bA}
\safemath{\matB}{\bB}
\safemath{\matC}{\bC}
\safemath{\matD}{\bD}
\safemath{\matE}{\bE}
\safemath{\matF}{\bF}
\safemath{\matG}{\bG}
\safemath{\matH}{\bH}
\safemath{\matI}{\bI}
\safemath{\matJ}{\bJ}
\safemath{\matK}{\bK}
\safemath{\matL}{\bL}
\safemath{\matM}{\bM}
\safemath{\matN}{\bN}
\safemath{\matO}{\bO}
\safemath{\matP}{\bP}
\safemath{\matQ}{\bQ}
\safemath{\matR}{\bR}
\safemath{\matS}{\bS}
\safemath{\matT}{\bT}
\safemath{\matU}{\bU}
\safemath{\matV}{\bV}
\safemath{\matW}{\bW}
\safemath{\matX}{\bX}
\safemath{\matY}{\bY}
\safemath{\matZ}{\bZ}
\safemath{\matzero}{\bmzero}

\safemath{\matDelta}{\bDelta}
\safemath{\matLambda}{\bLambda}
\safemath{\matPhi}{\bPhi}
\safemath{\matSigma}{\bSigma}
\safemath{\matOmega}{\bOmega}
\safemath{\matTheta}{\bTheta}

\safemath{\matidentity}{\matI}
\safemath{\matone}{\matO}

\safemath{\rnda}{A}
\safemath{\rndb}{B}
\safemath{\rndc}{C}
\safemath{\rndd}{D}
\safemath{\rnde}{E}
\safemath{\rndf}{F}
\safemath{\rndg}{G}
\safemath{\rndh}{H}
\safemath{\rndi}{I}
\safemath{\rndj}{J}
\safemath{\rndk}{K}
\safemath{\rndl}{L}
\safemath{\rndm}{M}
\safemath{\rndn}{N}
\safemath{\rndo}{O}
\safemath{\rndp}{P}
\safemath{\rndq}{Q}
\safemath{\rndr}{R}
\safemath{\rnds}{S}
\safemath{\rndt}{T}
\safemath{\rndu}{U}
\safemath{\rndv}{V}
\safemath{\rndw}{W}
\safemath{\rndx}{X}
\safemath{\rndy}{Y}
\safemath{\rndz}{Z}

\safemath{\rveca}{\bimA}
\safemath{\rvecb}{\bimB}
\safemath{\rvecc}{\bimC}
\safemath{\rvecd}{\bimD}
\safemath{\rvece}{\bimE}
\safemath{\rvecf}{\bimF}
\safemath{\rvecg}{\bimG}
\safemath{\rvech}{\bimH}
\safemath{\rveci}{\bimI}
\safemath{\rvecj}{\bimJ}
\safemath{\rveck}{\bimK}
\safemath{\rvecl}{\bimL}
\safemath{\rvecm}{\bimM}
\safemath{\rvecn}{\bimN}
\safemath{\rveco}{\bomO}
\safemath{\rvecp}{\bimP}
\safemath{\rvecq}{\bimQ}
\safemath{\rvecr}{\bimR}
\safemath{\rvecs}{\bimS}
\safemath{\rvect}{\bimT}
\safemath{\rvecu}{\bimU}
\safemath{\rvecv}{\bimV}
\safemath{\rvecw}{\bimW}
\safemath{\rvecx}{\bimX}
\safemath{\rvecy}{\bimY}
\safemath{\rvecz}{\bimZ}

\safemath{\rvecxi}{\bmxi}
\safemath{\rveclambda}{\bmlambda}
\safemath{\rvecmu}{\bmmu}
\safemath{\rvectheta}{\bmtheta}
\safemath{\rvecphi}{\bmphi}

\safemath{\rmatA}{\bimA}
\safemath{\rmatB}{\bimB}
\safemath{\rmatC}{\bimC}
\safemath{\rmatD}{\bimD}
\safemath{\rmatE}{\bimE}
\safemath{\rmatF}{\bimF}
\safemath{\rmatG}{\bimG}
\safemath{\rmatH}{\bimH}
\safemath{\rmatI}{\bimI}
\safemath{\rmatJ}{\bimJ}
\safemath{\rmatK}{\bimK}
\safemath{\rmatL}{\bimL}
\safemath{\rmatM}{\bimM}
\safemath{\rmatN}{\bimN}
\safemath{\rmatO}{\bimO}
\safemath{\rmatP}{\bimP}
\safemath{\rmatQ}{\bimQ}
\safemath{\rmatR}{\bimR}
\safemath{\rmatS}{\bimS}
\safemath{\rmatT}{\bimT}
\safemath{\rmatU}{\bimU}
\safemath{\rmatV}{\bimV}
\safemath{\rmatW}{\bimW}
\safemath{\rmatX}{\bimX}
\safemath{\rmatY}{\bimY}
\safemath{\rmatZ}{\bimZ}

\safemath{\rmatDelta}{\bimDelta}
\safemath{\rmatLambda}{\bimLambda}
\safemath{\rmatPhi}{\bimPhi}
\safemath{\rmatSigma}{\bimSigma}
\safemath{\rmatOmega}{\bimOmega}
\safemath{\rmatTheta}{\bimTheta}

\usepackage{amssymb}
\usepackage{amsfonts}
\usepackage{mathrsfs}
\usepackage{xspace}
\usepackage{bm}
\usepackage{fancyref}
\usepackage{textcomp}

\usepackage{multirow}
\usepackage{stmaryrd}

\newenvironment{textbmatrix}{	\setlength{\arraycolsep}{2.5pt}%
								\big[\begin{matrix}}{\end{matrix}\big]%
								\raisebox{0.08ex}{\vphantom{M}}}

\def\be{\begin{equation}}
\def\ee{\end{equation}}
\def\een{\nonumber \end{equation}}
\def\mat{\begin{bmatrix}}
\def\emat{\end{bmatrix}}
\def\btm{\begin{textbmatrix}}
\def\etm{\end{textbmatrix}}

\def\ba#1\ea{\begin{align}#1\end{align}}
\def\bas#1\eas{\begin{align*}#1\end{align*}}
\def\bs#1\es{\begin{split}#1\end{split}} 
\def\bg#1\eg{\begin{gather}#1\end{gather}}
\def\bml#1\eml{\begin{multline}#1\end{multline}}
\def\bi#1\ei{\begin{itemize}#1\end{itemize}}

\newcommand{\lefto}{\mathopen{}\left}

\DeclareMathOperator{\tr}{tr}				
\DeclareMathOperator{\diag}{diag}			
\DeclareMathOperator*{\argmin}{arg\;min}		
\DeclareMathOperator*{\argmax}{arg\;max}		
\DeclareMathOperator{\kron}{\otimes}			
\DeclareMathOperator{\Exop}{\opE}			

\newcommand{\Ex}[2]{\ensuremath{\Exop_{#1}\lefto[#2\right]}} 	

\safemath{\dirac}{\delta}					
\safemath{\krond}{\dirac}					

\safemath{\upto}{\uparrow}
\safemath{\downto}{\downarrow}
\safemath{\iu}{j}							
\safemath{\ev}{\lambda}						
\safemath{\hilseqspace}{l^{2}}				
\newcommand{\banachfunspace}[1]{\setL^{#1}}	
\safemath{\hilfunspace}{\banachfunspace{2}}	

\safemath{\SNR}{\textsf{SNR}} 				
\safemath{\PAR}{\textsf{PAR}} 				
\safemath{\No}{N_0}							
\safemath{\Es}{E_s}							
\safemath{\Eb}{E_b}							
\safemath{\EbNo}{\frac{\Eb}{\No}}
\safemath{\EsNo}{\frac{\Es}{\No}}

\DeclareMathOperator{\CHop}{\ensuremath{\opH}} 
\safemath{\tvir}{\rndh_{\CHop}}				
\safemath{\tvtf}{\rndl_{\CHop}}				
\safemath{\spf}{\rnds_{\CHop}}				
\safemath{\bff}{H_{\CHop}}					

\safemath{\ircf}{r_{h}}						
\safemath{\tftvcf}{r_{s}}					
\safemath{\tfcf}{r_{l}}						
\safemath{\bfcf}{r_{H}}						

\safemath{\tcorr}{c_h}						
\safemath{\scf}{c_{s}}						
\safemath{\tfcorr}{c_{l}}					
\safemath{\fcorr}{c_{H}}						

\safemath{\mi}{I}							
\safemath{\capacity}{C}						

\safemath{\normal}{\mathcal{N}}			
\safemath{\jpg}{\mathcal{CN}}			
\safemath{\mchain}{\leftrightarrow}		

\safemath{\dB}{\,\mathrm{dB}}
\safemath{\dBm}{\,\mathrm{dBm}}
\safemath{\Hz}{\,\mathrm{Hz}}
\safemath{\kHz}{\,\mathrm{kHz}}
\safemath{\MHz}{\,\mathrm{MHz}}
\safemath{\GHz}{\,\mathrm{GHz}}
\safemath{\s}{\,\mathrm{s}}
\safemath{\ms}{\,\mathrm{ms}}
\safemath{\mus}{\,\mathrm{\text{\textmu}s}}
\safemath{\ns}{\,\mathrm{ns}}
\safemath{\ps}{\,\mathrm{ps}}
\safemath{\meter}{\,\mathrm{m}}
\safemath{\mm}{\,\mathrm{mm}}
\safemath{\cm}{\,\mathrm{cm}}
\safemath{\m}{\,\mathrm{m}}
\safemath{\W}{\,\mathrm{W}}
\safemath{\mW}{\, \mathrm{mW}}
\safemath{\J}{\,\mathrm{J}}
\safemath{\K}{\,\mathrm{K}}
\safemath{\bit}{\,\mathrm{bit}}
\safemath{\nat}{\,\mathrm{nat}}

\safemath{\define}{\triangleq}			

\safemath{\equivalent}{\sim}
\safemath{\distas}{\sim}					
\safemath{\sdiff}{\Delta}				

\safemath{\reals}{\mathbb{R}}
\safemath{\positivereals}{\reals_{+}}
\safemath{\integers}{\mathbb{Z}}
\safemath{\posint}{\integers_{+}}
\safemath{\naturals}{\mathbb{N}}
\safemath{\posnaturals}{\naturals_{+}}
\safemath{\complexset}{\mathbb{C}}
\safemath{\rationals}{\mathbb{Q}}

\newcommand*{\fancyrefapplabelprefix}{app}		
\newcommand*{\fancyrefthmlabelprefix}{thm}		
\newcommand*{\fancyreflemlabelprefix}{lem}		
\newcommand*{\fancyrefcorlabelprefix}{cor}		
\newcommand*{\fancyrefdeflabelprefix}{def}		
\newcommand*{\fancyrefproplabelprefix}{prop}	
\newcommand*{\fancyrefobslabelprefix}{obs}		
\newcommand*{\fancyrefalglabelprefix}{alg}		
\newcommand*{\fancyrefasmlabelprefix}{asm}	    
\newcommand*{\fancyrefasmslabelprefix}{asms}	    
\newcommand*{\fancyreftbllabelprefix}{tbl}	    
\newcommand*{\fancyrefestilabelprefix}{esti}	    

\frefformat{vario}{\fancyrefseclabelprefix}{Section~#1}
\frefformat{vario}{\fancyrefthmlabelprefix}{Theorem~#1}
\frefformat{vario}{\fancyreflemlabelprefix}{Lemma~#1}
\frefformat{vario}{\fancyrefcorlabelprefix}{Corollary~#1}
\frefformat{vario}{\fancyrefdeflabelprefix}{Definition~#1}
\frefformat{vario}{\fancyrefobslabelprefix}{Observation~#1}
\frefformat{vario}{\fancyrefasmlabelprefix}{Assumption~#1}
\frefformat{vario}{\fancyrefasmslabelprefix}{Assumptions~#1}
\frefformat{vario}{\fancyreffiglabelprefix}{Figure~#1}
\frefformat{vario}{\fancyrefapplabelprefix}{Appendix~#1} 
\frefformat{vario}{\fancyrefproplabelprefix}{Proposition~#1}
\frefformat{vario}{\fancyrefalglabelprefix}{Algorithm~#1}
\frefformat{vario}{\fancyrefeqlabelprefix}{(#1)}
\frefformat{vario}{\fancyreftbllabelprefix}{Table~#1}
\frefformat{vario}{\fancyrefestilabelprefix}{Estimator~#1}
\newtheorem{thm}{Theorem}

\newtheorem{lem}{Lemma} 

\newtheorem{rem}{Remark}

\safemath{\dictab}{[\,\dicta\,\,\dictb\,]}

\safemath{\ysig}{\bmy}
\safemath{\ysighat}{\hat{\ysig}}
\safemath{\ysigdim}{M}
\safemath{\xsig}{\bmx}
\safemath{\xsigdim}{N}
\safemath{\nx}{n_x}
\safemath{\zsig}{\bmz}
\safemath{\zsigdim}{\ysigdim}
\safemath{\rsig}{\bmr}
\safemath{\Adict}{\bA}
\safemath{\Adicttilde}{\widetilde{\Adict}}
\safemath{\Adictdim}{\outputdim\times\xsigdim}
\safemath{\avec}{\bma}
\safemath{\avectilde}{\tilde{\avec}}
\safemath{\Bdict}{\bB}
\safemath{\Bdicttilde}{\widetilde{\Bdict}}
\safemath{\Cdict}{\bC}
\safemath{\cvec}{\bmc}
\safemath{\Ddict}{\bD}
\safemath{\Ddictdim}{\ysigdim\times\xsigdim}
\safemath{\dvec}{\bmd}
\safemath{\Ddicttilde}{\widetilde{\bD}}
\safemath{\Bonb}{\bB}
\safemath{\bvec}{\bmb}
\safemath{\Bonbdim}{\ysigdim\times\ysigdim}
\safemath{\noise}{\bmn}
\safemath{\noisedim}{\ysigim}
\safemath{\err}{\bme}
\safemath{\errdim}{\ysigdim}
\safemath{\errset}{\setE}
\safemath{\nerr}{n_e}
\safemath{\delop}{\bP_\errset}
\safemath{\delopc}{\bP_{{\errset}^c}}

\safemath{\cplxi}{\imath}
\safemath{\cplxj}{\jmath}

\safemath{\dict}{\matD}
\safemath{\inputdim}{N}		
\safemath{\outputdim}{M}		
\safemath{\sparsity}{S}	
\safemath{\inputdimA}{{N_a}}	
\safemath{\inputdimB}{{N_b}}	
\safemath{\elemA}{{n_a}}	
\safemath{\elemB}{{n_b}}	
\safemath{\resA}{\matR_a}	
\safemath{\resB}{\matR_b}	
\safemath{\subD}{\matS} 
\safemath{\subA}{\matS_a} 
\safemath{\subB}{\matS_b} 
\safemath{\dicta}{\matA} 	
\safemath{\dictb}{\matB} 	
\safemath{\hollowS}{H}
\safemath{\hollowA}{H_a}
\safemath{\hollowB}{H_b}
\safemath{\cross}{Z}
\safemath{\coh}{\mu_d}			
\safemath{\coha}{\mu_a}			
\safemath{\cohb}{\mu_b}			
\safemath{\mubs}{\nu}	
\safemath{\cohm}{\mu_m} 
\safemath{\dictset}{\setD}	
\safemath{\dictsetp}{\dictset(\coh,\coha,\cohb)}	
\safemath{\dictsetgen}{\dictset_\text{gen}}
\safemath{\dictsetgenp}{\dictsetgen(\coh)}
\safemath{\dictsetonb}{\dictset_\text{onb}}
\safemath{\dictsetonbp}{\dictsetonb(\coh)}

\safemath{\leftside}{U}
\safemath{\rightsideA}{R_a}
\safemath{\rightsideB}{R_b}

\safemath{\indexS}{\setI_S} 

\safemath{\na}{n_a}			
\safemath{\nb}{n_b}			
\safemath{\coeffa}{p_i}	
\safemath{\coeffb}{q_j}	
\safemath{\seta}{\setP}		
\safemath{\setb}{\setQ}     
\safemath{\setw}{\setW}	
\safemath{\setz}{\setZ}	
\safemath{\cola}{\veca}		
\safemath{\colb}{\vecb}		
\safemath{\cold}{\vecd}		
\safemath{\inputvec}{\vecx} 	
\safemath{\error}{\vece}	
\safemath{\noiseout}{\vecz} 	
\safemath{\inputvecel}{x}
\safemath{\inputveca}{\vecx_a}
\safemath{\inputvecb}{\vecx_b}
\safemath{\outputvec}{\vecy}	
\safemath{\lambdamin}{\lambda_{\mathrm{min}}}

\safemath{\elltwo}{\ell_2}
\safemath{\ellone}{\ell_1}
\safemath{\ellzero}{\ell_0}
\safemath{\ellinf}{\ell_\infty}
\safemath{\ellinftilde}{\ell_{\widetilde\infty}}
\safemath{\licard}{Z(\coh,\coha,\cohb)}
\safemath{\xsol}{\hat{x}}
\safemath{\xbord}{x_b}		
\safemath{\xstat}{x_s}		
\safemath{\xstatLone}{\tilde{x}_s}
\safemath{\order}{\mathcal{O}} 
\safemath{\scales}{\Theta} 
\safemath{\ones}{\mathbf{1}} 
\safemath{\zeroes}{\mathbf{0}} 
\safemath{\thlone}{\kappa(\coh,\cohb)} 
\safemath{\constoneA}{\delta} 
\safemath{\constoneB}{\epsilon} 
\safemath{\nlarge}{L}				   
\safemath{\sumlarge}{S_\nlarge}
\safemath{\maxlarger}{P_\nlarge}	   
\safemath{\Pzero}{\textrm{P0}}	
\safemath{\Pone}{\textrm{P1}}
\safemath{\vecfir}{\vecw}			 
\safemath{\vecsec}{\vecz}
\safemath{\elvecfir}{w}              
\safemath{\elvecsec}{z}				 
\safemath{\nlargefir}{n}
\safemath{\normout}{\gamma}
\safemath{\auxfun}{h}
\safemath{\supp}{\textrm{supp}}

\safemath{\indexa}{\ell}
\safemath{\indexb}{r}
\safemath{\indexc}{i}
\safemath{\indexd}{j}

\safemath{\project}{P}

\allowdisplaybreaks

\usepackage{color}

\newcommand*\pct{\scalebox{0.9}{\%}}

\safemath{\Herm}{\textnormal{H}}
\safemath{\Tran}{\textnormal{T}}
\newcommand{\fakeparagraph}[1]{\noindent {\bf #1 }}

\title{Asynchronous Multi Agent Active Search}

\newcommand*{\affaddr}[1]{#1} 
\newcommand*{\affmark}[1][*]{\textsuperscript{#1}}
\newcommand*{\email}[1]{\textit{#1}}

\author{%
	Ramina Ghods\affmark[\normalfont 1], Arundhati Banerjee\affmark[\normalfont 2], Jeff Schneider\affmark[\normalfont 1,]\affmark[\normalfont 2]\\ \\
	\affaddr{School of Computer Science,
		Carnegie Mellon University}\\
		\affaddr{\affmark[1]Robotics Institute}
	\affaddr{\affmark[2]Department of Machine Learning}\\
		\email{\{rghods, arundhat, schneide\}@cs.cmu.edu}\\
}

\begin{document}

%
%
%
%
%
%




\maketitle
\begin{abstract}
Active search refers to the problem of efficiently locating targets in an unknown environment by actively making data-collection decisions, and has many applications including detecting gas leaks, radiation sources or human survivors of disasters using aerial and/or ground robots (agents). 
Existing active search methods are in general only amenable to a single agent, or if they extend to multi agent they require a central control system to coordinate the actions of all agents. However, such control systems are often impractical in robotics applications.
%
In this paper, we propose two distinct active search algorithms called SPATS (Sparse Parallel Asynchronous Thompson Sampling) and LATSI (LAplace Thompson Sampling with Information gain) that allow for multiple agents to independently make data-collection decisions without a central coordinator.
Throughout we consider that targets are sparsely located around the environment in keeping with compressive sensing assumptions and its applicability in real world scenarios.
Additionally, while most common search algorithms assume that agents can sense the entire environment (e.g. compressive sensing) or sense point-wise (e.g. Bayesian Optimization) at all times, we make a realistic assumption that each agent can only sense a contiguous region of space
at a time.
%
We provide simulation results as well as theoretical analysis to demonstrate the efficacy of our proposed algorithms.
\end{abstract}



%
\section{Introduction}
\label{sec:introduction}
%
Active search (also referred to as active sensing, robotic sensing, seek and sample) defines the problem of efficiently locating targets in an unknown environment by interactively collecting data
and finds use in applications such as detecting gas leaks, pollution sources or search and rescue missions \cite{ma2017active,rolf2018successive,flaspohler2019information}. In the field of active learning, most existing active search algorithms are developed for a single agent and are not extendable to multi agent scenarios (see related work in \fref{sec:relatedWork}).
As an example, \cite{braun2015info} uses information greedy approaches to decide on best sensing actions for its agent. If we were to use multiple agents for this info-greedy method, all agents would make the same exact decision at each time step wasting resources of other agents. For other active learning algorithms that are extendable to multi agent scenarios, they usually need a central control system to coordinate the sensing actions of all agents \cite{azimi2012batch,gu2014batch}. In this paper, however, we are focusing on robotics applications where a central coordination of agents proves infeasible
due to unreliable communication and limited onboard computing resources
\cite{robin2016multi}. Such a constraint is very common in multi-robot applications of search and rescue and target detection or localization \cite{yan2013survey,murphy2004human,feddema2002decentralized,jennings1997cooperative}.
To clarify, one could still assume communication channels between agents to create coordination but a central coordinator that expects synchronicity is not feasible as any communication mishap could disrupt the entire process.
%
%
While communication constraints have been highly discussed in robotics literature, there has not been many work on multi-robot data collection (task decomposition in robotics) without a central coordinator \cite{yan2013survey,robin2016multi}. 
%
Another consideration of this paper is a realistic assumption on the sensing actions called region sensing initially introduced by \cite{ma2017active}. Inspired by search robots, we assume that each agent
 senses an average value of a contiguous region (block) of the space at each time step. The size of the sensing block models the distance of the agent from the region. We furthermore model noise in our observations in accordance to this distance. Specifically, we assume sensing a larger contiguous region (modelling farther distance from region) inflicts a larger noise value on the resulting observation. 
Lastly, as an essential part of the real-world applications of active search, we assume targets are sparsely located around the environment. 
%
%
%
%
\subsection{Contributions}
\begin{itemize}[leftmargin=5mm]
\item We propose two novel algorithms, SPATS (Sparse Parallel Asynchronous Thompson Sampling) and LATSI (LAplace Thompson Sampling with Information gain), to actively locate targets in an unknown environment. SPATS is a completely nonparametric algorithm with a probabilistic exploration approach that does not need any prior information about the signal of interest. LATSI leverages the benefits of mutual information with probabilistic exploration in the search space. 

\item SPATS and LATSI have three main features that collectively distinguish them in robotics applications of active search. 1) They are multi agent methods where agents can asynchronously make independent data-collection decisions without a central coordinator.  2) They are developed given a practical region sensing assumption. 3) They consider sparse signal recovery.
\item We demonstrate the efficacy of SPATS and LATSI with an extensive set of simulation results in an asynchronous multi agent setting. We provide theoretical analysis on the benefits of SPATS.
\end{itemize}
While there have been many sparse recovery algorithms proposed in the literature, to the best of our knowledge there is no algorithm proposed that develops sparse estimators for active learning methods along with multi agent structure and region sensing assumptions. In this paper, we show how sparsity in its nature limits the exploration factor in active learning methods and how a practical region sensing assumption exacerbates this situation. We propose SPATS and LATSI to strategically address such region sensing assumptions to successfully recover sparse signals. 

\subsection{Related Work}\label{sec:relatedWork}

A prominent approach to estimating sparse signals is compressive sensing (CS) \cite{candes2006robust,donoho2006compressed}. There has been a large number of work on adaptive CS that enables the ability to make online and adaptive measurements to estimate sparse signals and thus is applicable to active search problems \cite{braun2015info,haupt2009adaptive,haupt2009compressive,davenport2012compressive,malloy2014near}. Unfortunately, such adaptive CS methods are sequential and therefore not extendable to multi agent scenarios. Furthermore, CS algorithms in general assume that every measurement matrix can sense the entire environment with arbitrary coefficients which is not a practical assumption for active search problems with region sensing constraints. 

Another area of work are multi-armed bandits. \cite{abbasi2012online} and \cite{carpentier2012bandit} are two multi-armed bandit algorithms that include a sparsity assumption on their hyperparameter. However, the focus of these algorithms are not on estimating the sparse parameter since they are solving for a different reward function. 
%
There have been other Bayesian Optimization (BO) and active learning methods proposed for active search. \cite{marchant2012bayesian} uses BO to develop a spatial mapping of a region whereas we are interested in locating targeted signals. \cite{carpin2015uavs} uses BO for localization of single wireless devices but only focuses on point sensing actions. \cite{ma2017active,rajan2015bayesian,jedynak2012twenty} aim at locating targets by optimizing some notion of Shannon information. Unfortunately, all of the aforementioned active learning algorithms are developed for single agent applications, and except for \cite{ma2017active}, they mostly lack any realistic assumptions on sensing actions.

Regarding multi-agent algorithms, one well-known approach is to use active learning performed in parallel computation settings. Such algorithms in general require a central control system to manage the actions of all the agents by optimizing a batch of actions at each time step and therefore are not applicable to our problem setting \cite{azimi2012batch,gu2014batch,azimi2010batch}. Another multi-agent area of work are mobile sensor networks (MSN) \cite{nguyen2019distributed,scopus2-s2.0-85067810688,la2014cooperative} where multiple mobile sensors/agents reconstruct a scalar map of sensory values in an entire area. MSNs typically consider some form of region sensing assumption on their actions, however, one substantial point of difference of these MSN algorithms 
is the presence of a very constricting sensor network where sensors only follow certain flocking patterns and can only communicate with their close neighbors. Such constrictions are unnecessary in our applications. 

In robotics research, methods that deal with active search generally aim at autonomously building topological (identify obstacles and clearways) and/or spatial maps of a region. Our active search problem differs from topological mapping techniques such as SLAM \cite{leonard1991simultaneous, huang2019survey} and can be most closely related to spatial mapping. For example, \cite{rolf2018successive} identifies strong signals in environments with background information using trajectory planning with confidence intervals; but, unlike our problem setting, their algorithm is developed for a single agent performing point sensing observations. \cite{li2017potential} formulates a multi agent game theoretic approach to coordinate unmanned aerial vehicles for cooperative search and surveillance. While they require the actions of neighboring agents for optimal action selection, we find such requirements to be non-essential in our probabilistic approach.

\fakeparagraph{Notation}
Lowercase and uppercase boldface letters represent column vectors and matrices, respectively. For a matrix $\bA$, its transpose is
$\bA^\Tran$.
%
The $\ell_1$ and $\ell_2$-norm of~$\bma$ are denoted by $\|\bma\|_1$ and $\|\bma\|_2$.
The $N\times N$ identity matrix is denoted by $\bI_{N}$.
%
The Kronecker product is~$\kron$, and
$\diag(\bma)$ is a square matrix with $\bma$ on the main diagonal.
%
%
For a set $\setS$, $|\setS|$ denotes number of elements in that set.



\section{Problem Formulation: Multi Agent Active Search with Region Sensing}
\label{sec:formulation}

\begin{figure}
	\centering
	\begin{subfigure}{0.43\linewidth}
		\includegraphics[width=\linewidth]{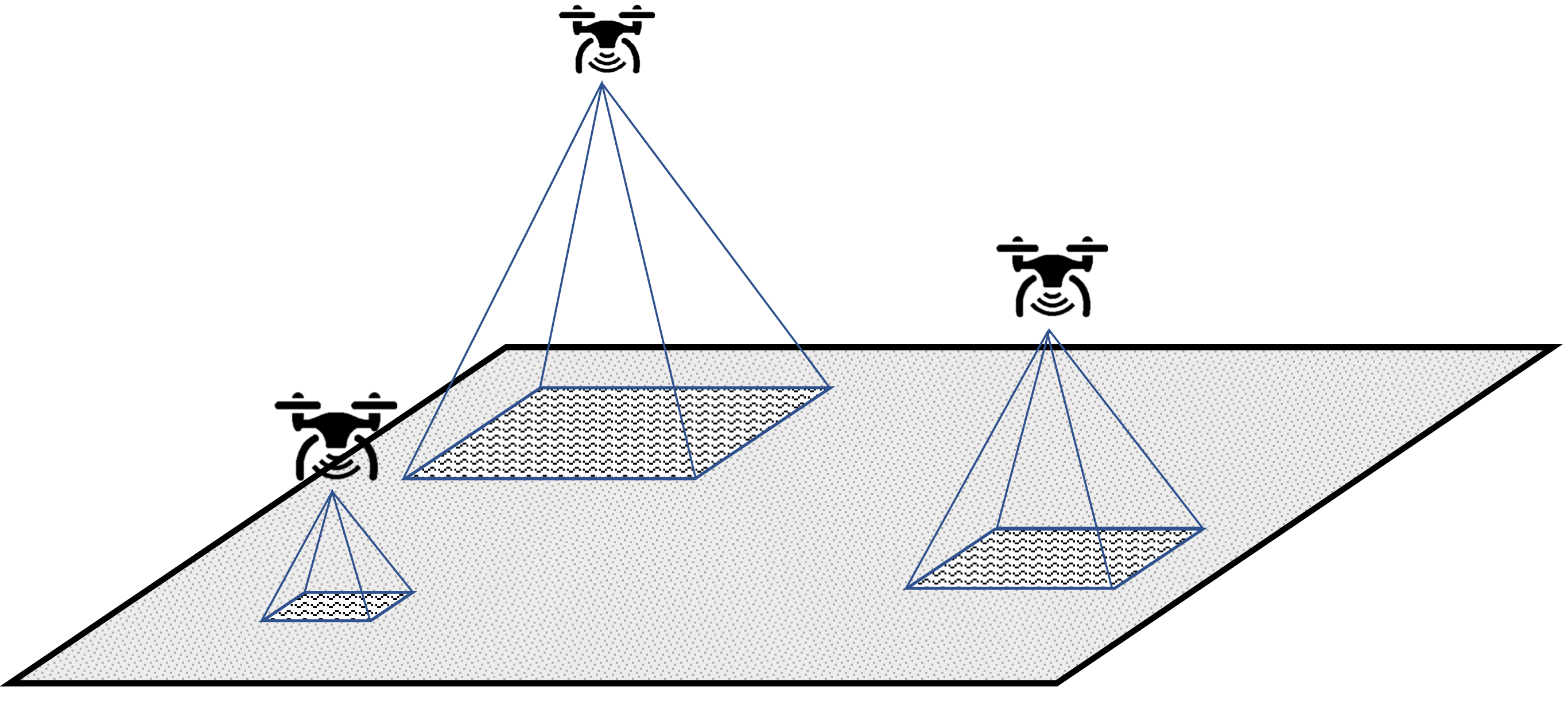}
		\caption{Two-dimensional multi agent active search}
		\label{fig:drones}
	\end{subfigure}
	\hspace*{15mm}
	\begin{subfigure}{0.43\linewidth}
		\includegraphics[width=\linewidth]{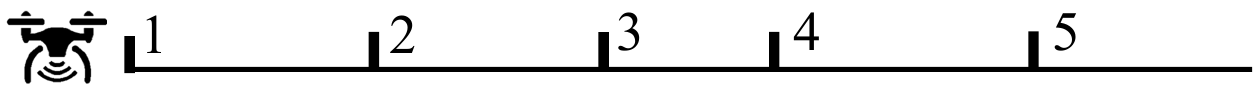}\caption{Single agent}\label{fig:single}
		\includegraphics[width=\linewidth]{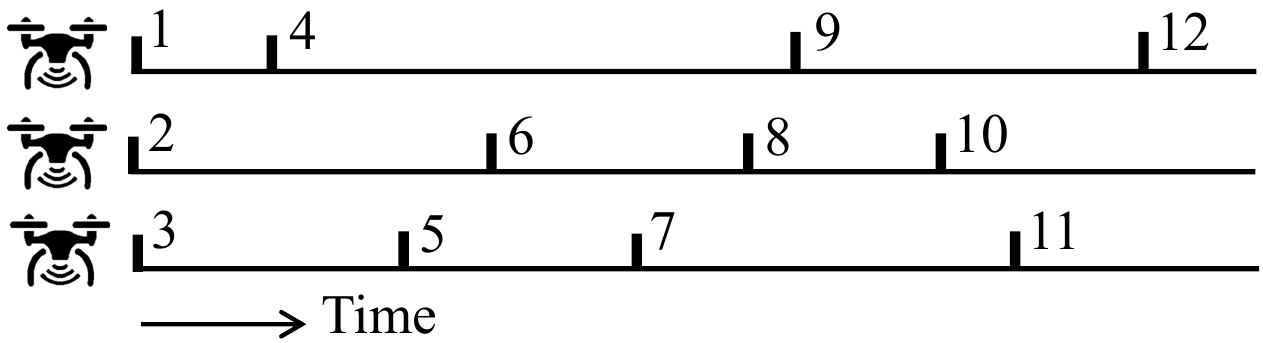}
		\caption{Multi agent (asynchronous)}
		\label{fig:multi}
	\end{subfigure}
	\caption{(a) An illustration of active search for $d=2$. (b), (c) Single and multi agent settings. Here, the small vertical lines indicate the start of $t$'th task. In single agent, tasks start sequentially. In multi agent, task $t$ can start before all previous $t-1$ tasks are finished. 
	}
	\label{SinglevsMulti}
\end{figure}
\fref{fig:drones} illustrates a multi agent active search problem for a two-dimensional environment. Our goal is to efficiently search for targets in an unknown environment by actively taking sensing actions given all the observations thus far. 
This can be thought of as an \emph{active learning} problem (referred to as ``Design of Experiment'' in statistical literature)\cite{activelearning}.  
In particular, we are interested in recovering the sparse d-dimensional matrix $\bB \in \reals^{n_1 \times  \dots \times n_d}$ with minimum measurements. We have no knowledge on the true prior distribution of matrix $\bB$ other than knowing it is sparse. Defining $\bm\beta \in \reals^n$ as a flattened (vectorized) version of matrix $\bB$ with $n\!=\! n_1 \!\times\! ... \!\times\! n_d$, we can write each sensing operation at time step $t$ as:
\begin{align}\label{eq:formulation}
y_t = \bmx_t^\Tran \bm\beta + \epsilon_t, \quad \epsilon_t \sim \setN(0,\sigma^2), t=1,...,T.
\end{align}
Here, $y_t$ is the observation and vector $\bmx_t \in \reals^n$ is the sensing action at time step $t$. We call the set of $(\bmx_t,y_t)$ the measurement at time step $t$. Our objective is to estimate the k-sparse vector $\bm\beta$ ($k\!\! \ll \!\!n$) with as few number of measurements $T$ as possible. Here, we are interested in rectangular sensing actions referred to as \emph{region sensing} \cite{ma2017active}. Precisely, in the original d-dimensional space, our sensing action will be a d-dimensional contiguous rectangle (region) with weights $w_t$ inside the rectangle and zeros outside. As an example, if $d=1$, the sensing action becomes $\bmx_t = [0,...,0,w_t,...,w_t,0,...,0]^\Tran$. 
This constraint models a robot sensing a region of the search space as illustrated in \fref{fig:drones}.
Furthermore, we dedicate a fixed amount of power to each sensing action by letting $\|\bmx_t\|_2=1$.
 This way sensing a larger region at a farther distance from the region would model larger distortion.
\paragraph{Multi Agent Setting}
To actively locate targets, at each time step $t \!\le\! T$, we choose a sensing action $\bmx_t$ given all the available measurements thus far in the set $\bD_{t-1}$. 
For a single agent this procedure is sequential as in \fref{fig:single}  where at time step $t$ the agent uses all previous sequential measurements $\bD_{t-1} \!=\! \{(\bmx_j,y_j)| \,j \!=\! \{1,..,t\!-\!1\}\!\}$ to make a decision.
In this paper, however, we are interested in an asynchronous parallel approach with multiple agents independently making data-collection decisions as in \fref{fig:multi}. Here, asynchronicity means that agents don't wait on results from other agents; instead, an agent starts a new query immediately after its previous data acquisition is completed using all the measurements available thus far; e.g. in \fref{fig:multi}, second agent queries $t=6$'th action before tasks $4$ and $5$ are completed using available measurements $\bD_{t-1} \!=\! \{(\bmx_j,y_j)| \,j \!=\! \{1,2,3\}\!\}$.

For easier computations, we can write a compact model of all the available measurements in $\bD_{t-1}$.
For example for sequential $\bD_{t-1}$, by defining $\bmy\!=\![y_1,...,y_{t-1}]^\Tran, \bX\!=\![\bmx_1^T\!,...,\bmx_{t-1}^\Tran]^\Tran\!$ we can write the model in \fref{eq:formulation} as:
%
\begin{align}\label{eq:formulation2}
\bmy = \bX \bm\beta + \bm\epsilon, \quad  \bm\epsilon \sim \setN(0,\sigma^2 \bI_{t-1}).
\end{align}
\section{Asynchronous Multi Agent Thompson Sampling}
\label{sec:TS}

In order to develop an asynchronous multi agent algorithm without central coordination, we borrow ideas from \cite{kandasamy2018parallelised} who develop asynchronous parallel Thompson Sampling (TS) for Bayesian Optimization (BO). We propose extending this idea to active search problems.
TS is an exploration-exploitation algorithm originally introduced for clinical trials by \cite{thompson1933likelihood} and later rediscovered for multi-armed bandits \citep{wyatt1998exploration,strens2000bayesian,russo2018tutorial}. 
\cite{kandasamy2018parallelised} shows that TS for BO can be developed for an asynchronous parallel setting where each agent makes independent and intelligent decision on the next action given the available measurements. This feature makes TS an ideal solution to our multi agent active search problem.
As proven in this reference, asynchronous parallel TS outperforms existing parallel BO methods. 
%

%


We now review TS for active learning. 
We start with the single agent setting as introduced in \cite{kandasamy2019myopic}. 
We are interested in recovering the $n$-dimensional vector $\bm\beta \sim p_0$. We actively query actions $\bmx_t$ and observe their outcome $y_t$ where the likelihood $p(y_t | \bmx_t,\bm\beta)$ is known. 
%
To query the best action, we maximize a reward function $\lambda(\bm\beta^\star, \bD_{t-1})$. As an example, the reward function can be $\textstyle \lambda(\bm\beta^\star, \bD_{t-1}) = -\|\bm\beta^\star - \hat{\bm\beta}(\bD_{t-1})\|_2^2$, where $\bm\beta^\star$ is our belief of the true $\bm\beta$, and $\hat{\bm\beta}(\bD_{t-1})$ is the estimated value of $\bm\beta$ given all the available measurements $\bD_{t-1}$ (e.g. maximum likelihood estimate). 
We are interested in the myopic policy which selects action $\bmx$ that maximizes the expected reward of time step $t$, i.e. 
\vspace*{-3mm}
\begin{align}\label{eq:expectedReward}
\lambda^+(\bm\beta^\star,\bD_{t-1},\bmx) = \Ex{y|\bmx,\bm\beta^\star}{\lambda(\bm\beta^\star, \bD_{t-1} \cup (\bmx,y))}.
\end{align}
Here, the best reward would be the one that has access to the true value of $\bm\beta$, i.e. $\lambda^+(\bm\beta,\bD_{t-1},\bmx)$.
Not knowing the true value of $\bm\beta$, TS will sample it from the current posterior distribution of $\bm\beta$ conditioned on the measurements $\bD_{t-1}$, i.e. $\bm\beta^\star \! \sim p(\bm\beta | \bD_{t-1})$. Hence, according to TS, the best action $\bmx_t$ is one that maximizes the reward assuming that the sample $\bm\beta^\star$ is the true value of $\bm\beta$. 
\begin{algorithm}[b]
	\caption{Asynchronous Multi Agent Thompson Sampling}
	\begin{algorithmic}
		\STATE {\bfseries Assume:} prior $\bm\beta \sim p_0$ and likelihood $p(y_t| \bmx_t,\bm\beta)$
		\STATE{\bfseries For}{ $t=1,...,T$}
		\STATE \hspace{0.4cm}Wait for an agent to finish; {for the freed agent:}
		\STATE \hspace*{0.8cm}Sample $\bm\beta^\star \sim p(\bm\beta | \bD_{t-1})$ \hfill \textit{Posterior Sampling}
		\STATE \hspace*{0.8cm}Select $\bmx_t = \argmax_{\bmx} \lambda^+(\bm\beta^\star,\bD_{t-1},\bmx)$ \hfill \textit{Design}
		\STATE \hspace*{0.8cm}Observe $y_t$ given action $\bmx_t$; update \& share measurements $\bD_{t} = \bD_{t-1} \cup (\bmx_t,y_t)$
	\end{algorithmic}
	\label{alg:TS}
\end{algorithm}
The case of multi agent directly follows from \cite{kandasamy2018parallelised}. Consider $g$ agents planning on taking $T$ measurements of an environment. 
Say an agent finishes making an observation and is ready to choose the $t$'th action. Using measurements available from all agents so far ($|\bD_{t-1}|\! \le \!t\!-\!1$), it will update and sample the posterior (\textit{posterior sampling}), select its next sensing action that maximizes the reward (\textit{design}), evaluate its action and share the observations with other agents.
\fref{alg:TS} summarizes this process.
%
%
%
%
%
%
%

%
%
%
%
%
\subsection{Thompson Sampling with Sparse Prior}\label{sec:LTS}
In order to develop TS in \fref{alg:TS} for our active search problem in \fref{sec:formulation}, we start by first establishing the prior $p(\bm\beta)$ and likelihood distribution $p(y_t | \bmx_t,\bm\beta)$. As for the prior, our knowledge is limited to the presence of sparsity. Hence, we will assume $\bm\beta$ has a Laplace distribution with independent entries and a tunable parameter $b$, i.e. $\textstyle p(\bm\beta) = \frac{1}{(2b)^n}\exp(-\frac{\|\bm\beta\|_1}{b})$. Laplace distribution translates to an $\ell_1$-norm regularization term in the cost function which has been shown to introduce sparsity into the estimator \cite{williams1995bayesian,tibshirani1996regression,chen2001atomic}. For the likelihood distribution, the sensing model in 
\fref{eq:formulation2} gives $p(\bmy|\bX,\bm\beta) = \setN(\bX \bm\beta,\sigma^2 \bI_{t-1})$. 
Next step is to derive the posterior sampling and design stages of \fref{alg:TS} using this prior and likelihood. \fref{app:app-LTS} provides a detailed derivation of these two stages. We call the resulting algorithm Laplace-TS.

\paragraph{Facing the Failure Mode of TS with Single Agent}
%
%
Unfortunately, Laplace-TS with single agent leads to poor performance that is on par with a point-wise algorithm that exhaustively searches all locations one at a time.
%
%
We can associate this poor performance with one of the failure modes of TS discussed in Sec. 8.2 of the tutorial by \cite{russo2018tutorial}. According to this tutorial, TS faces a dilemma when solving certain kinds of active learning problems. One such scenario are problems that require a careful assessment of information gain. In general, by optimizing the expected reward, TS always restricts its actions to those that have a chance in being optimal which in our case are sparse sensing actions restricted further by the region sensing constraint.
However, in active learning problems such as ours, suboptimal actions (i.e. nonsparse sensing actions) can carry additional information regarding the parameter of interest. \fref{app:LTSfail} includes simulation results as well as an example to further illustrate the failure mode in active search problems.
In the next section, we will modify Laplace-TS and propose two algorithms that can bypass this failure mode.

%
%
%
%
%
%
%


%
%
%
\section{Our Proposed Algorithms: SPATS and LATSI}
%
%
%
%
\subsection{SPATS: Sparse Parallel Asynchronous Thompson Sampling}\label{sec:SPATS}
%
%
Per our discussion in \fref{sec:LTS}, introducing sparsity with Laplace prior into TS algorithm limited its ability to explore queries. With this in mind, one might conclude that choosing non-sparse samples in the posterior sampling stage of \fref{alg:TS} should solve this problem. However, this strategy will still face the failure mode of TS because it is the sparse estimator in the design stage that is limiting the feasible sensing actions. The next logical solution would then be to make both the estimator and posterior sampling procedures non-sparse. Even though with this strategy we will avoid the failure mode of TS, without taking advantage of the prior information about sparsity, the resulting non-sparse TS will be performing no better than exhaustively searching the entire space.
To overcome this issue, we propose making an assumption on the prior distribution of both the sampling and estimation procedures that the neighbouring entries of the sparse vector $\bm \beta$ are temporally correlated, i.e. $\bm \beta$ is block sparse. Such temporal correlation creates the most compatible results to the region sensing constraint which only approves sensing actions with a single non-zero block of sensors. Furthermore, we expect block sparsity to introduce exploration ability while also keeping sparsity a useful information in the recovery process. In particular, by gradually reducing the length of the blocks from a large value, we gently trade exploration with exploitation capability over time. 
%

%
%
In short, borrowing ideas from a block sparse Bayesian framework introduced in \cite{zhang2011sparse},
we use a block sparse prior $p(\bm \beta) = \setN(\bZero_{n \times 1},\bm \Sigma_0)$, where:
%
\resizebox{.3\hsize}{!}{$
\bm \Sigma_0 = \text{diag} \left([ \gamma_1 \bB_1,...,\gamma_M \bB_M]\right)
,$}
%
%
with $\gamma_m$ and $\bB_m \in \reals^{L \times L}$ ($m=1,...,M$) as hyperparameters. Here, $\gamma_m$ controls the sparsity of each block as is the case in sparse Bayesian learning methods \cite{tipping2001sparse,wipf2004sparse}, i.e. when $\gamma_m = 0$, the corresponding block $m$ is zero. Here, $L$ is the length of the blocks that we will gradually reduce in the TS process. To avoid overfitting while estimating these hyperparameters, \cite{zhang2013extension} suggests one matrix $\bB$ to model all block covariances, namely $\bm \Sigma_0 = \text{diag}(\bm \gamma) \kron \bB$, where, $\bm\gamma$ is the vector containing all elements of $\gamma_m$ for $m=1,...,M$. 
\fref{app:blockTS} provides a detailed derivation of \fref{alg:TS} with this prior for the active search problem in \fref{sec:formulation}.
\begin{algorithm}
	\caption{SPATS}
	\begin{algorithmic}
		\STATE {\bfseries Assume:} Sensing model \fref{eq:formulation}; sparse signal $\beta$; $g$ agents; block length $L$
		\STATE {\bfseries Set:} $\bD_{0} = \varnothing$; $L = n/g$; $\gamma_m = 1$;
		$\bB:$ random highly correlated covariance matrix
		\STATE{\bfseries For }{$t=1,...,T$}
		\STATE \hspace{0.4cm}Wait for an agent to finish; {for the freed agent:}
		\STATE \hspace{0.8cm}Sample $\bm\beta^\star \sim p(\bm\beta | \bD_{t-1},\bm \gamma,\bB)$ from \fref{eq:blockposterior} in \fref{app:blockTS}
		\STATE \hspace{0.8cm}Select $\bmx_t \!=\! \argmax_{\bmx} \lambda^+(\bm\beta^\star,\bD_{t-1},\bmx)$ using \fref{eq:blockestimator} in \fref{app:blockTS}
		\STATE \hspace{0.8cm}Observe $y_t$ given action $\bmx_t$; update \& share  measurements $\bD_{t} = \bD_{t-1} \cup (\bmx_t,y_t)$
		\STATE \hspace{0.8cm}Update hyper parameters $\bm \gamma$ and $\bB$ using EM algorithm in \fref{eq:blockEM} in \fref{app:blockTS}
		\STATE \hspace{0.8cm}{\bfseries if} {$t \, \pct \, g = 0$} {\bfseries then } $L = L/2$
	\end{algorithmic} \label{alg:SPATS}
\end{algorithm}
 \fref{alg:SPATS} called SPATS summarizes our results in this section. SPATS has much lower computational cost than Laplace-TS algorithm since it does not require a Gibbs sampler. Furthermore, unlike Laplace-TS, SPATS is completely nonparametric and does not need to know the sparsity rate or any other prior information about the true signal $\bm \beta$. 
%
%
%
%
%
%
%
\subsubsection{Theoretical Bounds for a Sparse Model}
We now provide theoretical analysis testifying to the benefits of SPATS. SPATS has two aspects that distinguish it from a na\"ive TS developed for sparse signals. One is using a block sparse prior with varying block length and two is using multiple agents. In what follows, we introduce two theorems to investigate the benefits of each aspect separately. First in \fref{thm:singleAgent}, for a sparse model with single agent setting we will compute and compare upper bounds on the expected regret of two TS algorithms with a 1-sparse and a 1-block sparse prior with one nonzero block. The 1-block sparse prior closely imitates SPATS's performance with a region sensing assumption. See proof in \fref{app:thmSingleAgent}.

\begin{thm}\label{thm:singleAgent}
	Consider an active search problem with a 1-sparse true parameter $\bm\beta \in \reals^n$ and reward function $\setR(\bmx,\bm\beta) \!=\! (\bmx^\Tran \bm\beta)^2$ for action $\bmx \!\in\! \reals^n$ chosen from set of actions $\setX$ that satisfy region sensing in \fref{sec:formulation}. Consider two single agent TS algorithms where one assumes a 1-sparse prior and another uses a 1-block sparse prior with varying block length as defined in \fref{alg:SPATS}. Then, the expected regret $\Ex{}{\text{Reg(T)}} \!=\! \Ex{}{\!\sum_{t=1}^{T} \setR(\bmx^\star\!,\bm\beta) \!-\! \setR(\bmx_t,\bm\beta)\!}$ for each algorithm is upper-bounded by:
	\begin{align}\label{eq:regret_1}
&\text{1-sparse prior}: \textstyle \Ex{}{\text{Reg}(T)} \leq  \left(\log\!\left(|\setX|\right) \sum_{t=1}^{\min\{T,n-1\}} \frac{\left(1-\frac{t}{n}\right)\left(1-\frac{1}{n-t+1}\right)}{\left(\frac{n-t-1}{n-t} \log\left(\frac{n-t}{n-t-1}\right)+\frac{1}{n-t} \log\left(n-t\right)\right)}\right)^{1/2}\\\label{eq:regret_block}
&\text{1-block sparse prior}: \textstyle \Ex{}{\text{Reg}(T)} \!\leq \!\!\left( \!\!\log\!\left(|\setX|\right) \!\sum_{t=1}^{\min\{T,\log_2(n)\!\}} \!\! {\Big(\!1\!-\frac{1}{n-\left(\sum_{t^\prime=1}^{t-1}\frac{n}{2^{t^\prime}}\right)}\!\Big)^2}\!\!/{\log(2)} \!\!\right)^{1/2}
	\end{align}
\end{thm}
A simple comparison of \fref{eq:regret_1} and \fref{eq:regret_block} in \fref{thm:singleAgent} shows that using TS with a block sparse prior and varying block length significantly reduces the regret bounds comparing to TS that is using the true 1-sparse prior. 
Next, we will compute and compare an upper bound on the expected regret of a single-agent and an asynchronous multi-agent TS algorithm. To the best of our knowledge, only theoretical analysis for asynchronous parallel TS has been provided by \cite{kandasamy2018parallelised} which is limited to Gaussian Processes. In the following theorem, we provide theoretical guarantees for an asynchronous multi-agent active search problem with a sparse model with proof in \fref{app:thmMultiAgent}.
%
\begin{thm}\label{thm:multiAgent}
Consider the active search problem in \fref{thm:singleAgent}. Let us propose two TS algorithms with a 1-sparse prior where one is single agent and another uses $g$ agents in an asynchronous parallel setting. Then, the expected regret as defined in \fref{thm:singleAgent} for each algorithms is:
	\begin{align}\label{eq:regret_single}
&\text{single agent: } \quad \textstyle \Ex{}{\text{Reg}(T)} = T_n - \frac{T_n(T_n+1)}{2n}, \quad T_n = \min\{T,n\} \\\label{eq:regret_multi}
&\text{multi agent: } \quad \textstyle \Ex{}{\text{Reg}(T)} \leq  T_n - \frac{T_n(T_n+1)}{2n} \!+\!\frac{T_n(2g-1)}{n}, \quad T_n \!= \min\{T,n\!+\!g\} 
\end{align}
\end{thm}
A simple analysis of \fref{eq:regret_multi} shows that for $g\ll n$ and $g \ll T$ (which is a reasonable assumption), the third term in the bound will be upper bounded by $2g+1$. As a result, the difference in expected regret between single agent and asynchronous multi agent is negligible in terms of number of measurements $T$. Hence, we can conclude that by dividing the same number of measurements $T$ between $g$ agents, multi agent algorithm achieves same regret $g$ times faster than single agent setting.
\begin{rem}
	\fref{thm:multiAgent} shows that our asynchronous multi agent algorithm performs on par with an optimal multi agent system with a central planner. This result is a consequence of the central planner's regret being bounded by the single agent in terms of number of measurements $T$ \cite{kandasamy2017asynchronous}.
\end{rem}

\subsection{LATSI: Laplace TS with Information gain}
\label{sec:LATSI}

In \fref{sec:LTS}, we discussed how single agent Laplace-TS fails due to a careless assessment of the information gain. To combat this issue, \cite{russo2017learning} proposes a new reward function which is a combination of expected regret and the mutual information between the optimal action and the next observation. They show that their algorithm called Information Directed Sampling (IDS) can considerably improve the performance of single agent Laplace-TS. Unfortunately,
computing the mutual information as introduced in the IDS algorithm is not computationally feasible for our problem. Specifically, IDS proposes sampling techniques to approximate the mutual information between the optimal action and the next observation. However, sampling the optimal action would require computing it for each sample which considering the region sensing assumption is quite expensive. 

Another information-theoretic active search algorithm is Region Sensing Index (RSI) by \cite{ma2017active} that recognizes region sensing constraints. RSI searches for sparse signals by maximizing the mutual information between the next observation and the true parameter $\bm \beta$. RSI does not extend well to multi agent settings as without randomness in its reward function, all agents will solve for the same sensing action. However, inspired by the IDS algorithm, we propose combining the reward of Laplace-TS with the mutual information computed in RSI. We call this algorithm LATSI summarized in \fref{alg:LATSI} with additional details in \fref{app:LATSI}. We will provide simulation results in the next section, showing that while LATSI improves Laplace-TS, for $k>1$ LATSI performs poorly comparing to SPATS due to RSI's poor approximation of mutual information for that case.

\begin{algorithm}[t]
	\caption{LATSI}
	\begin{algorithmic}
		\STATE {\bfseries Assume:} Sensing model \fref{eq:formulation}; sparse signal $\beta$; $g$ agents
		\STATE {\bfseries Set:} $\bD_{0} \!\!=\!\! \varnothing$; $\bm\tau$: randomly initialized; $\alpha, \eta$ for tuning;
		$p_0:$ 1-sparse uniform distribution on $\bm\beta$
		\STATE{\bfseries For }{$t=1,...,T$}
		\STATE \hspace{0.4cm}Wait for an agent to finish; {for the freed agent:}
		\STATE \hspace{0.8cm}Sample $\bm\beta^\star \sim p(\bm\beta | \bD_{t-1},\bm \tau)$ (Gibbs sampler) \fref{eq:betaGibbs}
		\STATE \hspace{0.8cm}Select $\bmx_t \!=\! \argmax_{\bmx} \bm{R}^+(\bm\beta^\star, \bD_{t-1}, \{\bmx,y\})$\\[1mm]
		
		\hspace{1cm}\big($\bm{R}^+ \!= \!\frac{I(\bm\beta^\star; y | \bmx, \pi_{t-1})}{\text{average }I\text{ over all }\bmx} \!+\! \alpha \! \times \! \frac{\lambda^+(\bm\beta^\star,\bD_{t-1},\bmx)}{\text{average }\lambda^+\text{ over all }\bmx}$ using $I$ from \cite{ma2017active} and $\lambda^+$ in \fref{eq:lambda+}\big)\\[1mm]
		\STATE \hspace{0.8cm}Observe $y_t$ given action $\bmx_t$; update \& share measurements  $\bD_{t} = \bD_{t-1} \cup (\bmx_t,y_t)$
		\STATE \hspace{0.8cm}Update $\bm \tau$ using EM algorithm in \fref{eq:EMsolution} and $p(\bm\beta | \bD_{t-1},\bm \tau)$ following RSI-A in \cite{ma2017active}
	\end{algorithmic} \label{alg:LATSI}
\end{algorithm}

\section{Numerical Results}
\label{sec:results}

We now compare the performance of our SPATS and LATSI against the information-theoretic approach called RSI proposed in \cite{ma2017active}. In this section, we focus on 2-dimensional search spaces ($d=2$), where we estimate a k-sparse signal $\bm \beta$ with length $n=8\times16$ and two sparsity rates of $k=1, 5$. Here, $\bm\beta$ is generated with a randomly uniform sparse vector. We set the noise variance to $\sigma^2 = 1$. For LATSI, we set the tuning parameters $\alpha,\eta\!=\!1$. Note that neither SPATS nor LATSI are aware of the true uniform sparse prior or sparsity rate $k$. We then vary the number of measurements $T$ and plot the mean and standard error of the full recovery rate over $50$ random trials. The full recovery rate is defined as the rate at which an algorithm correctly recovers the entire vector $\bm \beta$ over random trials. To further demonstrate the efficacy of SPATS and LATSI, we provide additional experiments for $d=1$, larger length $n$, $k=10$ and a sensitivity analysis for LATSI in \fref{app:experiments}.
%
%
\subsection{Single Agent}
\label{sec:single-agent}
\begin{figure}
	\vspace*{3mm}
	\centering
	\begin{subfigure}{0.495\linewidth}
		\includegraphics[width=0.495\linewidth]{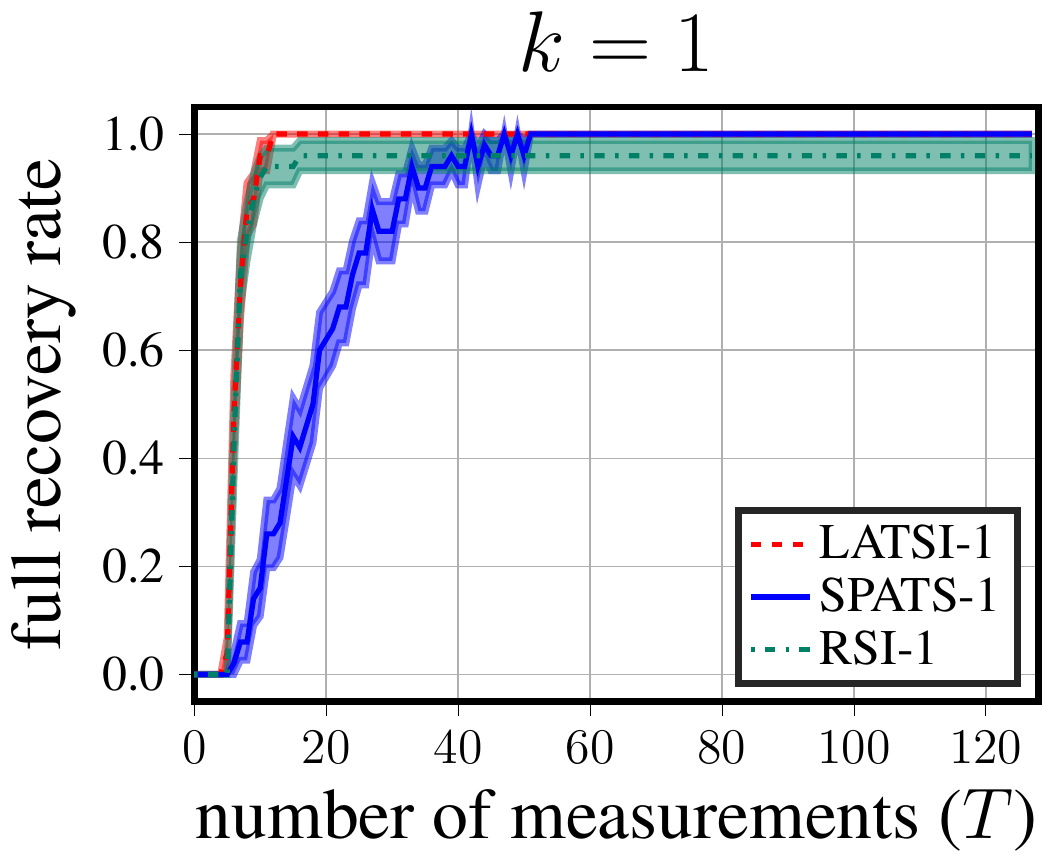}
		\includegraphics[width=0.495\linewidth]{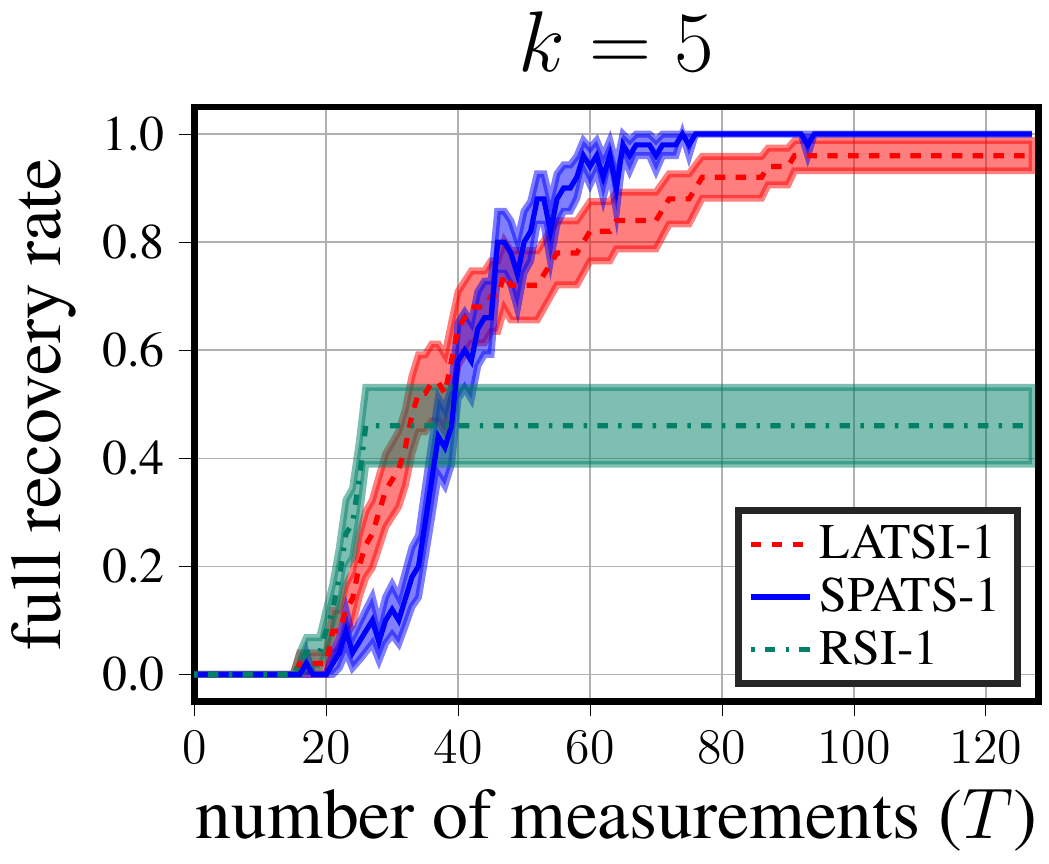}
		\caption{single agent}
		\label{fig:RSI_TS_1ag_plots}
	\end{subfigure}
	\begin{subfigure}{0.495\linewidth}
		\includegraphics[width=0.495\linewidth]{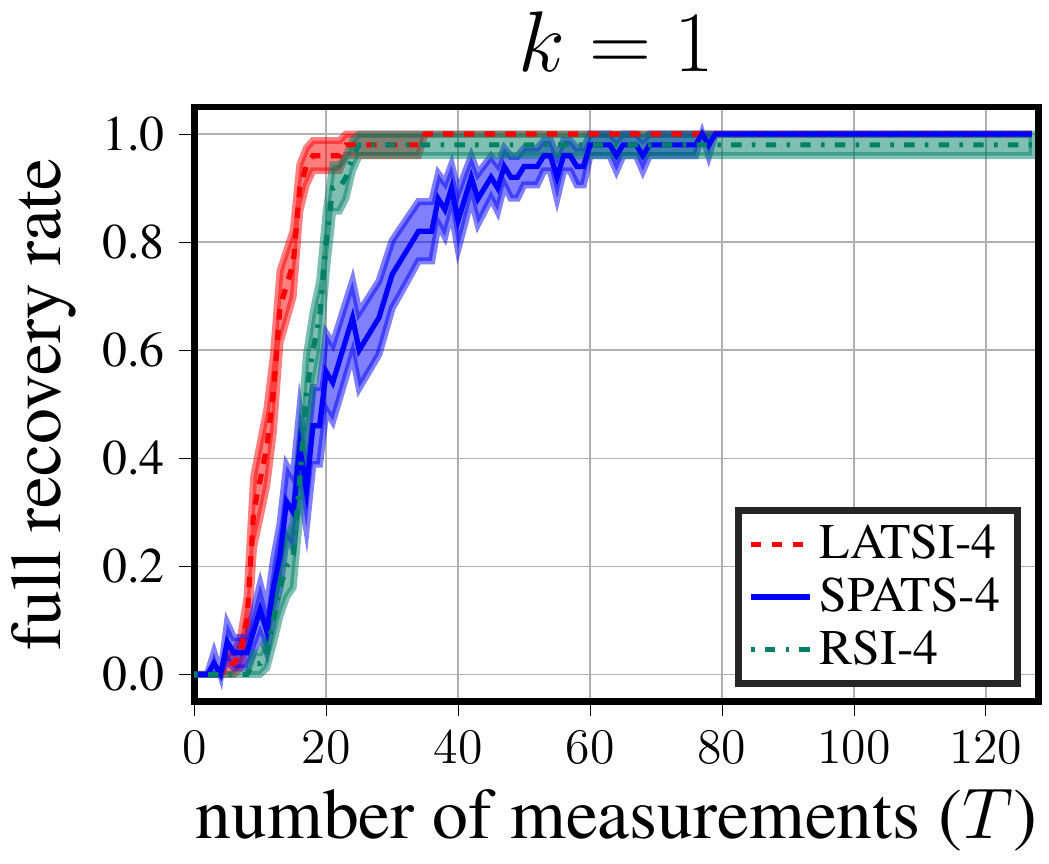}
		\includegraphics[width=0.495\linewidth]{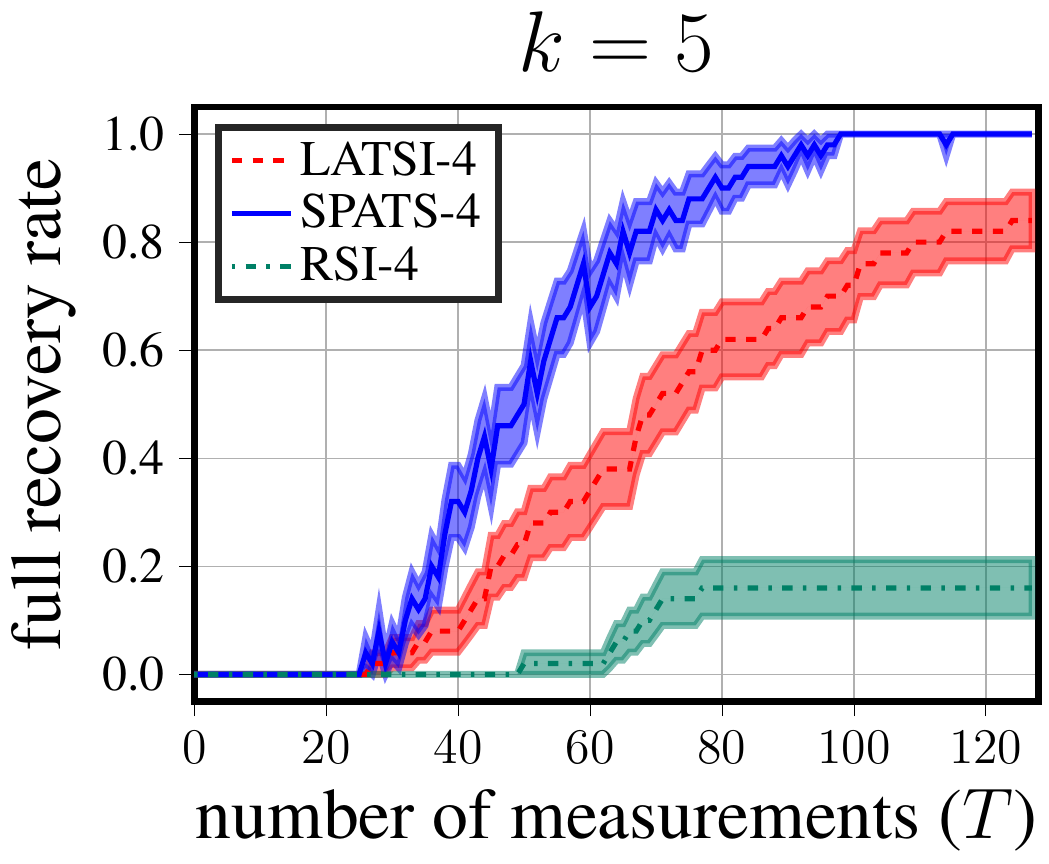}
		\caption{4 agents}
		\label{fig:ALL_plots}
	\end{subfigure}
	\caption{Full recovery rate of SPATS, LATSI and RSI for 1 and 4 agents for sparsity rates $k=1, 5$}
\end{figure}

In a single agent setting, \fref{fig:RSI_TS_1ag_plots} shows that for $k=1$, RSI and LATSI outperform SPATS. The reason is that RSI has a very accurate approximation of mutual information for $k=1$ and consequently it is difficult for our SPATS to win over the information-optimal algorithms of RSI and LATSI. On the other hand, for higher sparsity rate of $k=5$, SPATS outperforms RSI and LATSI. This is a result of poor approximation of mutual information for $k>1$ by RSI. Specifically, for $k>1$ RSI recovers the support of $\bm \beta$ by repeatedly applying RSI assuming $k=1$. The authors use this strategy to avoid the large cost of computing mutual information for $k>1$. Since our proposed LATSI is a combination of RSI and Laplace-TS, its performance is tied to that of both RSI and SPATS. 
\begin{figure}[H]
	\centering
	\begin{subfigure}{0.245\linewidth}
		\includegraphics[width=\linewidth]{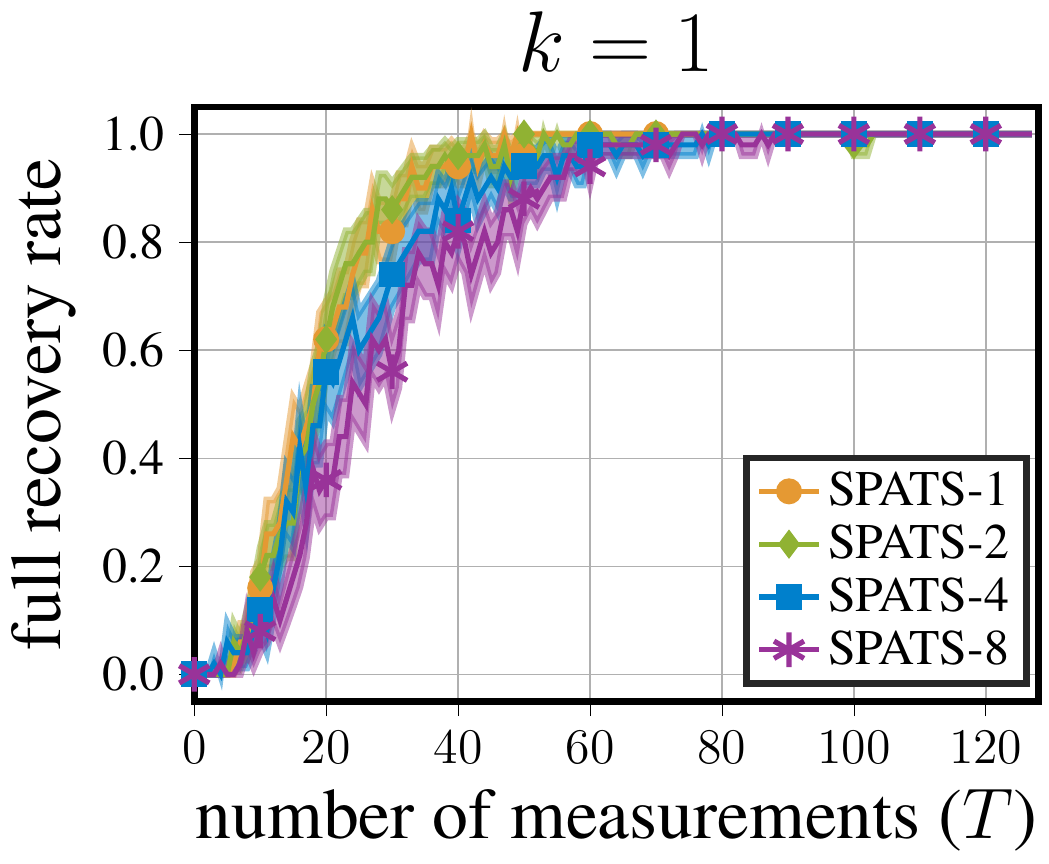}
	\end{subfigure}
	\begin{subfigure}{0.245\linewidth}
		\includegraphics[width=\linewidth]{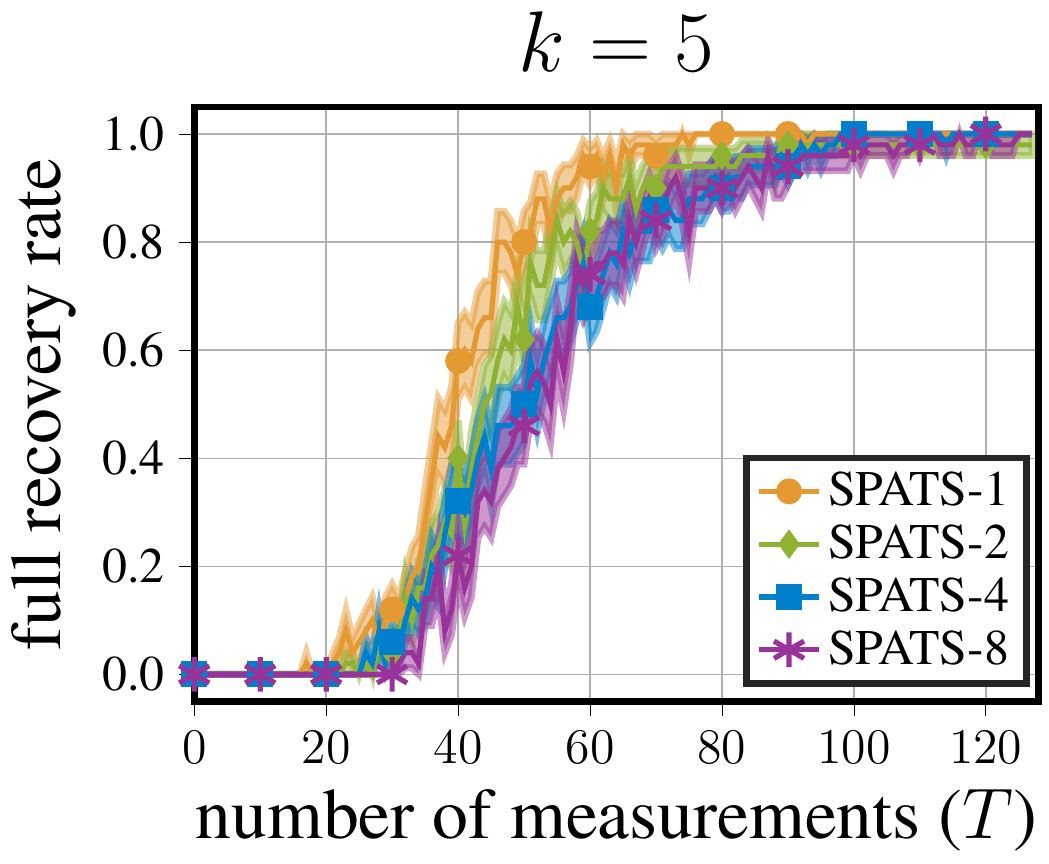}
	\end{subfigure}
	\begin{subfigure}{0.245\linewidth}
		\includegraphics[width=\linewidth]{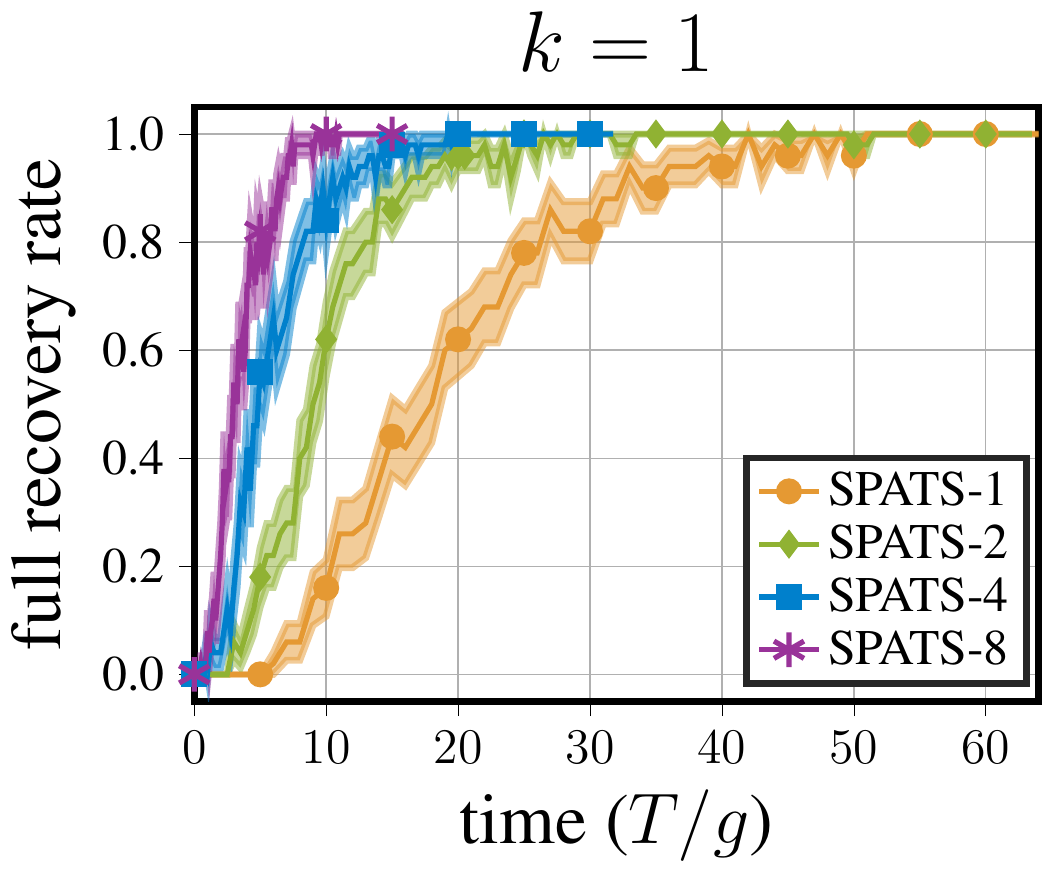}
	\end{subfigure}
	\begin{subfigure}{0.245\linewidth}
		\includegraphics[width=\linewidth]{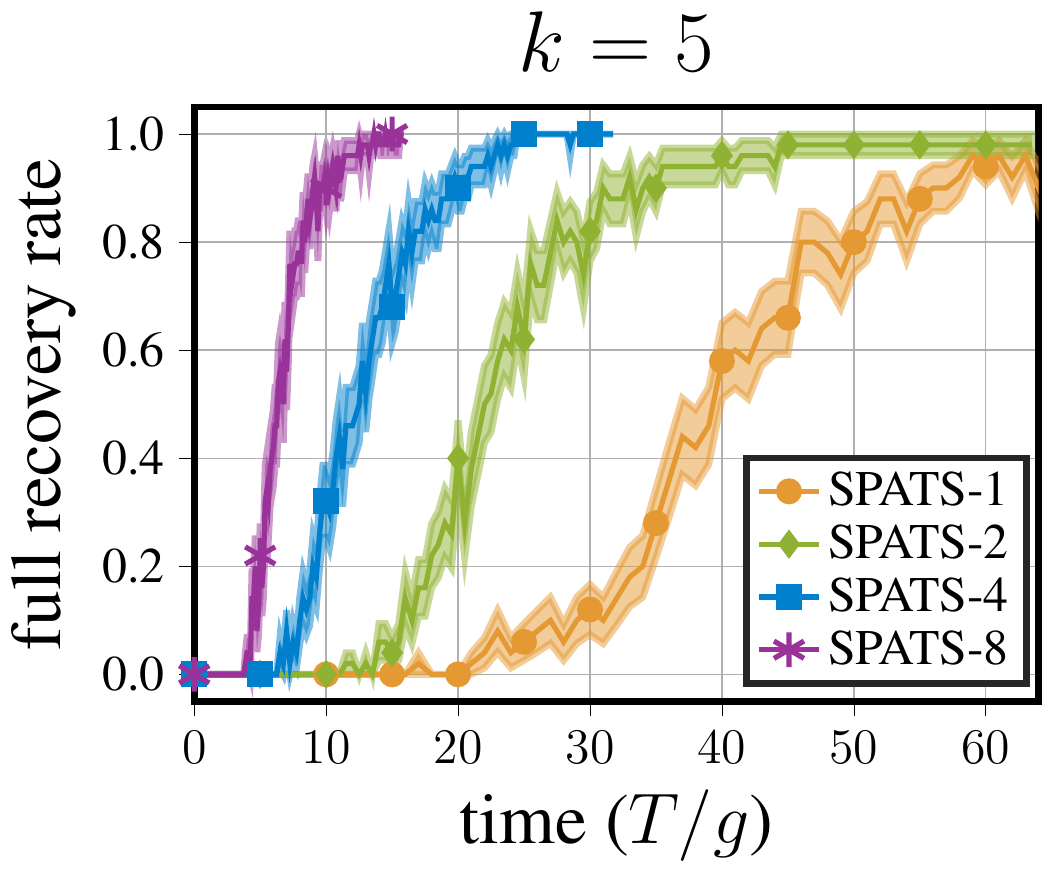}
	\end{subfigure}
	\caption{Full recovery rate of SPATS with 1, 2, 4 and 8 agents for sparsity rates $k=1, 5$}
	\vspace*{-2.2mm}
	\label{fig:TS_plots}
\end{figure}
\begin{figure}[H]
	\centering
	\begin{subfigure}{0.245\linewidth}
		\includegraphics[width=\linewidth]{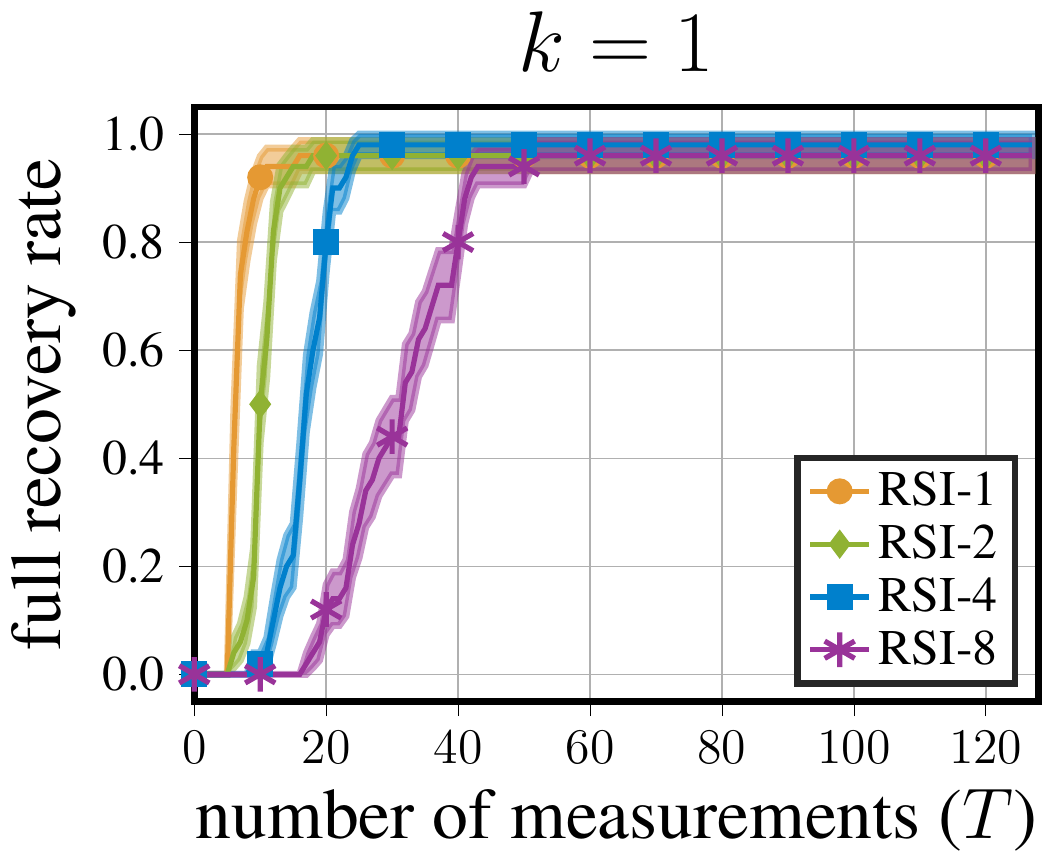}
	\end{subfigure}
	\begin{subfigure}{0.245\linewidth}
		\includegraphics[width=\linewidth]{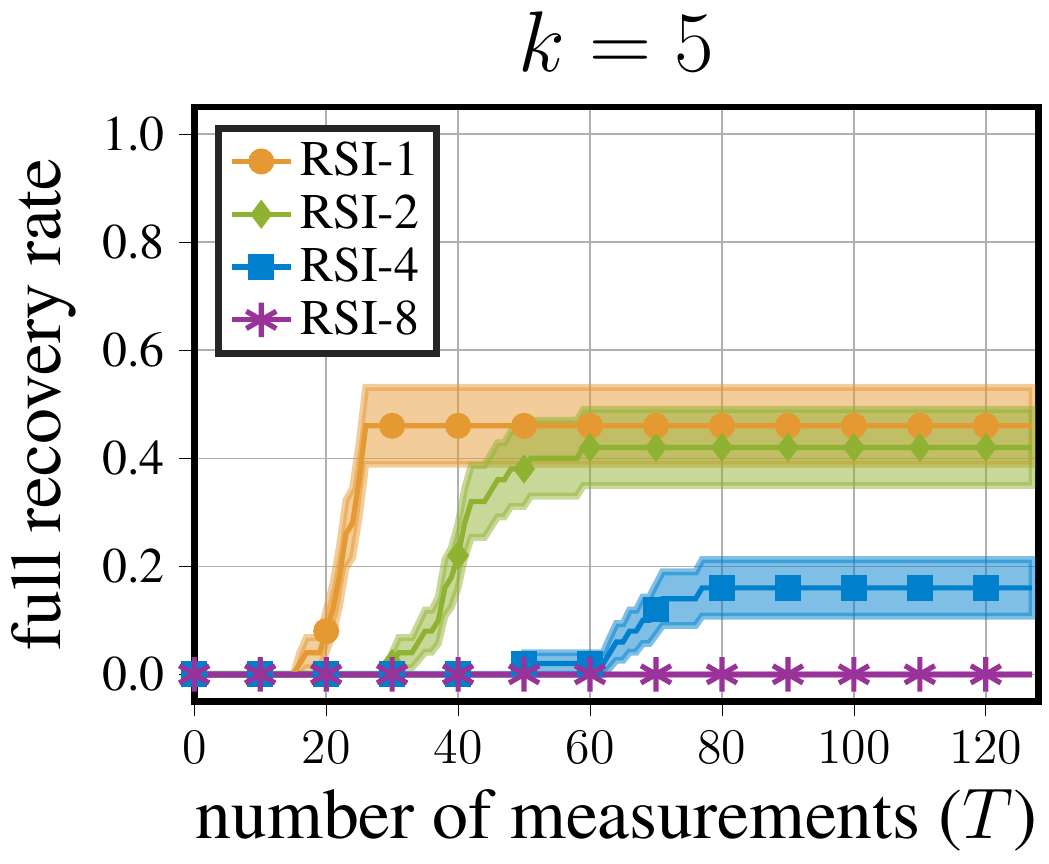}
	\end{subfigure}
	\begin{subfigure}{0.245\linewidth}
		\includegraphics[width=\linewidth]{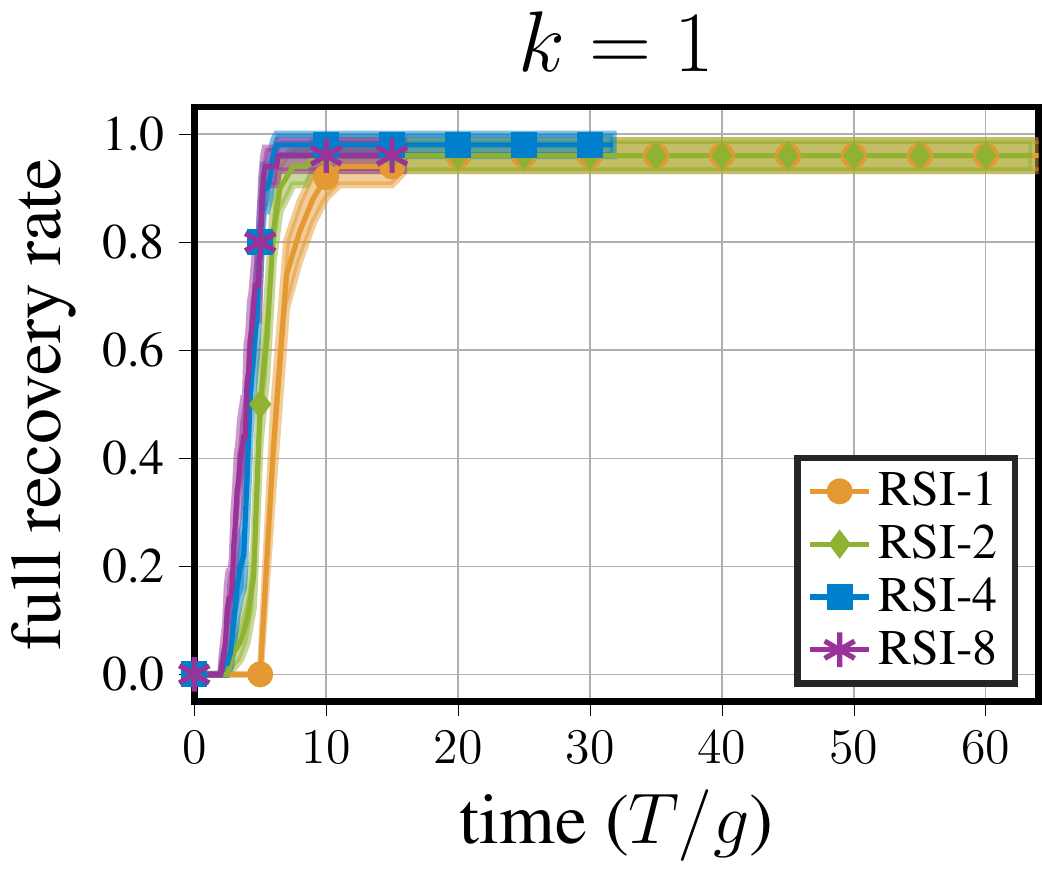}
	\end{subfigure}
	\begin{subfigure}{0.245\linewidth}
		\includegraphics[width=\linewidth]{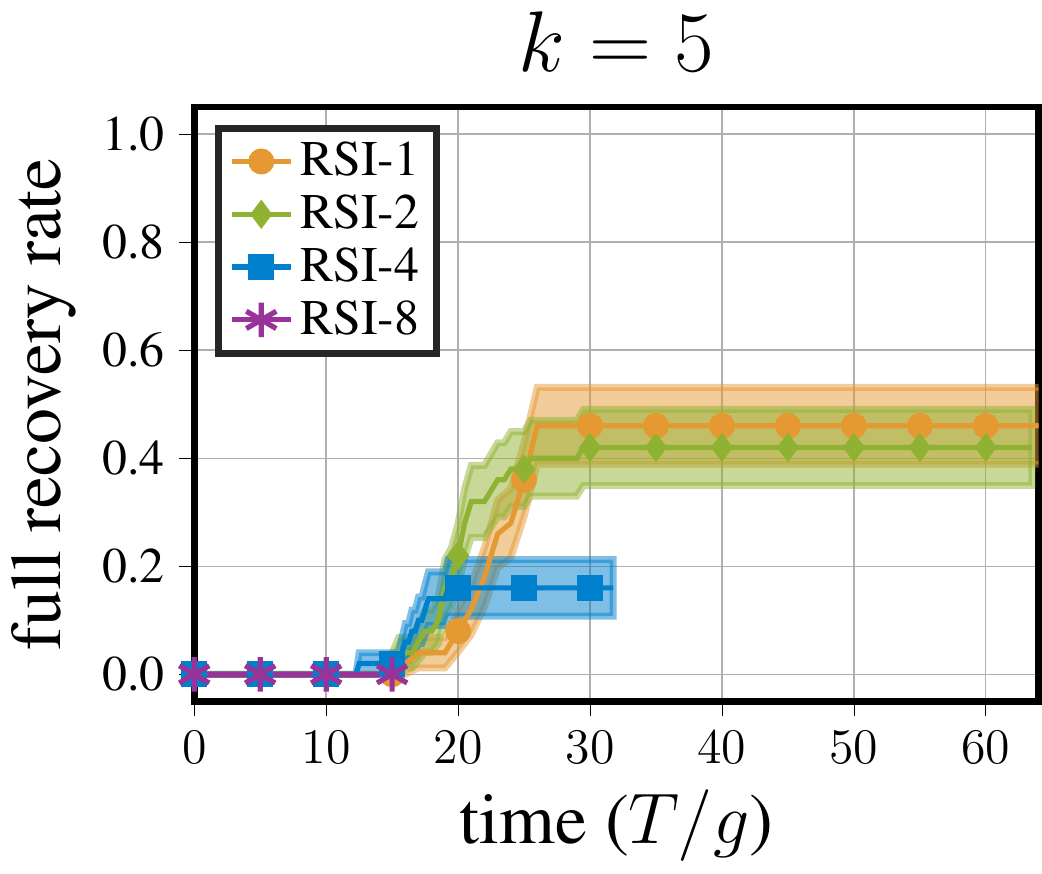}
	\end{subfigure}
	\caption{Full recovery rate of RSI with 1, 2, 4 and 8 agents for sparsity rates $k=1, 5$}
	\vspace*{-2.2mm}
	\label{fig:RSI_plots}
\end{figure}
\begin{figure}[H]
	\centering
	\begin{subfigure}{0.245\linewidth}
		\includegraphics[width=\linewidth]{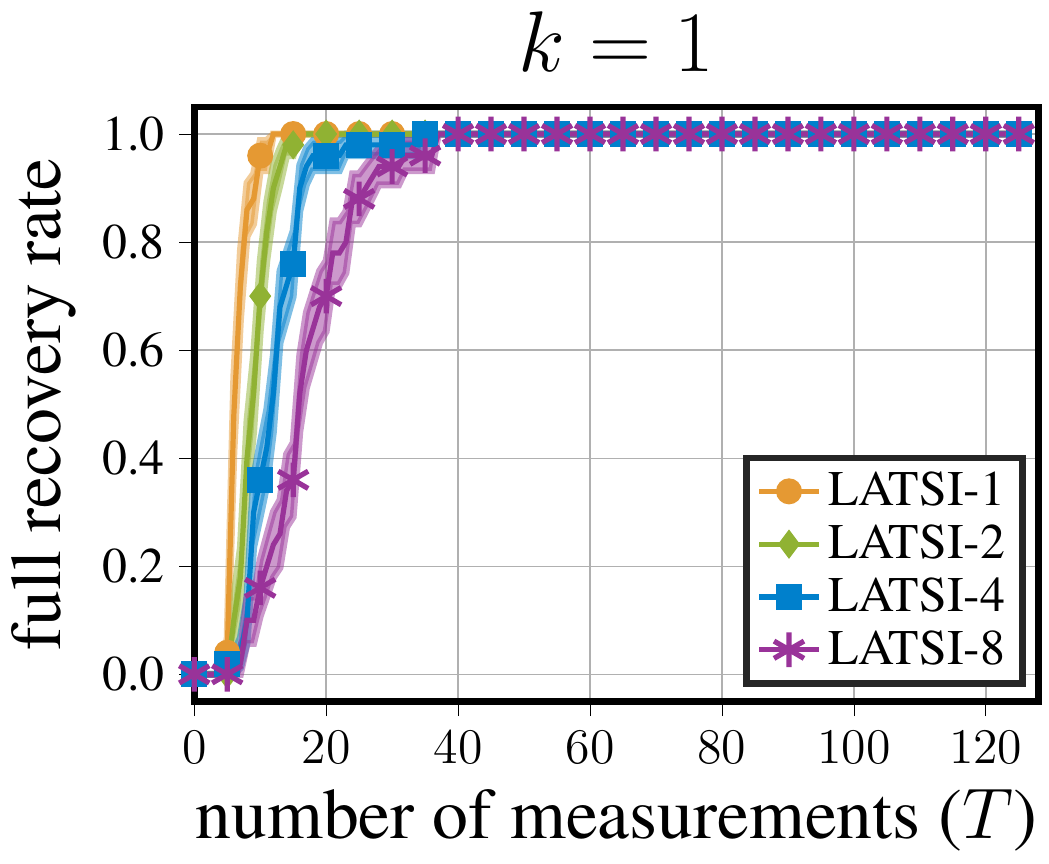}
	\end{subfigure}
	\begin{subfigure}{0.245\linewidth}
		\includegraphics[width=\linewidth]{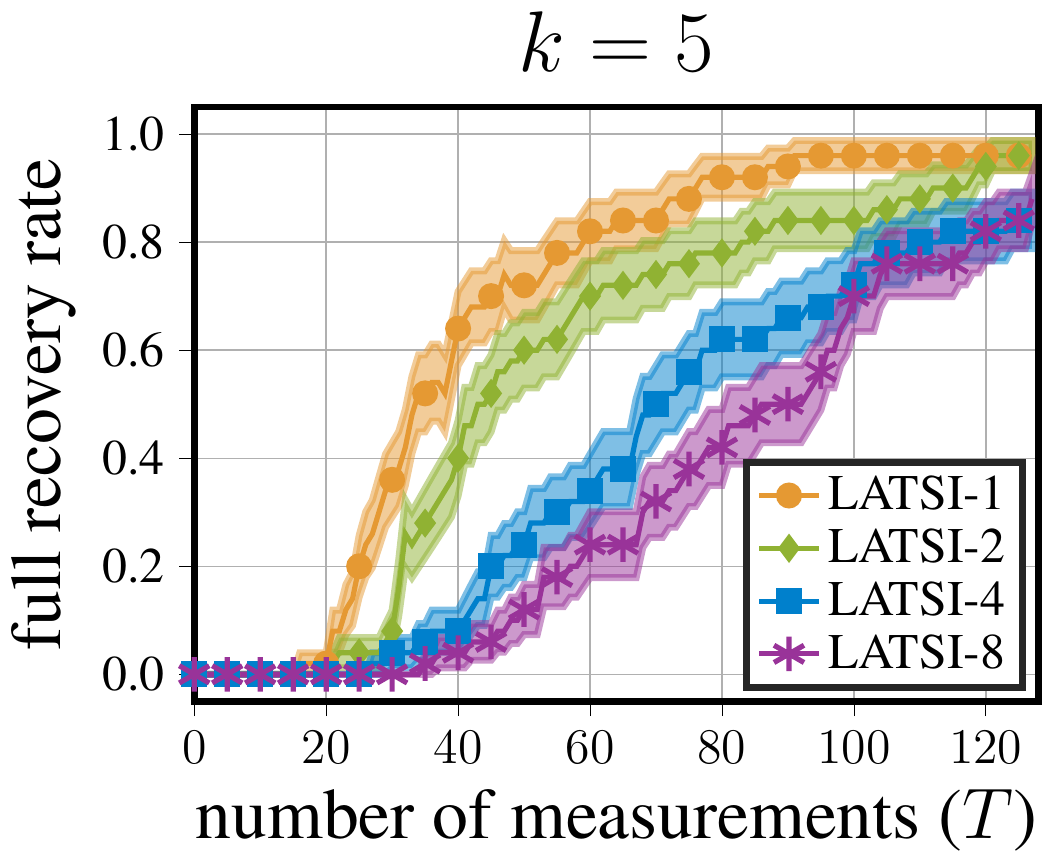}
	\end{subfigure}
	\begin{subfigure}{0.245\linewidth}
		\includegraphics[width=\linewidth]{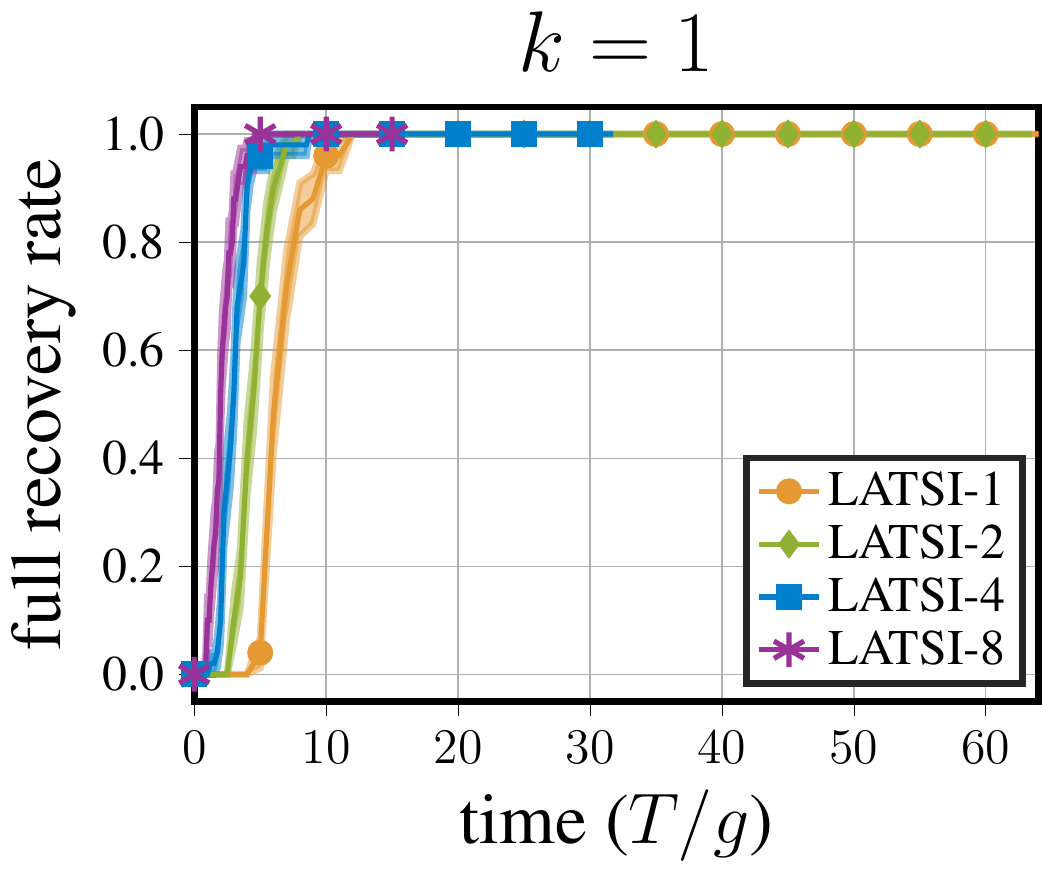}
	\end{subfigure}
	\begin{subfigure}{0.245\linewidth}
		\includegraphics[width=\linewidth]{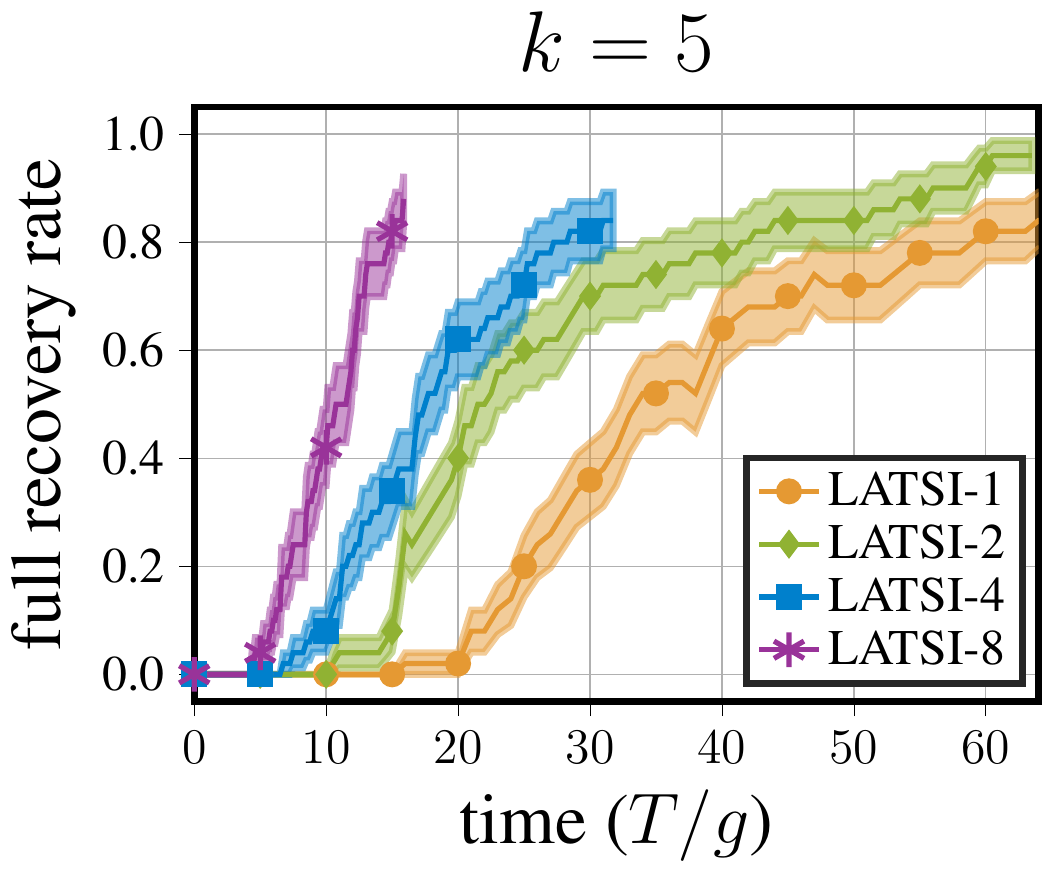}
	\end{subfigure}
	\caption{Full recovery rate of LATSI with 1, 2, 4 and 8 agents for sparsity rates $k=1, 5$}
	\label{fig:LATSI_plots}
\end{figure}

\subsection{Multi Agent}\label{sec:multi-agent}
%
\fref{fig:TS_plots}, \ref{fig:RSI_plots} and \ref{fig:LATSI_plots} show the performance of SPATS, RSI and LATSI in a multi agent setting, respectively. Each figure consists of 4 sub-figures where the left two illustrate the full recovery rate for $k=1$ and $k=5$ as a function of number of measurements taken by all the agents ($T$). To better demonstrate the multi agent performance, we also plot full recovery rate as a function of time which is computed by dividing the number of measurements $T$ by the number of agents $g$.
In each sub-figure we vary the number of available agents between $1,2,4$ and $8$.
Furthermore, in all subsequent plots, LATSI-g, RSI-g or SPATS-g indicate the corresponding algorithm with $g$ agents available.

%

\textbf{SPATS: }
We see in \fref{fig:TS_plots} that by increasing the number of agents $g$, SPATS become $g$ times faster as evident in the right two sub-figures. From the left two sub-figures, we can draw a similar conclusion. That is, increasing the number of agents from $1$ to $2$ to $4$ and $8$ hardly changes the total number of measurements required for a given recovery rate, i.e. the average number of sensing actions per agent is improved about $g$ times.
This result demonstrates that SPATS can efficiently perform active search in an asynchronous parallel fashion. 

%
%
%
%
\textbf{RSI: }We extend the RSI algorithm of \cite{ma2017active} to multi agent setting by allowing each agent to independently choose its sensing action given RSI's acquisition function and utilizing the available measurements from other agents. Looking at \fref{fig:RSI_plots} for both $k=1$ and $k=5$, 
we see a significant deterioration in full recovery rate as a function of $T$ as the number of agents increases. The reason is that without randomness in RSI's reward function, agents that are working at the same time are repeating the same sensing actions. For $k=5$, this performance reduction is also obvious as a function of time. However, for $k=1$ RSI performs slightly better in time by increasing agents. The reason for this contradicting behavior is that RSI's performance for $k=1$ is so close to optimal (binary search) that it reaches recovery rate of $1$ before the multi agent system can negatively affect it.

\textbf{LATSI: }Looking at \fref{fig:LATSI_plots}, we see that similar to SPATS, LATSI's multi-agent performance improves in time by increasing the number of agents. 

\textbf{SPATS vs. LATSI vs. RSI: } In \fref{fig:ALL_plots}, we plot all three algorithms against each other for $4$ agents. Here, for $k=1$, RSI and LATSI outperform SPATS due to their information-theoretic approach in computing the reward function. For $k=5$, SPATS outperforms both RSI and LATSI. This is because SPATS is carefully designed to use randomness from TS in its reward function such that multiplying the number of agents would multiply its recovery rate. Furthermore, LATSI performs significantly better than RSI due to the probabilistic exploration aspect of TS in its reward function.
\section{Conclusions}
\label{sec:conclusions}
We have proposed two novel algorithms - SPATS and LATSI which are suitable for the recovery of sparse targets in a multi agent (parallel) asynchronous active search problem with a region sensing constraint. We have discussed the role of sparsity in our design principle and also compared the limitations of a purely information theoretic approach in this setting. An interesting direction for future work would involve constraining the total travel distance of the agents in addition to the number of measurements. Moreover, one could use continuous sensing along a trajectory as an objective rather than only sensing at stopping points.

\bibliographystyle{ieeetr}

\bibliography{confs-jrnls,publishers,RegionSensing}
\balance

%
%
%
%
%
%
%
%
%
%
%
%

\clearpage

\rule{\textwidth}{0.13cm}
\begin{center}
	\LARGE \bf 
	Appendix: \\[0.1cm] Asynchronous Multi Agent Active Search
\end{center}
\rule{\textwidth}{0.04cm}

\vspace{0.3cm}


\appendix

\begin{abstract}
	This appendix includes derivations for the proposed Thompson Sampling algorithms Laplace-TS, SPATS and LATSI, discussion and example on failure mode of Laplace-TS, theoretical proofs for \fref{thm:singleAgent} and \fref{thm:multiAgent} and additional numerical results.
\end{abstract}

\paragraph{Notation}
Lowercase and uppercase boldface letters represent column vectors and matrices, respectively. For a matrix $\bA$, its transpose
$\bA^\Tran$, and the $k$th row and $\ell$th column entry is~$A_{k,\ell}$. For a vector~$\bma$, the $k$th entry is~$a_k$, and the sub-vector containing the $i$th to $j$th entries (excluding $j$) is $\bma[i:j]$ (Python notation). 
The $\ell_1$ and $\ell_2$-norm of~$\bma$ are denoted by $\|\bma\|_1$ and $\|\bma\|_2$,
and $|\bma|$ represents its absolute value applied element-wise.
The Kronecker product is~$\kron$,
and the trace operator is $\tr(\cdot)$. The $N\times N$ identity matrix is denoted by $\bI_{N}$.
%
$\diag(\bma)$ is a square matrix with $\bma$ on the main diagonal.
%
%
$\emptyset$ denotes an empty set and for a set $\setS$, $|\setS|$ shows the number of elements in that set. $\setA \land \setB$ depicts the logical \emph{AND} between two sets $\setA$ and $\setB$. We use symbols $\triangleq$ and $\mathbbm{1}(.)$ as the symbols for mathematical definition and indicator function, respectively.

\section{Laplace-TS: Multi Agent Thompson Sampling with Sparse Prior}\label{app:app-LTS}

%
%
We will now derive the posterior sampling and design stages of Laplace-TS algorithm introduced in \fref{sec:LTS}.
As detailed in this section, we use a Laplace sparse prior $p(\bm\beta) = \frac{1}{(2b)^n}\exp(-\frac{\|\bm\beta\|_1}{b})$ and likelihood distribution $p(\bmy|\bX,\bm\beta) = \setN(\bX \bm\beta,\sigma^2 \bI_{t-1})$ for this problem.

\fakeparagraph{Posterior Sampling } In this stage (also referred to as inference stage in active learning algorithms), we regularly update the posterior distribution given the available data sequence $\bD_{t-1}$. 
Using Bayes rule we have:
\begin{align*}
p(\bm\beta | \bD_{t-1}) = p(\bm\beta | \bX, \bmy) = \frac{1}{Z} p(\bmy | \bX, \bm\beta) p(\bm\beta).
\end{align*}

Unfortunately, computing the normalization factor $Z$ above is intractable for a Bayesian likelihood distribution with a Laplace prior. To compute this posterior, we borrow ideas from \cite{figueiredo2003adaptive}. In particular, we substitute the Laplace prior distribution with a zero-mean Gaussian prior per entry $p(\beta_i|\tau_i) = \setN(0,\tau_i)$, where its variance $\tau_i$ has an exponential hyper prior $p(\tau_i)=\frac{\eta}{2} \exp(-\frac{\eta}{2}\tau_i)$ with $\tau_i \ge 0$.

Integrating out $\tau_i$ to compute $p(\beta_i)$, we see that this new prior is equivalent to the original Laplace prior per entry $\beta_i$:
\begin{align*}
p(\beta_i) = \int p(\beta_i|\tau_i)p(\tau_i) \, \text{d} \tau_i = \frac{\sqrt{\eta}}{2} \exp(-\sqrt{\eta}|\beta_i|).
\end{align*}
Here, $\sqrt{\eta}=1/b$ is the scaling hyperparameter in Laplace distribution and for best performance needs to be tuned given the sparsity rate of the original signal $\bm \beta$. If we were able to observe variance $\tau_i$, the posterior distribution of $\bm\beta$ given data sequence $\bD_{t-1}$ would become:
\begin{align}\nonumber
p(\bm\beta|\bD_{t-1},\bm\tau) &= \frac{1}{Z} \,\, p(\bmy | \bX, \bm\beta) p(\bm\beta|\bm\tau) \\\nonumber
&= \frac{1}{Z} \,\, \setN(\bX \bm\beta,\sigma^2 \bI_{t-1}) \times \setN(0,\text{diag}(\bm\tau)) \\
&= \setN(\bm\mu_{\bm\beta}(\bm \tau),\bm\Sigma_{\bm\beta}(\bm \tau)), \label{eq:betagiventau}
\end{align}
with, 
\vspace{-5mm}
\begin{align}\nonumber
\bm\Sigma_{\bm\beta}(\bm \tau) &= \left(\left(\text{diag}(\bm\tau)\right)^{-1}+\frac{1}{\sigma^2} \bX^\Tran \bX \right)^{-1},\\
\bm\mu_{\bm\beta} (\bm \tau)& = \frac{1}{\sigma^2} \bm\Sigma_{\bm\beta}(\bm \tau) \bX^\Tran \bmy. \label{eq:meanVar}
\end{align}
Here, $\bm\tau$ is the vector containing all elements of $\tau_i$ for $i=1,...,n$. Since we cannot observe $\bm\tau$, we use Gibbs sampling \cite{gelfand1990sampling} to iteratively sample $\bm\tau$ and $\bm\beta$ from their conditional posterior distributions. For the given prior, \cite{bae2004gene} has computed the conditional distributions to iteratively sample $\bm\tau$ and $\bm\beta$ as follows:
\begin{align}\nonumber
\bm\beta^\star &\sim \setN(\bm\mu_{\bm\beta}(\tilde{\bm \tau}),\bm\Sigma_{\bm\beta}(\tilde{\bm \tau}))\label{eq:betaGibbs}\\
\tilde{\tau}_i^{-1} &\sim InvGauss(\frac{\sqrt{\eta}}{\beta_i^\star},\eta),\quad i=1,...,n.
\end{align}
%
%
%
%
%
%
%
%
%
%
\fakeparagraph{Design }
In this stage, we wish to compute the expected reward $ \lambda^+(\bm\beta^\star,\bD_{t-1},\bmx)$ and optimize it for best sensing action $\bmx_t$. Let us use the reward function $\lambda(\bm\beta^\star, \bD_{t-1}\cup (\bmx,y)) = -\|\bm\beta^\star - \hat{\bm\beta}(\bD_{t-1}\cup (\bmx,y))\|_2^2$. In order to compute the expected reward in \fref{eq:expectedReward}, we need to design an estimator $\hat{\bm\beta}(\bD_{t-1}\cup (\bmx,y))$ whose expectation we can compute. While there has been many sparse recovery algorithms proposed in the literature \cite{marques2018review}, for many of these well-known thresholding or iterative algorithms (e.g. \cite{pati1993orthogonal,needell2010cosamp,blumensath2009iterative,daubechies2004iterative,maleki2011approximate}) computing the expectation in \fref{eq:expectedReward} is not tractable. 
%
%
%
Alternatively, we propose using maximum a posteriori (MAP) estimate of the posterior $p(\bm \beta|\bD_{t-1},\bm \tau)$ in \fref{eq:betagiventau} where we estimate $\bm \tau$ using Expectation-Maximization (EM) proposed in \cite{figueiredo2003adaptive} with the E-step and M-step for $j = 1,..., J$ iterations as follows:
\begin{align}\nonumber
&\text{\textbf{E-step}}: \bm\tau^{(j)} =  \,\, |\hat{\bm\beta}^{(j-1)}|/\eta\\
&\text{\textbf{M-step}}: \hat{\bm\beta}^{(j)} = \bm\mu_{\bm\beta} (\bm \tau^{(j)}).
\label{eq:EMsolution}
\end{align}
Since many elements of $\hat{\bm\beta}^{(j)}$ and consequently $\bm\tau^{(j)}$ are expected to be zero,  to avoid inverting $\text{diag}(\bm\tau)$ we can rewrite the M-step as follows: $\hat{\bm\beta}^{(j)} =\frac{1}{\sigma^2} \bm \Sigma^{(j)} \bX^\Tran \bmy$, with $\bm \Sigma^{(j)} = \bU \left( \bI_{n}+ \frac{1}{\sigma^2} \bU \bX^\Tran \bX \bU \right)^{-1} \bU$ and $\bU = \left(\text{diag}(\bm\tau^{(j)})\right)^{1/2}$.

%

Since the posterior is Gaussian, its MAP estimate is the mean of the distribution, i.e. $\hat{\bm\beta}^{(J)}$ in \fref{eq:EMsolution}.
%
Adding $(\bmx,y)$ to the data sequence $\bD_{t-1}$,
%
%
%
our MAP estimate becomes:
\vspace*{-1.5mm}
\begin{align*}
\hat{\bm\beta}(\bD_{t-1} \cup (\bmx,y)) =
\underbrace{\bU \! \left(\! \sigma^2 \bI_{n}+ \bU \begin{bmatrix} \bX^\Tran\!&\!\bmx \end{bmatrix} \! \begin{bmatrix} \bX\\\bmx^\Tran \end{bmatrix} \bU \right)^{-1} \! \bU }_{\triangleq \, \bmq} \begin{bmatrix} \bX^\Tran \!&\!\bmx \end{bmatrix}  \begin{bmatrix} \bmy\\y \end{bmatrix}.
\end{align*}
\vspace{-1mm}
Now, we can derive the expected reward in \fref{eq:expectedReward} as follows:
\begin{align}\nonumber
\lambda^+(\bm\beta^\star\!\!,\bD_{t-1},\bmx) &= \Ex{y|\bmx,\bm\beta^\star}{-\|\bm\beta^\star\!\! - \hat{\bm\beta}(\bD_{t-1}\cup (\bmx,y))\|_2^2}
= - \Ex{y|\bmx,\bm\beta^\star}{\| \bm q\bX^\Tran \bmy + \bm q \bmx y - \bm\beta^\star \|_2^2}\\\nonumber
&=- \| \bm q\bX^\Tran \bmy - \bm\beta^\star \|_2^2
-\|\bm q \bmx\|_2^2 \Ex{y|\bmx,\bm\beta^\star}{y^2}-2 \left(\bm q\bX^\Tran \bmy - \bm\beta^\star\right)^\Tran \bm q \bmx \Ex{y|\bmx,\bm\beta^\star}{y}\\\label{eq:lambda+}
&=- \| \bm q\bX^\Tran \bmy - \bm\beta^\star \|_2^2-\|\bm q \bmx\|_2^2 \left(\sigma^2 + (\bmx^\Tran \bm\beta^\star)^2 \right) -2 \left(\bm q\bX^\Tran \bmy - \bm\beta^\star\right)^\Tran \!\! \bm q \bmx \bmx^\Tran \bm\beta^\star.
\end{align}
%
%
%
%
With the posterior sampling and design stage developed, we can now run \fref{alg:TS} for the active search problem in \fref{sec:formulation}. Let us call this algorithm Laplace-TS. In the next section, we will evaluate the performance of this algorithm.


\section{Failing Performance of Laplace-TS in Single Agent Setting}\label{app:LTSfail}

\begin{figure}[H]
	\centering
	\begin{subfigure}{0.49\linewidth}
		\centering
		\includegraphics[width=0.75\linewidth]{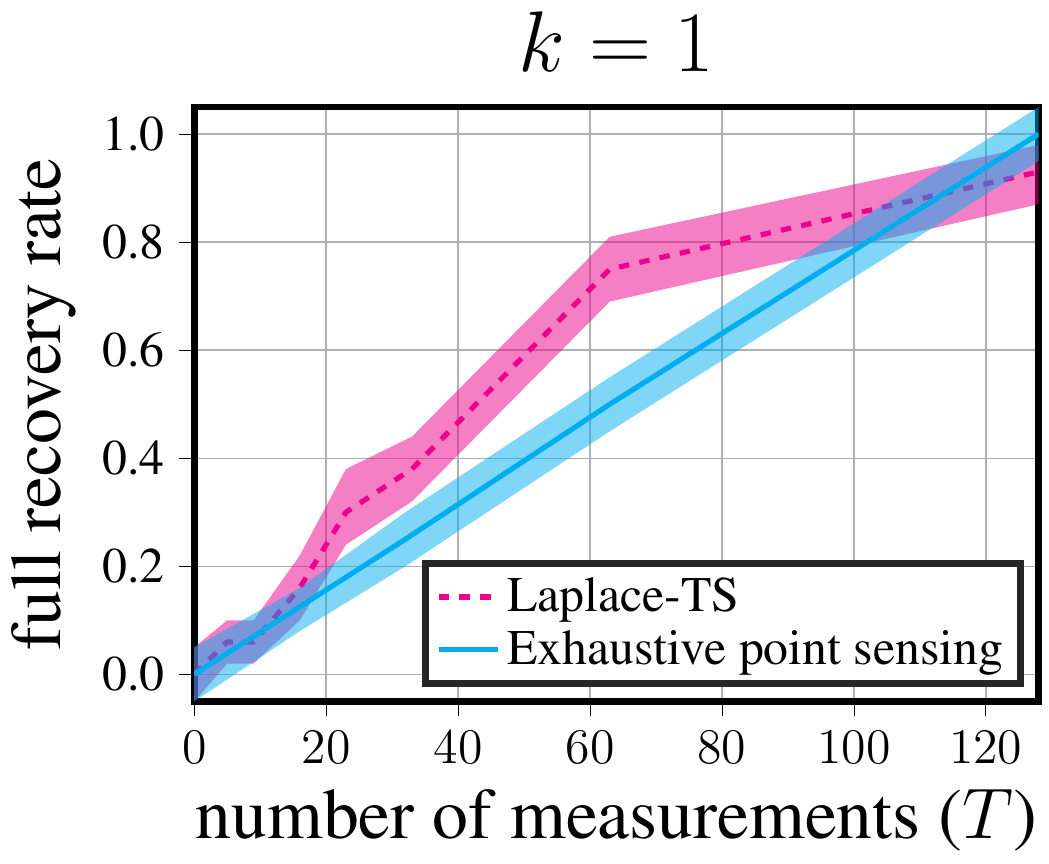}
	\end{subfigure}
	\begin{subfigure}{0.49\linewidth}
		\centering
		\includegraphics[width=0.75\linewidth]{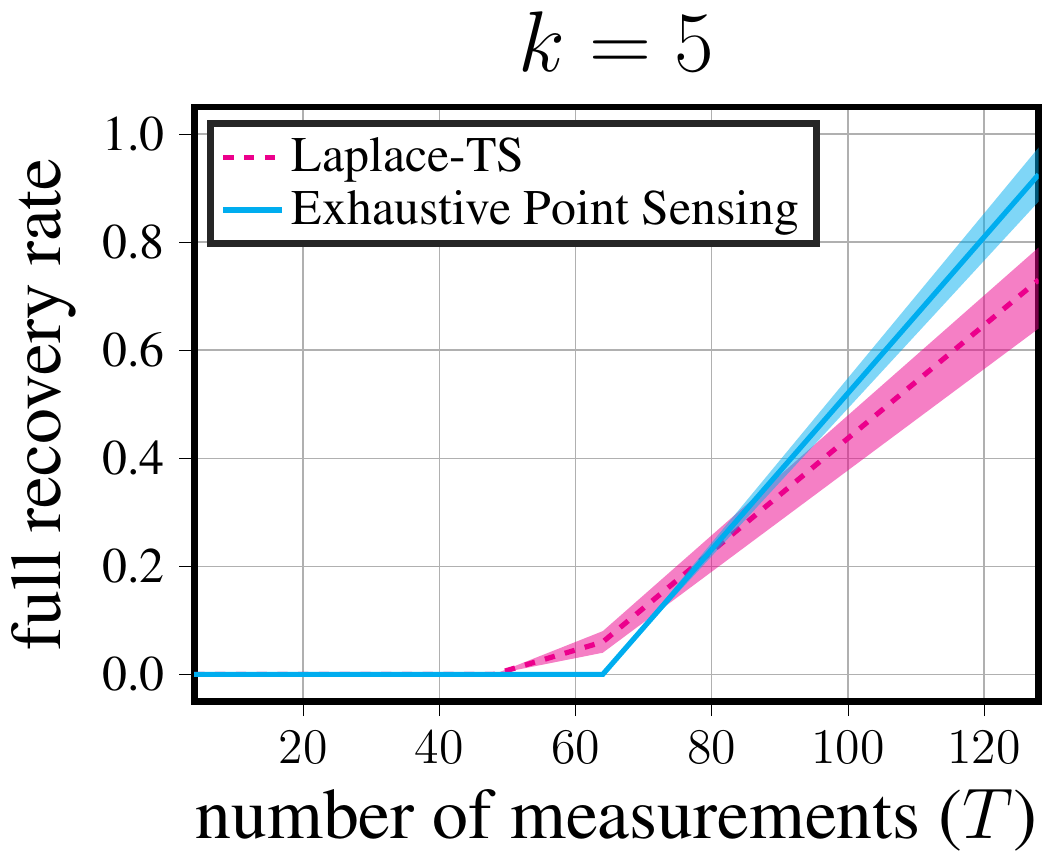}
	\end{subfigure}
	\caption{Full recovery rate of Laplace-TS with single agent for sparsity rate $k=1$}
	\label{fig:LaplaceTS}
\end{figure}

\fref{fig:LaplaceTS} illustrates simulation results of the full recovery performance of Laplace-TS compared to a simple point algorithm that exhaustively searches the entire action space one location at a time with one agent. Here, we are plotting full recovery rate of vector $\bm \beta$ with size $n=128$ as a function of number of measurements $T$ for two sparsity rates of $k=1$ and $k=5$. The curves show mean and standard error of full recovery over $50$ random trials. 
This figure demonstrates that, unfortunately, Laplace-TS with one agent leads to poor performance that is on par with a point-wise algorithm that exhaustively searches all locations one at a time. 

We can associate this poor performance with one of the failure modes of TS discussed in Sec. 8.2 of the tutorial by \cite{russo2018tutorial}. According to this tutorial, TS faces a dilemma when solving certain kinds of active learning problems. One such scenario are problems that require a careful assessment of information gain. In general, by optimizing the expected reward, TS always restricts its actions to those that have a chance in being optimal. 
However, in certain active learning problems, suboptimal actions can carry additional information regarding the parameter of interest. 

We provide the following example to better understand the situation in our problem formulation.
Let us
assume there is no noise in the sensing model \fref{eq:formulation} ($\epsilon_t=0$) and that there is only one non-zero element ($k=1$). Under such conditions, the problem amounts to finding the location of the non-zero element in $\bm \beta$ which we call $\tilde{i}$. Thus, for every action $\bmx_t$, the observation $y_t = \bmx_t^\Tran \bm \beta$ is non-zero if $\tilde{i}$ is in the support of $\bmx_t$ and is zero, otherwise. Clearly, a binary search can locate $\tilde{i}$ in $\log(n)$ steps. 
TS, however, in each time step estimates the sample $\bm\beta^\star$ with $\hat{\bm\beta}$ and picks the sensing action $\bmx_t = \hat{\bm\beta}$. Or, in case of perfect estimation, we will have $\bmx_t = \hat{\bm\beta} = \bm \beta^\star$.
So, unless $\bm \beta^\star$ is the true $\bm \beta$, observation $\bmy_t$ would be zero and TS will only manage to eliminate support of $\bm \beta^\star$ from the list of possibly correct supports. Therefore, TS ends up eliminating wrong supports one location at a time which explains its on par performance to an exhaustive point sensor.
Similarly, for $k >1$ TS is always limited to picking sensing actions $\bmx_t$ that are in the support of the sparse estimates of the samples. Adding the region sensing constraint will only aggravate this problem by further shrinking this support set.
%
In the next section, we will modify Laplace-TS and propose two algorithms that can bypass this failure mode.

\section{SPATS: Thompson Sampling with Block Sparse Prior}\label{app:blockTS}

In what follows, we will derive the posterior sampling and design stages for SPATS algorithm proposed in \fref{sec:SPATS}. SPATS is an asynchronous multi agent TS algorithm using a block sparse prior for the problem formulation in \fref{sec:formulation}. 

As discussed in \fref{sec:SPATS}, we use the following prior and likelihood distributions. The likelihood function is $p(\bmy|\bX,\bm\beta) = \setN(\bX \bm\beta,\sigma^2 \bI_{t-1})$ from \fref{eq:formulation2}. For prior distribution, as discussed in \fref{sec:SPATS} we use a block sparse prior $p(\bm \beta) = \setN(\bZero_{n \times 1},\bm \Sigma_0)$, where:
%
%
%
\begin{align}\label{eq:Sigma0}
\bm \Sigma_0 = \text{diag}(\bm \gamma) \kron \bB,
\end{align}
with $\bm\gamma \in \reals^M$ and $\bB \in \reals^{L \times L}$ ($M=n/L$) as hyperparameters.
%
%

\fakeparagraph{Posterior Sampling }
Similar to \fref{eq:betagiventau} and \fref{eq:meanVar} in computing posterior distribution for Laplace-TS, the posterior with a Gaussian likelihood and Gaussian prior becomes:
\begin{align}\label{eq:blockposterior}
p(\bm\beta|\bD_{t-1},\bm \gamma,\bB) = \setN(\bm\mu_{\bm\beta}(\bm \gamma,\bB),\bm\Sigma_{\bm\beta}(\bm \gamma,\bB)),
\end{align}
with, 
\begin{align*}
\bm\Sigma_{\bm\beta}(\bm \gamma,\bB) &= (\left(\bm \Sigma_0\right)^{-1}+\frac{1}{\sigma^2} \bX^\Tran \bX )^{-1},\\
\bm\mu_{\bm\beta} (\bm \gamma,\bB) &= \frac{1}{\sigma^2} \bm\Sigma_{\bm\beta}(\bm \gamma,\bB)\bX^\Tran \bmy.
\end{align*}

Using Expectation-Maximization proposed in \cite{zhang2011sparse}, we can estimate the hyperparameters $\bm \gamma$ and $\bB$ for $j=1,...,J$ iterations as follows.
\begin{align}\nonumber
\text{ \textbf{E-step}}: \,\, \bm\mu_{\bm\beta}^{(j)} &= \bm\mu_{\bm\beta}(\bm \gamma^{(j-1)},\bB^{(j-1)}),  \\\nonumber
\bm\Sigma_{\bm\beta}^{(j)} &= \bm\Sigma_{\bm\beta}(\bm \gamma^{(j-1)},\bB^{(j-1)})
\\\nonumber
\text{\textbf{M-step}}:\,\, \gamma_m^{(j)} &= \frac{\tr(\bB^{-1}(\bm \Sigma_{\bm\beta}^m+\bm \mu_{\bm\beta}^m \left({\bm \mu_{\bm\beta}^m}\right)^\Tran))}{L} , \quad (m=1,...,M)\\	\bB^{(j)} &= \frac{1}{M} \sum_{m=1}^{M} \frac{\Sigma_{\bm\beta}^m+\bm \mu_{\bm\beta}^m \left({\bm \mu_{\bm\beta}^m}\right)^\Tran}{\gamma_m}, \label{eq:blockEM}
\end{align}
with
$\bm \mu_{\bm\beta}^m = \mu_{\bm\beta}^{(j)}[(m-1)\times L:m \times L]$, and
$\bm \Sigma_{\bm\beta}^m = \bm \Sigma_{\bm\beta}^{(j)}[(m-1) \times L:m \times L,(m-1) \times L:m \times L]$.
%
%

\fakeparagraph{Design}
Next, for the estimator $\hat{\bm \beta}(\bD_{t-1})$, we use the MAP estimator given by $\bm\mu_{\bm\beta}^{(J)}$. Particularly, we have:
\begin{align}\label{eq:blockestimator}
\hat{\bm\beta}(\bD_{t-1} \!\cup\! (\bmx,y)) \!= \! \underbrace{\left(\!\! \sigma^2 \bm \Sigma_0^{-1}\!\!+\!\!  \begin{bmatrix} \bX^\Tran\!\!&\!\!\bmx \end{bmatrix} \!\! \begin{bmatrix} \bX\\\bmx^\Tran \end{bmatrix} \! \right)^{\!\!\! -1}  }_{\bmq} \!\!\!  \begin{bmatrix} \bX^\Tran \!\!&\!\!\bmx \end{bmatrix}  \!\! \begin{bmatrix} \bmy\\y \end{bmatrix}
\end{align} 
with $\bm \Sigma_0 = \text{diag}(\bm \gamma^{(J)}) \kron \bB^{(J)}$.
Hence, the reward function for this estimator is given by \fref{eq:lambda+} with $\bmq$ from \fref{eq:blockestimator}.

\section{Proof for \fref{thm:singleAgent}}\label{app:thmSingleAgent}
Since this theorem focuses on single agent settings, both proposed TS algorithms in the theorem have the theoretical guarantees as well as the limits of TS bounds as summarized in \cite[Ch. 8]{russo2018tutorial}. As discussed in this chapter, there are different theoretical approaches proposed in literature for regret bounds on TS. For this proof, we use regret bounds computed via information theory proposed in \cite{russo2016information}. According to this paper, the resulting bounds better describe the benefits of prior distribution which helps point to failure modes of TS and potential solutions. 
Defining $\pi_t$ as the distribution of actions for an online optimization policy at time $t$, \cite{russo2016information} shows that for this policy, the expected regret defined in \fref{thm:singleAgent} is bounded by:
\begin{align}\label{eq:infoRegretBound}
\Ex{}{\text{Reg}(T)} \le \sqrt{T\bar{\Psi}_t(\pi) \text{H}(\bmx^\star)},
\end{align}
where, $\text{H}(\bmx^\star)$ is the entropy for optimal action $\bmx^\star$ and $\bar{\Psi}_t(\pi)$ is the average expected information ratio defined as follows.
\begin{align}\label{eq:infoRatio}
\bar{\Psi}_t(\pi)=\frac{1}{T} \sum_{t=1}^{T} \Ex{\pi_t} {\Psi_t(\pi_t)} \quad \& \quad \Psi_t(\pi_t) \triangleq  \frac{\Delta_t(\pi_t)^2}{\setI_t(\pi_t)},
\end{align}
where, $\Delta_t(\pi_t)$ and $\setI_t(\pi_t)$ are the single-period expected regret and information gain over policy $\pi_t$ at time step $t$ defined below. For the single-period expected regret we have:
\begin{align}\label{eq:singleReg}
\Delta_t(\pi_t) = \sum_{\bmx \in \setX} \pi(\bmx) \, \Delta_t(\bmx) \quad \& \quad \Delta_t(\bmx) = \Ex{}{\setR_t(\bmx^\star,\bm\beta) - \setR_t(\bmx,\bm\beta)},
\end{align}
where, $\setR_t(\bmx,\bm\beta)$ is the reward function for action $\bmx$ at time $t$.
And, for information gain, we have:
\begin{align}\label{eq:infoGain}
\setI_t(\pi_t) = \sum_{\bmx \in \setX} \pi(\bmx)\, \setI_t(\bmx^\star;y|\bmx),
\end{align}
where, $\setI_t(\bmx^\star;y|\bmx)$ is the mutual information between optimal action $\bmx^\star$ and observation $y=(\bmx^T \bm\beta)^2$ at time $t$.

To compute the regret bound in \fref{eq:infoRegretBound}, we need to first focus on computing single-period expected regret and information gain in \fref{eq:singleReg} and \fref{eq:infoGain}. Computing these terms for a general problem formulation can be difficult. However, for an online optimization problem with a finite prior distribution and finite action set, \cite[Algorithm 1]{russo2017learning} has provided the necessary formulation to compute these terms. Since our problem formulation falls under the category of this framework, we can take advantage of their formulations as follows. For single-period expected regret, we have:
\begin{align}\label{eq:finiteSingleReg}
\Delta_t(\bmx) = \setR(\bmx^\star) - \sum_{\bm\beta}{p(\bm\beta) \setR(\bmx,\bm\beta)},
\end{align}
and, for the mutual information we have:
\begin{align}\label{eq:finiteInfoGain}
\setI_t(\bmx^\star;y|\bmx) = \sum_{\bmx^\star,y} p(\bmx^\star,y|\bmx) \log \left(\frac{p(\bmx^\star,y|\bmx)}{p(\bmx^\star)p(y|\bmx)}\right),
\end{align}
where,
\begin{align}\nonumber
p(\bmx^\star) &= \sum_{\bm\beta \in \setB(\bmx^\star)} p(\bm\beta|\bD_{t-1}), \\\nonumber
p(y|\bmx) &= \sum_{\bm\beta} p(\bm\beta|\bD_{t-1}) p(y|\bmx,\bm\beta),\\\label{eq:infoProbabilities}
p(\bmx^\star,y|\bmx) &= p(\bmx^\star) \sum_{\bm\beta \in \setB(\bmx^\star)} p(y|\bmx,\bm\beta),
\end{align}
with $\setB(\bmx) = \{\bm\beta: \bmx = \argmax_{\tilde\bmx} \Ex{y|\tilde\bmx,\bm\beta}{\setR(\tilde\bmx,\bm\beta)}\}$ as the set containing all vectors $\bm\beta$ for which action $\bmx$ is optimal.

Next, we will compute this regret bound for our TS algorithms with 1-sparse and block sparse prior introduced in \fref{thm:singleAgent}.
Our emphasis here is on understanding the benefits of a block sparse prior distribution. Hence, to simplify the discussion we assume both algorithms are aware of the true 1-sparse prior for $\bm\beta$.

\subsection*{1-sparse Prior}
Let us here establish two parameters $i^\star$ and $j$ that we will use throughout the proof.
1) Since the true parameter $\bm\beta$ is 1-sparse, we assume the location of the non-zero element of true $\bm\beta$ is $i^\star$. Given the uniform 1-sparse prior on true $\bm\beta$, $i^\star$ can be either of indices $1$ through $n$ with probability $1/n$.
2) Since our TS algorithm here assumes a 1-sparse prior distribution, the only feasible sensing actions in $\setX$ are 1-sparse with one nonzero element at location $j$.

With two parameters $i^\star$ and $j$, we can now compute the single-period expected regret in \fref{eq:finiteSingleReg} for $t=1$ as:
\begin{align*}
\Delta_1(\bmx) =\Delta_1(j)= 1 - \sum_{i^\star=1}^{n}{1/n \times \mathbbm{1}(j = i^\star)},
\end{align*}
Hence, the single-period expected regret $\Delta_1(\pi_1)$ over our policy becomes:
\begin{align*}
\Delta_1(\pi_1) = \sum_{\bmx \in \setX} \pi(\bmx) \, \Delta_1(\bmx) \stackrel{(a)}{=} \sum_{j=1}^{n} 1/n \, \Delta_1(j)= \frac{n-1}{n}
\end{align*}
Here, (a) follows from the fact that our TS algorithm at $t=1$ will pick all sensing actions with the nonzero element $j=1$ through $j=n$ equally likely. Next, we need to compute single-period expected regret in \fref{eq:finiteSingleReg} for $t=2$. For $t=2$, there are two policies for our TS algorithm depending on the action $\bmx_1$ chosen in previous time step. If $\bmx_1=\bmx^\star$, then our TS algorithm will be exploiting the same optimal action at time $t=2$ and set $\bmx_2 = \bmx^\star$ for which $\Delta_2(\bmx)=\Delta_2(\pi_2)=0$. The probability of finding $\bmx^\star$ at $t=1$ is $1/n$. However, if $\bmx_1 \ne \bmx^\star$, then our TS algorithm will be exploring the $n-1$ feasible actions that are not observed. Let us assume $j_1$ shows the location of nonzero element in $\bmx_1$, then:
\begin{align*}
\Delta_2(\bmx) =\Delta_2(j)= 1 - \sum_{i^\star=1,i^\star \ne j_1}^{n}{\frac{1}{n-1} \times \mathbbm{1}(j = i^\star)},
\end{align*}
Hence, for both policies the single-period expected regret $\Delta_2(\pi_2)$ becomes: 
\begin{align*}
\Delta_2(\pi_2) = \left\{
  \begin{array}{@{}ll@{}}
 0, & \text{prob}=\frac{1}{n}\\
\sum_{\bmx \in \setX} \pi(\bmx) \, \Delta_2(\bmx) = \sum_{j=1,j \ne j_1}^{n} \,\, \frac{1}{n-1} \, \Delta_2(j)= \frac{n-2}{n-1}, & \text{prob}=\frac{n-1}{n}
\end{array}\right.
\end{align*}
Similarly, we can conclude that for any $t>2$, we will have:
\begin{align}\label{eq:delta1sparse}
\Delta_t(\pi_t) = \left\{
\begin{array}{@{}ll@{}}
0, & \text{prob}=\frac{t-1}{n}\\
 \frac{n-t}{n-t+1}, & \text{prob}=\frac{n-t+1}{n}
\end{array}\right.
\end{align}
We will next compute the information gain in \fref{eq:finiteInfoGain} for time step $t$. Recall that our TS algorithm have two policies. One policy $\pi_t$ is to keep exploiting the optimal action if it has been discovered in the previous $t-1$ measurements $\bD_{t-1}$. Since $\Delta_t(\pi_t)=0$ for this policy, we have $\Psi_t(\pi_t)=  \frac{\Delta_t(\pi_t)^2}{\setI_t(\pi_t)}=0$ and hence there is no need to compute the information term $\setI_t(\pi_t)$ for this scenario. The second policy of our TS method is used if optimal action $\bmx^\star$ has not been found in the previous $t-1$ measurements. For this scenario, at time $t$ the TS algorithm picks the sensing action $\bmx_t$ out of $n-t+1$ measurements that have not been observed yet. For this policy, we can compute the probabilities in \fref{eq:infoProbabilities} as follows. First, for $p(\bmx ^\star)$, we have:
\begin{align*}
	p(\bmx ^\star) =\sum_{\bm\beta \in \setB(\bmx^\star)} p(\bm\beta=\bmx^\star|\bD_{t-1}) = \left\{
	\begin{array}{@{}ll@{}}
	\frac{1}{n-t+1}, \quad & \text{if} \,\,\,    \bmx^\star \notin \bD_{t-1} \,\, \land \,\, \bmx^\star \in \setX\\
	0,& \text{otherwise}
	\end{array}\right.
	\end{align*}
Second, for $p(y|\bmx)$ we have:
	\begin{align*}
	p(y|\bmx) &= p(y|j) =\left\{
	\begin{array}{@{}ll@{}}
	\sum_{i^\star=1,i^\star \ne \{j_1,...,j_{t-1}\}}^{n} \frac{1}{n-t+1} \,\,\, p(i^\star \ne j), \quad & \text{if} \, y=0 ,\\
	\sum_{i^\star=1,i^\star \ne \{j_1,...,j_{t-1}\}}^{n} \frac{1}{n-t+1}  \,\,\, p(i^\star = j), & \text{if} \, y=1 ,
	\end{array}\right.\\
	&=\left\{\begin{array}{@{}lll@{}}
	 \frac{n-t}{n-t+1}\quad & \text{if} \, y=0,\\
	\frac{1}{n-t+1} & \text{if} \, y=1,\\
		0 & \text{otherwise} ,
	\end{array}\right.
\end{align*}
where, $\{j_1,...,j_{t-1}\}$ denotes the location of the nonzero elements in the previous $t-1$ measurements $\bD_{t-1}$. Third, for $p(\bmx^\star,y|\bmx)$ we get:
\begin{align*}
p(\bmx^\star,y|\bmx) = \frac{1}{n-t+1} \sum_{\bm\beta =\bmx^\star} p(y|\bmx,\bm\beta)
=\left\{\begin{array}{@{}lll@{}}
\frac{1}{n-t+1}\quad & \text{if} \,\,\, y=0 \,\, \land \,\, \bmx^\star \ne \bmx,\\
\frac{1}{n-t+1} & \text{if} \,\,\, y=1 \,\, \land \,\, \bmx^\star = \bmx,\\
		0 & \text{otherwise} .
\end{array}\right.
\end{align*}
Using these three probability densities, the information gain in \fref{eq:finiteInfoGain} becomes:
\begin{align*}
\setI_t(\bmx^\star;y|\bmx) &= \sum_{\substack{\bmx^\star \in \setX,\bmx^\star \notin \bD_{t-1}\\\bmx^\star \ne \bmx,y=0}} \frac{1}{n-t+1} \log \left(\frac{\frac{1}{n-t+1}}{\frac{1}{n-t+1} \times \frac{n-t}{n-t+1} }\right) \\
&+ \sum_{\substack{\bmx^\star \in \setX,\bmx^\star \notin \bD_{t-1}\\\bmx^\star = \bmx,y=1}} \frac{1}{n-t+1} \log \left(\frac{\frac{1}{n-t+1}}{\frac{1}{n-t+1} \times \frac{1}{n-t+1} }\right)\\
&= \frac{n-t}{n-t+1}\log \left(\frac{n-t+1}{n-t}\right) + \frac{1}{n-t+1}\log \left(n-t+1\right),
\end{align*}
resulting in:
\begin{align}\label{eq:info1sparse}
\setI_t(\pi_t) = \sum_{\bmx \in \setX,\bmx^\star \notin \bD_{t-1}} \frac{1}{n-t+1}\, \setI_t(\bmx^\star;y|\bmx) = \setI_t(\bmx^\star;y|\bmx).
\end{align}
Now, using \fref{eq:delta1sparse} and \fref{eq:info1sparse} , we can compute the average expected information ratio in \fref{eq:infoRatio} as:
\begin{align*}
\bar{\Psi}_t(\pi)=\frac{1}{T} \sum_{t=1}^{T} \left[\frac{t-1}{n} \times 0 + \frac{n-t+1}{n} \times \frac{\left(\frac{n-t}{n-t+1}\right)^2}{\frac{n-t}{n-t+1}\log \left(\frac{n-t+1}{n-t}\right) + \frac{1}{n-t+1}\log \left(n-t+1\right)}\right],
\end{align*}
which together with an upperbound $\text{H}(\bmx^\star) < \log\left(|\setX|\right)$ and \fref{eq:infoRegretBound} gives the following regret bound:
\begin{align*}
\Ex{}{\text{Reg}(T)} \le \left(\log\left(|\setX|\right) \sum_{t=1}^{T} \frac{\frac{(n-t)^2}{n(n-t+1)}}{\frac{n-t}{n-t+1}\log \left(\frac{n-t+1}{n-t}\right) + \frac{1}{n-t+1}\log \left(n-t+1\right)} \right)^{1/2}.
\end{align*}
Lastly, the regret bound we computed above is only applicable when $T\le n$. Since there are $n$ 1-sparse actions available for a vector $\bm\beta$ with length $n$, after $T=n$ actions the algorithm has surely found the optimal action. As a result, $\Delta_t(\pi_t)=0$ for $t\ge n$ which gives us the regret bound:
\begin{align}\label{eq:regret_1_app}
	\textstyle \Ex{}{\text{Reg}(T)} \leq  \left(\log\!\left(|\setX|\right) \sum_{t=1}^{\min\{T,n-1\}} \frac{\left(1-\frac{t}{n}\right)\left(1-\frac{1}{n-t+1}\right)}{\left(\frac{n-t-1}{n-t} \log\left(\frac{n-t}{n-t-1}\right)+\frac{1}{n-t} \log\left(n-t\right)\right)}\right)^{1/2}
\end{align}

\begin{rem}
	We noticed a minor mistake in expected regret term \fref{eq:regret_1} in \fref{thm:singleAgent} after paper submission. Equation \fref{eq:regret_1_app} above describes the corrected term. This does not affect any of the conclusions in the main paper.
\end{rem}

\subsection*{block sparse prior}
We will again start with establishing two parameters $i^\star$ and $m_t$ that we will use throughout the proof.
1) We assume the location of the non-zero element of true $\bm\beta$ is $i^\star$. Given the uniform 1-sparse prior on true $\bm\beta$, $i^\star$ can be either of indices $1$ through $n$ with probability $1/n$.
2) Since our TS algorithm here assumes a 1-block sparse prior distribution, the feasible sensing actions in $\setX$ at time step $t$ are 
$	\bmx = [ \gamma_1 \bmb,...,\gamma_{M_t} \bmb]^\Tran$ where only one of paramaters $\gamma_1$ through $\gamma_{M_t}$ is nonzero
and $\bmb = [\frac{1}{\sqrt{L_t}},...,\frac{1}{\sqrt{L_t}}]$ are the blocks with length $L_t$. We call the location of the nonzero block $m_t$ which can be either of indices $1$ through $M_t=n/L_t$ with probability $1/M_t$ at time step $t$.

With two parameters $i^\star$ and $m_t$, we can now compute the single-period expected regret $\Delta_1(\pi_1)$ in \fref{eq:singleReg} for $t=1$ as:
\begin{align*}
\Delta_1(\pi_1) &=   \sum_{\bmx \in \setX} \pi(\bmx) \, \Delta_1(\bmx) \stackrel{(b)}{=} \sum_{m_1=1}^{M_1} \frac{1}{M_1} \, \Delta_1(m_1) \\
&\stackrel{(c)}{=}  \sum_{m_1=1}^{M_1} \frac{1}{M_1} \left(1- \sum_{i^\star=1}^{n} 1/n \times \mathbbm{1}(i^\star \in \{m_1L_1+1,...,m_1L_1+L_1\}) \times \frac{1}{L_1}\right)=\frac{n-1}{n},
\end{align*}
here, (b) follows from the fact that our TS algorithm at $t=1$ will pick all sensing actions with the nonzero blocks $m_1=1$ through $m_1=M_1$ equally likely, and (c) follows from \fref{eq:finiteSingleReg}. 
For $t>2$, there are two policies depending on whether $i^\star$ has been discovered to belong to any of the elements of block $m_1$ or not. If $i^\star$ does not belong to block $m_1$, then:
\begin{align*}
\Delta_2(\pi_2) &= \sum_{\substack{\bmx \in \setX\\\bmx \notin \bD_{t-1}}} \pi(\bmx) \left(1- \hspace{-0.65cm}\sum_{\substack{i^\star=1\\i^\star \notin \{m_1L_1+1,...,m_1L_1+L_1\}}}^{n} \hspace{-1cm} \frac{1}{n-L_1} \times \mathbbm{1}(i^\star \in \{m_2L_2+1,...,m_2L_2+L_2\}) \times \frac{1}{L_2}\right)\\
&=1-\frac{1}{n-L_1}.
\end{align*}
Or if $i^\star$ does belong to block $m_1$, similarly we 
$
\Delta_2(\pi_2) 
=1-\frac{1}{L_1}.
$
For block length $L_1=n/2$, we will have the regret $\Delta_2(\pi_2) =1-\frac{1}{n/2}$ for both policies which is the smallest regret value. Similarly, for any $t>2$, we have:
\begin{align}\label{eq:deltaBlock}
\Delta_t(\pi_t) =1-\frac{1}{n-\sum_{t^\prime=1}^{t-1}\frac{n}{2^{t^\prime}}},
\end{align}
as long as $L_t = \frac{n}{2^t}$. 

We will next compute the information gain in \fref{eq:finiteInfoGain} for time step $t=1$. For that, we will first compute the three probability densities in \fref{eq:infoProbabilities}. For $p(\bmx ^\star)$, we have:
\begin{align*}
p(\bmx ^\star) =\sum_{\bm\beta \in \setB(\bmx^\star)} p(\bm\beta=\bmx^\star) = \left\{
\begin{array}{@{}ll@{}}
\frac{1}{n}, \quad & \text{if} \,\,\,    \bmx^\star \text{ is 1-sparse}\\
0,& \text{otherwise}
\end{array}\right.
\end{align*}
which follows the fact that the optimal action is 1-sparse. Second, for $p(y|\bmx)$ we have:
\begin{align*}
p(y|\bmx) &= p(y|m_1) =\left\{
\begin{array}{@{}ll@{}}
\sum_{i^\star=1}^{n} \frac{1}{n} \,\,\, p(i^\star \notin \{m_1L_1+1,...,m_1L_1+L_1\}), \quad & \text{if } \, y=0 ,\\
\sum_{i^\star=1}^{n} \frac{1}{n}  \,\,\, p(i^\star \in \{m_1L_1+1,...,m_1L_1+L_1\}), & \text{if } \, y=1/{L_1} ,
\end{array}\right.\\
&=\left\{\begin{array}{@{}lll@{}}
\frac{n-L_1}{n}\quad & \text{if } \, y=0,\\
\frac{L_1}{n} & \text{if } \, y=1/L_1,\\
0 & \text{otherwise} ,
\end{array}\right.
\end{align*}
Third, for $p(\bmx^\star,y|\bmx)$ we get:
\begin{align*}
p(\bmx^\star,y|\bmx) = \frac{1}{n} \sum_{\bm\beta =\bmx^\star} p(y|\bmx,\bm\beta)
=\left\{\begin{array}{@{}lll@{}}
\frac{1}{n}\quad & \text{if} \,\,\, y=0 \,\, \land \,\, \bmx ^\Tran \bmx^\star = 0,\\
\frac{1}{n} & \text{if} \,\,\, y=1/L_1 \,\, \land \,\, \bmx ^\Tran \bmx^\star \ne 0,\\
0 & \text{otherwise} .
\end{array}\right.
\end{align*}
Using these three probability densities, the information gain in \fref{eq:finiteInfoGain} becomes:
\begin{align*}
\setI_1(\bmx^\star;y|\bmx) &=  \sum_{\substack{\bmx^\star \in \setX\\\bmx^\Tran \bmx^\star \ne 0,y=1/L_1}} \frac{1}{n} \log \left(\frac{\frac{1}{n}}{\frac{1}{n} \times \frac{L_1}{n} }\right)+\sum_{\substack{\bmx^\star \in \setX\\\bmx^\Tran \bmx^\star =0,y=0}} \frac{1}{n} \log \left(\frac{\frac{1}{n}}{\frac{1}{n} \times \frac{n-L_1}{n} }\right) \\
&= \frac{L_1}{n}\log \left(\frac{n}{L_1}\right) + \frac{n-L_1}{n}\log \left(\frac{n}{n-L_1}\right).
\end{align*}
The information term above is maximized when $L_1=n/2$ which gives $\setI_1(\bmx^\star;y|\bmx)=\log \left(2\right)$.

Next, let us discuss information gain for $t=2$. Recall that for $t=2$, there are two policies depending on whether $i^\star$ has been discovered to belong to elements of block $m_1$ or not. 
Similar computations to that of $t=1$ shows that for the two policies we have:
 
\begin{align*}
\setI_2(\bmx^\star;y|\bmx) 
=\left\{\begin{array}{@{}ll@{}}
 \frac{L_2}{L_1}\log \left(\frac{L_1}{L_2}\right) + \frac{L_1-L_2}{L_1}\log \left(\frac{L_1}{L_1-L_2}\right),& \text{ if  \resizebox{.25\hsize}{!}{$i^\star \in \{m_1L_1+1,...,m_1L_1+L_1\} $}}\\
\frac{L_2}{n-L_1}\log \left(\frac{n-L_1}{L_2}\right) + \frac{n-L_1-L_2}{n-L_1}\log \left(\frac{n-L_1}{n-L_1-L_2}\right),& \text{ if  \resizebox{.25\hsize}{!}{$i^\star \notin \{m_1L_1+1,...,m_1L_1+L_1\} $}}
\end{array}\right.
\end{align*}
By setting $L_1 = n/2$ and $L_2 = n/4$, both policies will be maximized to give $\setI_2(\bmx^\star;y|\bmx)  = \log \left(2\right)$. Similarly, for any $t>2$, the information gain $	\setI_t(\bmx^\star;y|\bmx) $ is maximized with:
\begin{align}\label{eq:infoBlock}
	\setI_t(\bmx^\star;y|\bmx)  = \log \left(2\right),
\end{align}
as long as $L_t = \frac{n}{2^t}$. The fact that the varying block length $L_t = \frac{n}{2^t}$ maximizes information gain $\setI_t(\bmx^\star;y|\bmx) $ and minimizes regret term $\Delta_t(\pi_t)$ shows that this varying block length is the optimal approach for our TS algorithm finding a 1-sparse signal of interest. This result in essence describes our strategy for varying block length $L_t = \frac{L_{t-1}}{2}$in SPATS algorithm.

Finally, using \fref{eq:deltaBlock} and \fref{eq:infoBlock} with the upperbound $\text{H}(\bmx^\star) < \log\left(|\setX|\right)$, we can compute the expected regret in  \fref{eq:infoRegretBound} as:
\begin{align*}
\Ex{}{\text{Reg}(T)} \le \left(\log\left(|\setX|\right)  \sum_{t=1}^{T} \left[ \frac{\left( 1-\frac{1}{n-\sum_{t^\prime=1}^{t-1}\frac{n}{2^{t^\prime}}} \right)^2}{\log\left(2\right)}\right] \right)^{1/2},
\end{align*}
Lastly, the regret bound we computed above is only applicable when $T\le \log_2(n)$. Since the algorithm has a varying block length of $L_t = \frac{n}{2^t}$, after $T=\log_2(n)$ the block length is reduced to $1$ which certainly includes nonzero element of $\bm\beta$.
As a result, after $T=\log_2(n)$ actions the algorithm has surely found the optimal action. This result means that $\Delta_t(\pi_t)=0$ for $t > \log_2(n)$ which gives us the regret bound:
\begin{align}\label{eq:regret_block_app}
\textstyle \Ex{}{\text{Reg}(T)} \leq \left( \log \left(|\setX|\right) \sum_{t=1}^{\min\{T,\log_2(n)\}}  {\Big(1-\frac{1}{n-\left(\sum_{t^\prime=1}^{t-1}\frac{n}{2^{t^\prime}}\right)}\!\Big)^2}/{\log(2)} \right)^{1/2}.
\end{align}

\begin{rem}
	We noticed a minor mistake in expected regret term \fref{eq:regret_block} in \fref{thm:singleAgent} after paper submission. Equation \fref{eq:regret_block_app} above describes the corrected term. This does not affect any of the conclusions in the main paper.
\end{rem}

\section{Proof for \fref{thm:multiAgent}}\label{app:thmMultiAgent}
Unlike \fref{thm:singleAgent}, we use a direct approach to compute expected regret in this proof. 
Similar to \fref{app:thmSingleAgent}, to simplify the discussion, we assume both algorithms are aware of the prior distribution on true parameter $\bm\beta$.
 We will first compute the expected regret for the single agent algorithm.
\subsection*{Single Agent}

Since the true parameter $\bm\beta$ is 1-sparse, we assume the location of the non-zero element is $i^\star$. 
Recall the equation for expected regret:
\begin{align}\label{eq:reg1_single}
 \Ex{}{\text{Reg(T)}} \!=\!\Ex{i^\star,\bmx_1,...,\bmx_T}{\sum_{t=1}^{T} \left[\setR(\bmx^\star,\bm\beta) \!-\! \setR(\bmx_t,\bm\beta)\right]}\!\!\stackrel{(a)}{=} \!\sum_{t=1}^{T} \setR(\bmx^\star,\bm\beta) \!-\!\sum_{t=1}^{T}  \Ex{\bmx_t}{\setR(\bmx_t,\bm\beta)} ,
 \end{align}
  where, (a) follows from assuming $i^\star$ is a fixed unknown location without loss of generality. To compute the first term $\setR(\bmx^\star,\bm\beta) $, note that the optimal action $\bmx^\star$ is the 1-sparse vector with $i^\star$ as the nonzero element, i.e. $\setR(\bmx^\star,\bm\beta)=1$. 
  
  To compute the second term $\Ex{\bmx_t}{\setR(\bmx_t,\bm\beta)} $, remember that $\bmx_t$ for TS is selected by maximizing the reward function given the available measurements $\bD_{t-1}$. For a single agent setting, as depicted in \fref{fig:single}, we have $\bD_{t-1}=\{(\bmx_j,y_j)|j=1,..,t\!-\!1\}$. Since our algorithm has a 1-sparse prior, only feasible sensing actions are 1-sparse. If $\bmx^\star \in \bD_{t-1}$, then TS will be exploiting the same optimal action at time $t$ and set $\bmx_t = \bmx^\star$ for which $\Ex{\bmx_t}{\setR(\bmx_t,\bm\beta)|\bmx^\star \in \bD_{t-1}} =\Ex{\bmx_t}{\setR(\bmx_t,\bm\beta)|\bmx_t=\bmx^\star}=1$. On the other hand, if $\bmx^\star \notin \bD_{t-1}$, then TS will select action $\bmx_t$ out of $n-t+1$ actions that are not observed as part of $\bD_{t-1}$. Hence, the reward $\setR(\bmx_t,\bm\beta)$ is $1$ with probability $\frac{1}{n-t+1}$ and $0$ otherwise, i.e $\Ex{\bmx_t}{\setR(\bmx_t,\bm\beta)|\bmx^\star \notin \bD_{t-1}} =\frac{1}{n-t+1}$.
  
  Putting the information for the first and second term in \fref{eq:reg1_single}, we have:
\begin{align*}
	\Ex{}{\text{Reg(T)}} &=\sum_{t=1}^{T} \setR(\bmx^\star,\bm\beta) -\sum_{t=1}^{T}  \Ex{\bmx_t}{\setR(\bmx_t,\bm\beta)} \\
	&= T-\sum_{t=1}^{T}  \Big[p(\bmx^\star \in \bD_{t-1}) \times \Ex{\bmx_t}{\setR(\bmx_t,\bm\beta)|\bmx^\star \in \bD_{t-1}} \\
	&\hspace{15mm}+ p(\bmx^\star \notin \bD_{t-1}) \times \Ex{\bmx_t}{\setR(\bmx_t,\bm\beta)|\bmx^\star \notin \bD_{t-1}} \Big]\\
	&\stackrel{(b)}{=} T-\sum_{t=1}^{T}  \Big[\frac{t-1}{n} \times 1 + \frac{n-t+1}{n}\times \frac{1}{n-t+1} \Big]\\
	&= T - \frac{T(T+1)}{2n},
\end{align*}
where, (b) follows from $p(\bmx^\star \in \bD_{t-1}) = 1-  p(\bmx^\star \notin \bD_{t-1}) = \frac{t-1}{n}$ given that there are $t-1$ distinct measurements out of $n$ distinct possible actions until $\bmx^\star$ is found.

Lastly, the regret term we computed above is only applicable when $T\le n$. Since there are $n$ 1-sparse actions available for a vector $\bm\beta$ with length $n$, after $T=n$ actions the algorithm has surely found the optimal action. As a result, $\setR(\bmx^\star,\bm\beta) - \setR(\bmx_t,\bm\beta)=1-1=0$ for $t>n$ which gives us:
\begin{align*}
 \Ex{}{\text{Reg}(T)} = T_n - \frac{T_n(T_n+1)}{2n}, \quad T_n = \min\{T,n\}.
\end{align*}

\subsection*{Multi Agent}

Now, we will compute the expected regret for TS with asynchronous multi agent setting. Similar to single agent, the equation for expected regret follows from \fref{eq:reg1_single} with $i^\star$ as the location of the nonzero element of $\bm\beta$.
Again, for the first term we have $\setR(\bmx^\star,\bm\beta)=1$ where the optimal action $\bmx^\star$ is the 1-sparse vector with $i^\star$ as the nonzero element.
For the second term $\Ex{\bmx_t}{\setR(\bmx_t,\bm\beta)}$, we know that if $\bmx^\star \in \bD_{t-1}$, then TS will be exploiting the same optimal action at time $t$ and set $\bmx_t = \bmx^\star$ for which $\Ex{\bmx_t}{\setR(\bmx_t,\bm\beta)|\bmx^\star \in \bD_{t-1}} =1$. On the other hand, if $\bmx^\star \notin \bD_{t-1}$, then TS will select action $\bmx_t$ out of actions that are not observed as part of $\bD_{t-1}$. Precisely:
\begin{align}\nonumber
\Ex{}{\text{Reg(T)}} 
&= T-\sum_{t=1}^{T}  \Big[p(\bmx^\star \in \bD_{t-1}) \times \Ex{\bmx_t}{\setR(\bmx_t,\bm\beta)|\bmx^\star \in \bD_{t-1}} \\\label{eq:regret1_multi}
&\hspace{15mm}+ p(\bmx^\star \notin \bD_{t-1}) \times \Ex{\bmx_t}{\setR(\bmx_t,\bm\beta)|\bmx^\star \notin \bD_{t-1}} \Big].
\end{align}
However, computing $p(\bmx^\star \in \bD_{t-1})=1-p(\bmx^\star \notin \bD_{t-1})$ and $\Ex{\bmx_t}{\setR(\bmx_t,\bm\beta)|\bmx^\star \notin \bD_{t-1}}$ is not as straight forward as the single agent setting. In a single agent setting, $|\bD_{t-1}| = t-1$ and all $t-1$ actions are distinct until $\bmx^\star$ is found. In an asynchronous multi agent setting, $|\bD_{t-1}| < t-1$ and some of the actions in $\bD_{t-1}$ might be the same. Specifically, as illustrated in \fref{fig:multi} an action $t$ starts before all previous $t-1$ actions are completed and as a result there is a chance that action $t$ will equal the previous actions that were not included in $\bD_{t-1}$.
%
While we cannot precisely compute $p(\bmx^\star \in \bD_{t-1})$ in this asynchronous multi agent setting, in the following lemma we compute a lower bound for it with the proof provided in \fref{app:lowerbound}.

\begin{lem}\label{lem:lowerbound}
	Consider the active search problem in \fref{thm:multiAgent}. For an asynchronous multi agent TS algorithm, $p(\bmx^\star \in \bD_{t-1})$ is bounded by:
	\begin{align*}
	p(\bmx^\star \in \bD_{t-1}) &\ge 1 - \frac{(n-1)^{t}}{n^t}, \quad \text{if} \,\,\,\, t < g\\
p(\bmx^\star \in \bD_{t-1}) &\ge 1 - \frac{(n-1)^{g-1}(n-t+2g-1)}{n^g}, \quad \text{if} \,\,\,\, t \ge g
	\end{align*}
\end{lem}
Using the bound in \fref{lem:lowerbound} along with a na\"ive bound  $\Ex{\bmx_t}{\setR(\bmx_t,\bm\beta)|\bmx^\star \notin \bD_{t-1}} \ge 0$, we can upper bound the expected regret in \fref{eq:regret1_multi} as follows:
\begin{align*}
\Ex{}{\text{Reg(T)}} 
&\le T-\sum_{t=1}^{T}  \Big[p(\bmx^\star \in \bD_{t-1}) \times \Ex{\bmx_t}{\setR(\bmx_t,\bm\beta)|\bmx^\star \in \bD_{t-1}} \Big]\\
&= T - \sum_{t=1}^{g}  \Big[1 - \frac{(n-1)^{t}}{n^t} \times 1 \Big] - \sum_{t=g+1}^{T}  \Big[1 - \frac{(n-1)^{g-1}(n-t+2g-1)}{n^g} \times 1 \Big]\\
&\le T -g + \sum_{t=1}^{g} \Big[\frac{(n)^{t}}{n^t} \Big] - (T-g) +\sum_{t=g+1}^{T}  \Big[ \frac{(n)^{g-1}(n-t+2g-1)}{n^g} \Big]\\
&=  T  - \frac{T(T+1)}{2n} + \frac{T(2g-1)}{n} -\frac{3g^2-3g}{2n} \\
&\le  T  - \frac{T(T+1)}{2n} + \frac{T(2g-1)}{n}. 
\end{align*}
Lastly, the regret term we computed above is only applicable when $T\le n+g$. If $T> n+g$, then $|\bD_{t-1}|=n$ which includes all $n$ 1-sparse actions available for a vector $\bm\beta$ with length $n$. Hence, after $T=n+g$ actions, the algorithm has surely found the optimal action which results in $\setR(\bmx^\star,\bm\beta) - \setR(\bmx_t,\bm\beta)=1-1=0$, and consequently:
\begin{align*}
\Ex{}{\text{Reg}(T)} \leq  T_n - \frac{T_n(T_n+1)}{2n} \!+\!\frac{T_n(2g-1)}{n}, \quad T_n \!= \min\{T,n\!+\!g\} 
\end{align*}

\subsection{Proof for \fref{lem:lowerbound}}\label{app:lowerbound}
\begin{proof}


Since we need to upper bound the expected regret in \fref{eq:reg1_single}, we will lower bound $p(\bmx^\star \in \bD_{t-1})$. To compute this bound, we need to better understand $\bD_{t-1}$. For that, we will look at the example illustrated in \fref{fig:multi}. For this example, $\bD_{0}$ through $\bD_{11}$ have the following measurement sets:
\begin{align*}
\bD_{0} &= \emptyset \hspace{4.05cm} \bD_{6} = \{(\bmx_j,y_j)|j=\{1,3,2,5\}\}\\
\bD_{1} &= \emptyset \hspace{4.05cm} \bD_{7} = \{(\bmx_j,y_j)|j=\{1,3,2,5,6\}\}\\
\bD_{2} &= \emptyset \hspace{4.05cm} \bD_{8} = \{(\bmx_j,y_j)|j=\{1,3,2,5,6,4\}\}\\
\bD_{3} &= \{(\bmx_j,y_j)|j=\{1\}\} \hspace{1.51cm} \bD_{9} = \{(\bmx_j,y_j)|j=\{1,3,2,5,6,4,8\}\}\\
\bD_{4} &= \{(\bmx_j,y_j)|j=\{1,3\}\} \hspace{1.04cm} \bD_{10} = \{(\bmx_j,y_j)|j=\{1,3,2,5,6,4,8,7\}\}\\
\bD_{5} &= \{(\bmx_j,y_j)|j=\{1,3,2\}\} \quad \quad \bD_{11} = \{(\bmx_j,y_j)|j=\{1,3,2,5,6,4,8,7,9\}\}
\end{align*}
Our first observation is that $|\bD_{t-1}|=t-g$ which applies to any asynchronous example \cite{kandasamy2018parallelised}. Next, let us compute $p(\bmx^\star \in \bD_{t-1})$ as an example for $\bD_{6}$:
\begin{align}\nonumber
p(\bmx^\star \in \bD_{6}) &= 1 - p(\bmx^\star \notin \bD_{6}) = 1 - p\Big((\bmx^\star \neq \bmx_1) \land (\bmx^\star \neq \bmx_3) \land (\bmx^\star \neq \bmx_2) \land (\bmx^\star \neq \bmx_5)\Big)\\\nonumber
&= 1 - \Big(p(\bmx^\star \neq \bmx_1) \times p(\bmx^\star \neq \bmx_3|\bmx^\star \neq \bmx_1) \times p(\bmx^\star \neq \bmx_2| \bmx^\star \neq \bmx_1 , \bmx_3) \\\nonumber
&\hspace*{1cm}\times p(\bmx^\star \neq \bmx_5|\bmx^\star \neq \bmx_1, \bmx_3, \bmx_2)\Big)\\\nonumber
&\stackrel{(c)}{=} 1 - \Big(p(\bmx^\star \neq \bmx_1) \times p(\bmx^\star \neq \bmx_3|\emptyset) \times p(\bmx^\star \neq \bmx_2|\emptyset) \times p(\bmx^\star \neq \bmx_5|\bmx^\star \neq \bmx_1, \bmx_3)\Big)\\\label{eq:pxinD6}
&\stackrel{(d)}{=} 1 - \frac{n-1}{n} \times \frac{n-1}{n} \times \frac{n-1}{n} \times \frac{n-3}{n-2}.
\end{align}
Here, (c) follows from the fact that on picking $\bmx_3$ and $\bmx_2$, they depend on empty sets of measurements $\bD_{2}$ and $\bD_{1}$. And, on choosing sensing action $\bmx_5$, the algorithm has access to measurements $\bD_{4}= \{(\bmx_j,y_j)|j=\{1,3\}\}$. Consequently, in (d) we see that $\bmx_1$,$\bmx_3$ and $\bmx_2$ have $n$ possible actions to pick, while $\bmx_5$ will pick out of $n-2$ actions that had excluded the ones in $\bD_{4}$.

As evident in \fref{eq:pxinD6}, computing $p(\bmx^\star \in \bD_{t-1})$ depends on the specifics of every example and is different for all $\bD_{t-1}$. However, this probability can be lower-bounded for any asynchronous setting as follows.
We know that $\bD_{t-1}$ has a cardinality of $t-g$. As a result, each asynchronous example translates to a set $\bD_{t-1}$ with permutation of $t-g$ measurements picked out of all $t-1$ possible measurements $\{(\bmx_j,y_j)|\{j=1,...,t-1\}\}$. Out of all possible asynchronous examples, $\bD^W_{t-1}=\{(\bmx_j,y_j)|\{j=1,...,t-g\}\}$ is the worst case scenario in terms of finding the optimal action. In other words, for this measurement set, the algorithm has used the least amount of information to decide on each of the actions $\bmx_1$ through $\bmx_t$. Hence, the asynchronous example with $\bD^W_{t-1}$ has the lowest probability to pick the optimal sensing action $\bmx^\star$ and can be used to lower-bound $p(\bmx^\star \in \bD_{t-1})$:
\begin{align*}
p(\bmx^\star \in \bD_{t-1}) &\ge p(\bmx^\star \in \bD^W_{t-1})  \\
\end{align*}
Now, all we have to do is compute $p(\bmx^\star \in \bD^W_{t-1})$. For $t<g$, we have:
\begin{align*}
p(\bmx^\star \in \bD^W_{t-1}) &= 1 - p(\bmx^\star \notin \bD^W_{t-1})=1 - \Big(p(\bmx^\star \neq \bmx_1) \times p(\bmx^\star \neq \bmx_2|\emptyset) \times ... \times p(\bmx^\star \neq \bmx_t|\emptyset)\Big)\\
&= 1 - \underbrace{\frac{n-1}{n} \times \frac{n-1}{n} \times ... \times \frac{n-1}{n}}_{t \,\,\,\, \text{terms}} = 1 - \frac{(n-1)^t}{n^t}
\end{align*}
For $t\ge g$, we have:
\begin{align*}
p(\bmx^\star \in \bD^W_{t-1}) &= 1 - p(\bmx^\star \notin \bD^W_{t-1})\\
&=1 - \Big(p(\bmx^\star \neq \bmx_1) \times p(\bmx^\star \neq \bmx_2|\emptyset) \times ... \times p(\bmx^\star \neq \bmx_g|\emptyset)\\
&\hspace*{1cm} \times p(\bmx^\star \neq \bmx_{g+1}|\bmx^\star \neq \bmx_1) p(\bmx^\star \neq \bmx_{g+2}|\bmx^\star \neq \bmx_1,\bmx_2) \\
&\hspace*{1cm}\times ... \times p(\bmx^\star \neq \bmx_{t-g}|\bmx^\star \neq \bmx_1,\bmx_2,...,\bmx_{t-2g})\Big)\\
&= 1 - \underbrace{\frac{n-1}{n} \times \frac{n-1}{n} \times ... \times \frac{n-1}{n}}_{g \,\,\,\, \text{terms}} \times \frac{n-2}{n-1} \times \frac{n-3}{n-2} \times ... \times \frac{n-(t-2g)-1}{n-(t-2g)}\\
&= 1 - \frac{(n-1)^{g-1}(n-t+2g-1)}{n^g}
\end{align*}
\end{proof}


\section{Additional Details on LATSI}\label{app:LATSI}

In this section, we provide additional details on deriving LATSI algorithm we proposed in \fref{sec:LATSI}. First, let us do a short review on IDS and RSI algorithms:
\subsection*{Review of IDS (Information Directed Sampling) proposed by \cite{russo2017learning}}
There are certain online optimization problems such as our active search problem where traditional multi-armed bandit algorithms such as TS and Upper Confidence Bound (UCB) fail due to a careless assessment of the information gain (see \fref{app:LTSfail}).
IDS is an online optimization algorithm proposed by \cite{russo2017learning} that addresses this problem by introducing a novel reward function that balances between expected single-period regret and a measure of information gain. Specifically, defining $\pi_t^\text{IDS}$ as the action sampling distribution of IDS at time $t$, at each time step $t$, IDS is computed by:
\begin{align*}
\pi_t^\text{IDS} = \argmin_{\pi \in \setD(\setX)} \left\{\Psi_t(\pi) \triangleq \frac{\Delta_t(\pi)^2}{\setI_t(\pi)}\right\},
\end{align*}
where they call $\Psi_t(\pi)$ the information ratio of the action sampling distribution $\pi$, and $\Delta_t(\pi)$ and $\setI_t(\pi)$ are the single-period expected regret and information gain at time $t$ defined below. For the single-period expected regret, we have:
\begin{align*}
	\Delta_t(\pi) = \sum_{\bmx \in \setX} \pi(\bmx) \Ex{}{\setR_t(\bmx^\star) - \setR_t(\bmx)},
\end{align*}
where, $\setR_t(\bmx)$ is the reward function for action $\bmx$ at time $t$.
And, for information gain, we have:
\begin{align*}
	\setI_t(\pi) = \sum_{\bmx \in \setX} \pi(\bmx) \bI_t(\bmx^\star;y|\bmx),
\end{align*}
where, $\setI_t(\bmx^\star;y|\bmx)$ is the mutual information between optimal action $\bmx^\star$ and observations $y$ at time $t$ if sensing action $\bmx$ is chosen.

Once IDS computes the action sampling distribution at time $t$, it will choose an action by randomly sampling this distribution.

\subsection*{Review of RSI (Region Sensing Index) proposed by \cite{ma2017active}}
RSI is a single agent active search algorithm designed to locate sparse signals by actively making data-collection decisions. Similar to our problem formulation in \fref{sec:formulation}, RSI makes a practical assumption that at each time step the agent senses a contiguous region of the space. To decide on their next action, RSI at each time step chooses the sensing action $\bmx_t$ that maximizes the mutual information between the next observation $y_t$ and the signal of interest $\bm\beta$, i.e.
\begin{align*}
	\bmx_t = \argmax_{\bmx} \mathit{I}(\bm\beta;y|\bmx,\bD_{t-1}),
\end{align*}
where, the mutual information is computed using posterior distribution $p(\bm\beta|\bD_{t-1})= p_0(\bm\beta) \prod_{\tau=1}^{t-1} p(y_\tau|\bmx,\bm\beta)$ with a k-sparse uniform prior $p_0(\bm\beta)$ and same likelihood distribution as in \fref{eq:formulation}. 

Unfortunately, computing the mutual information $\mathit{I}(\bm\beta;y|\bmx,\bD_{t-1})$ has high complexity for sparsity rates of $k>1$.  In order to reduce this complexity for $k>1$ RSI recovers the support of $\bm \beta$ by repeatedly applying RSI assuming $k=1$.

\subsection*{Deriving LATSI}
Recall how IDS algorithm solves for the failure mode of TS by using a novel reward function that is the ratio of the expected single-period regret and information gain. Inspired by this algorithm, we propose revising the expected reward $\lambda^+(\bm\beta^\star,\bD_{t-1},\bmx)$ in \fref{eq:expectedReward} for Laplace-TS by adding a measure of information gain to it. For the information gain, we use the mutual information $\mathit{I}(\bm\beta;y|\bmx,\bD_{t-1})$ computed in RSI algorithm. 
Thus, we can say that LATSI is the Laplace-TS algorithm where the design stage has been replaced by:
\begin{align*}
\bmx_t \!&=\! \argmax_{\bmx} \bm{R}^+(\bm\beta^\star, \bD_{t-1}, \{\bmx,y\}),\\
\end{align*}
where,
\begin{align} \label{eq:LATSIreward}
\bm{R}^+ \!&= \!\frac{I(\bm\beta^\star; y | \bmx, \bD_{t-1})}{\text{average }I\text{ over all }\bmx} \!+\! \alpha \! \times \!
\frac{\lambda^+(\bm\beta^\star,\bD_{t-1},\bmx)}{\text{average }\lambda^+\text{ over all }\bmx}.
\end{align}
Here, we normalize and add a tuning parameter $\alpha$ to best control the importance of each term on the overall reward $\bm{R}^{+}$.


\section{Additional Numerical Results}\label{app:experiments}

We now provide additional simulation results to support our analysis in \fref{sec:results}.
\subsection*{1-dimensional Search Space}
\begin{figure}
	\centering
	\begin{subfigure}{0.245\linewidth}
		\includegraphics[width=\linewidth]{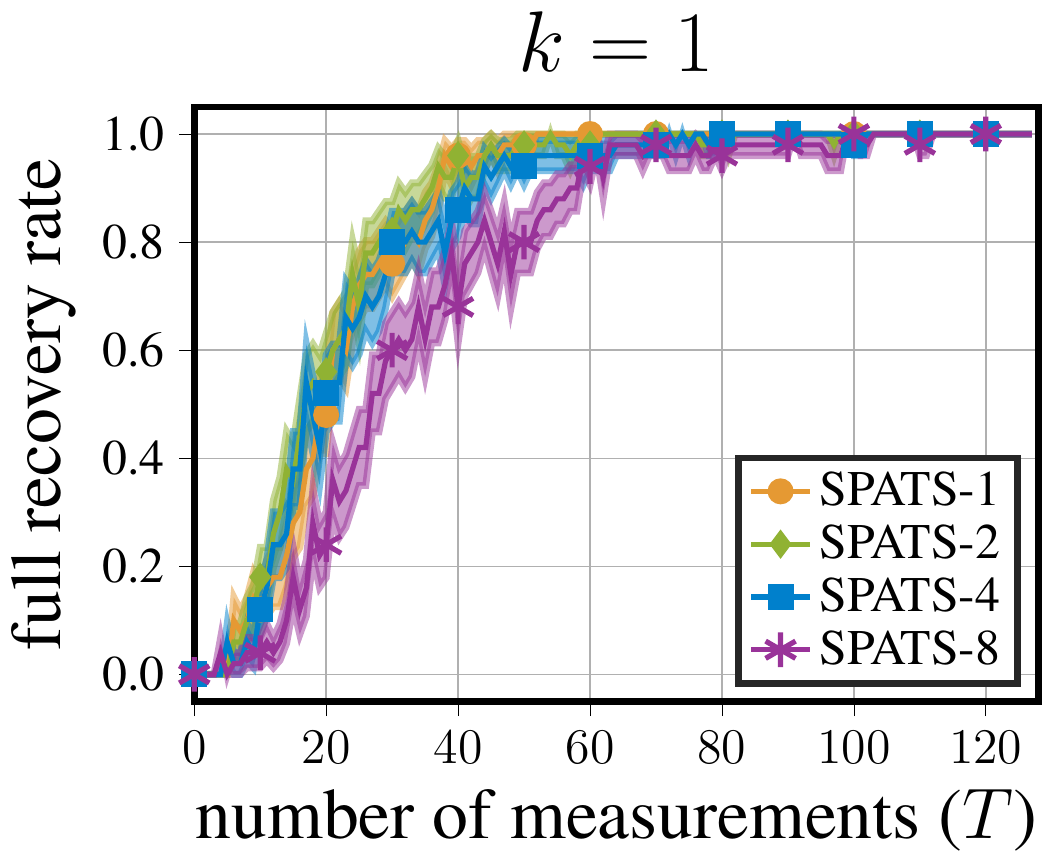}
	\end{subfigure}
	\begin{subfigure}{0.245\linewidth}
		\includegraphics[width=\linewidth]{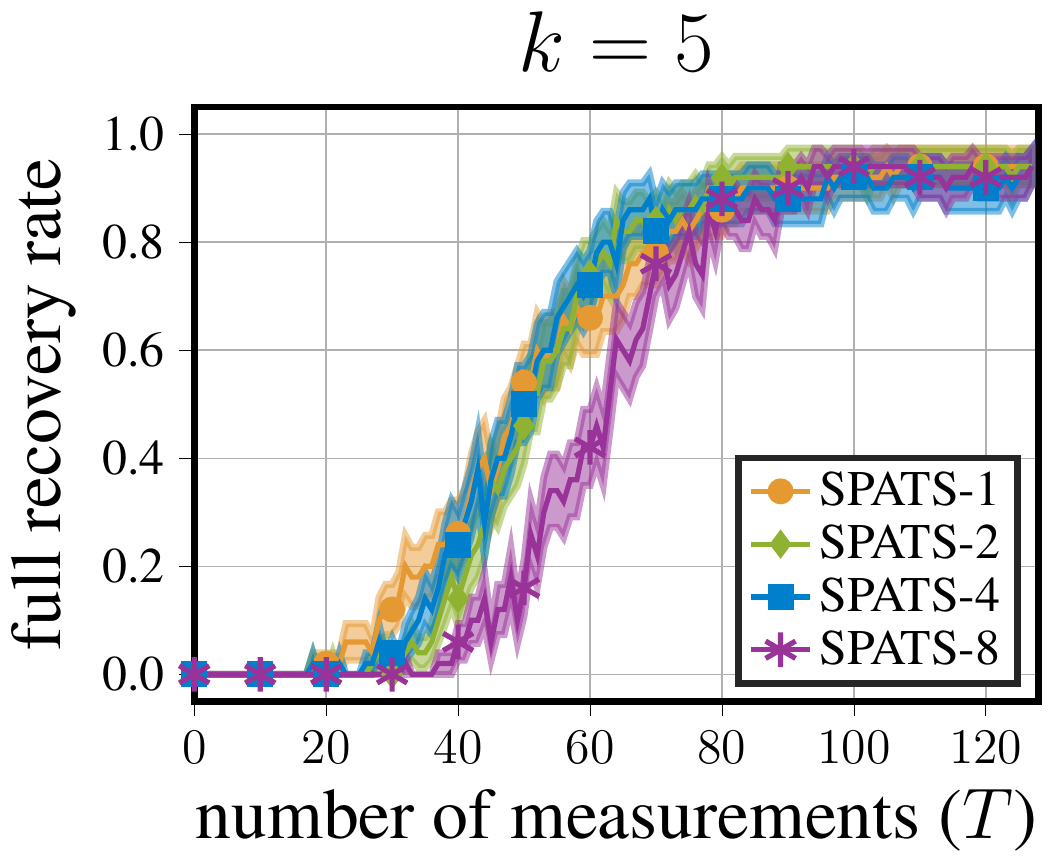}
	\end{subfigure}
	\begin{subfigure}{0.245\linewidth}
		\includegraphics[width=\linewidth]{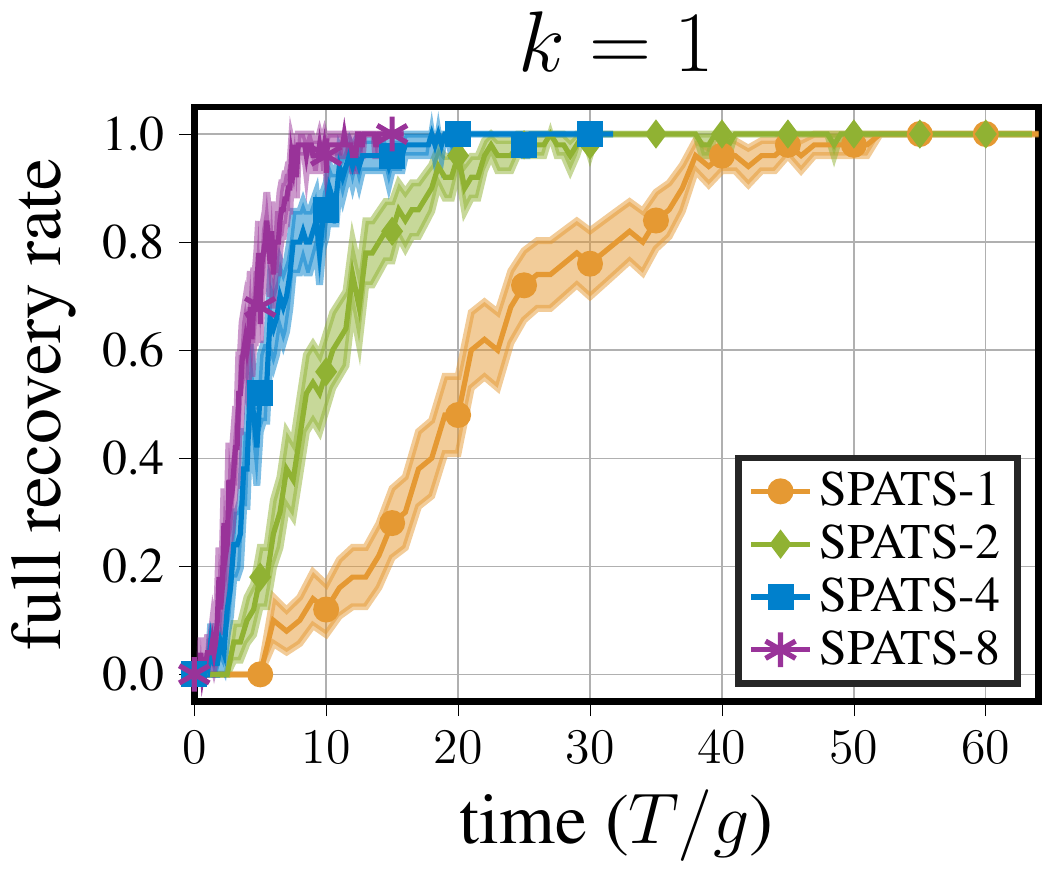}
	\end{subfigure}
	\begin{subfigure}{0.245\linewidth}
		\includegraphics[width=\linewidth]{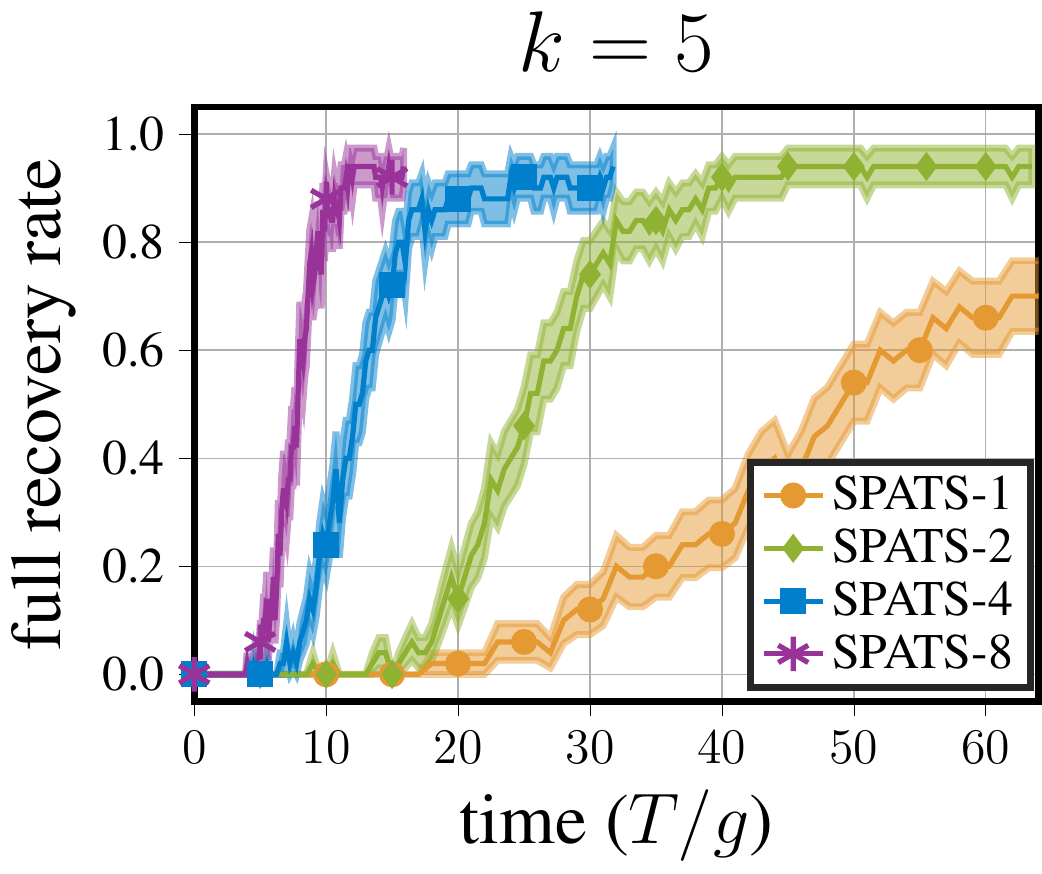}
	\end{subfigure}
	\caption{Full recovery rate of SPATS with 1, 2, 4 and 8 agents for $n = 128$, $k=1, 5$}
	\label{fig:TS_n128}
\end{figure}

\begin{figure}
	\centering
	\begin{subfigure}{0.245\linewidth}
		\includegraphics[width=\linewidth]{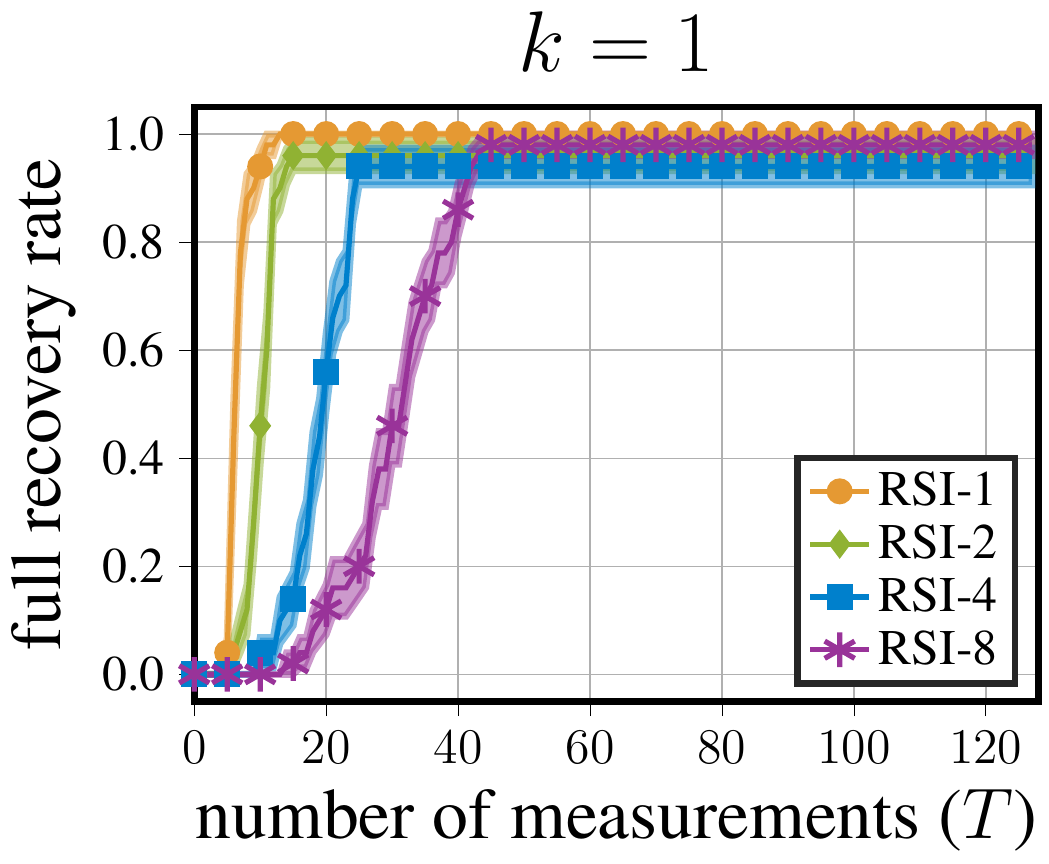}
	\end{subfigure}
	\begin{subfigure}{0.245\linewidth}
		\includegraphics[width=\linewidth]{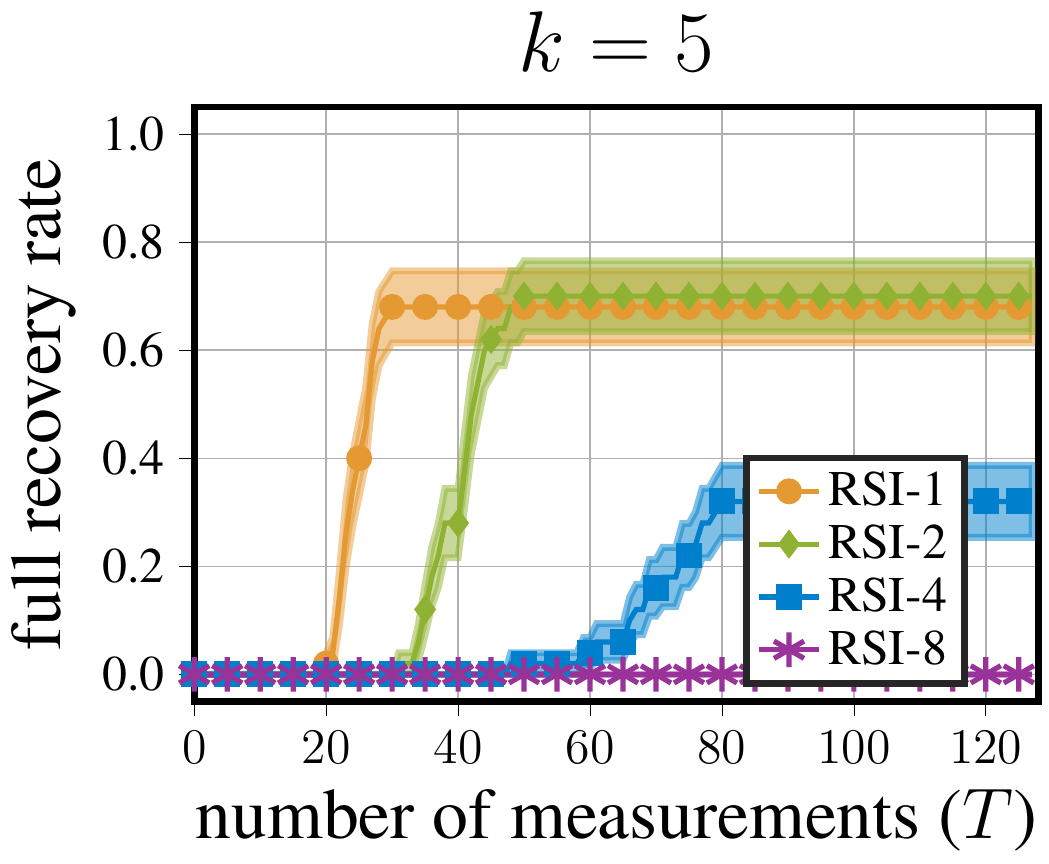}
	\end{subfigure}
	\begin{subfigure}{0.245\linewidth}
		\includegraphics[width=\linewidth]{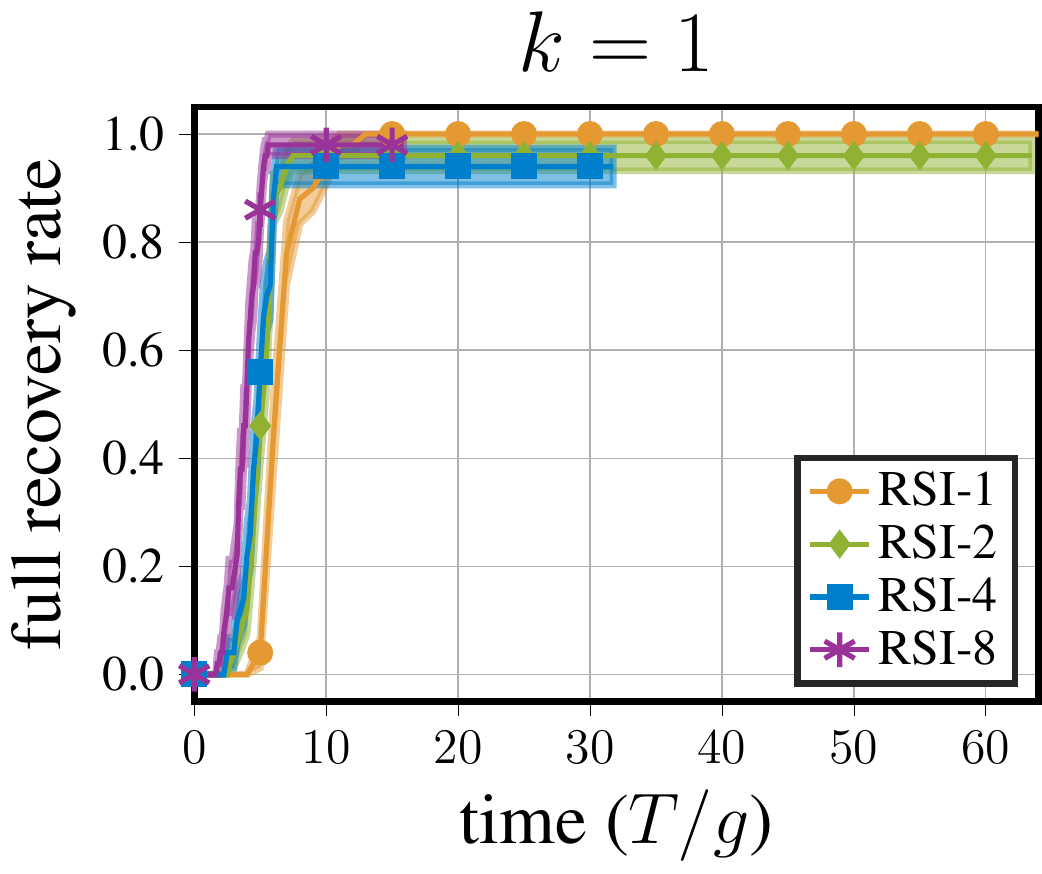}
	\end{subfigure}
	\begin{subfigure}{0.245\linewidth}
		\includegraphics[width=\linewidth]{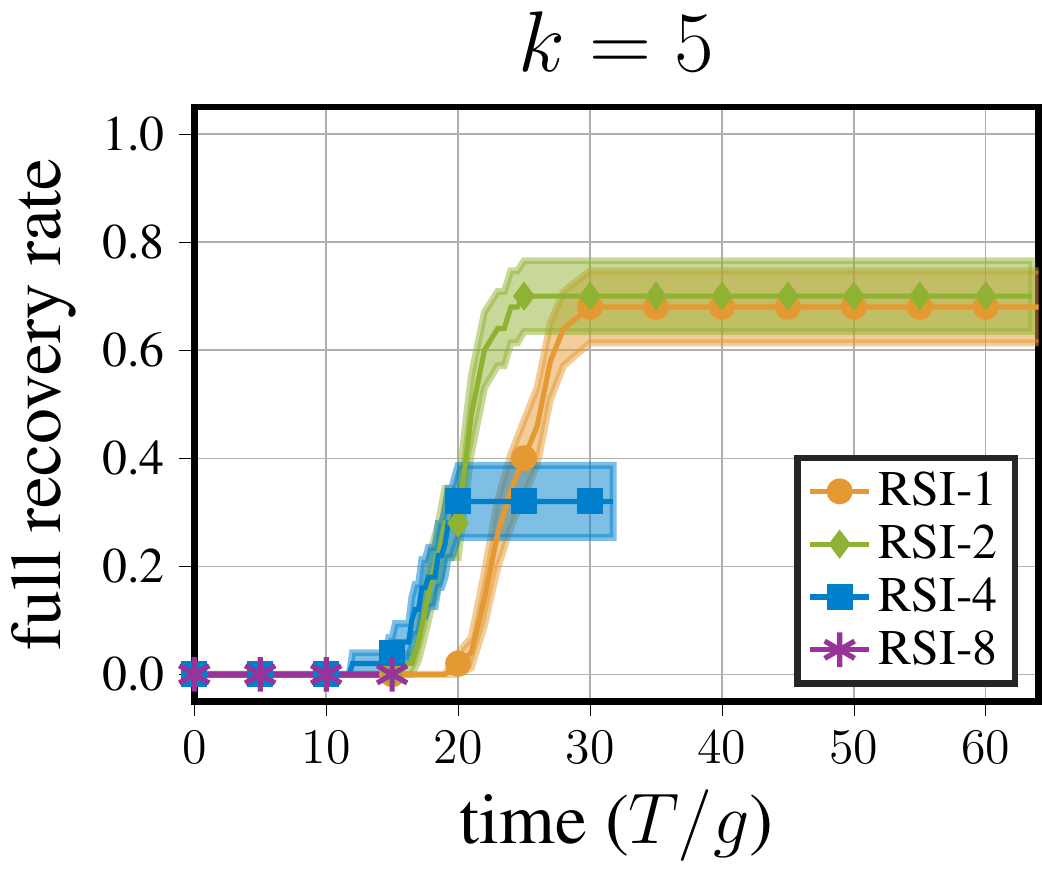}
	\end{subfigure}
	\caption{Full recovery rate of RSI with 1, 2, 4 and 8 agents for $n = 128$, $k=1, 5$}
	\label{fig:RSI_n128}
\end{figure}

\begin{figure}
	\centering
	\begin{subfigure}{0.245\linewidth}
		\includegraphics[width=\linewidth]{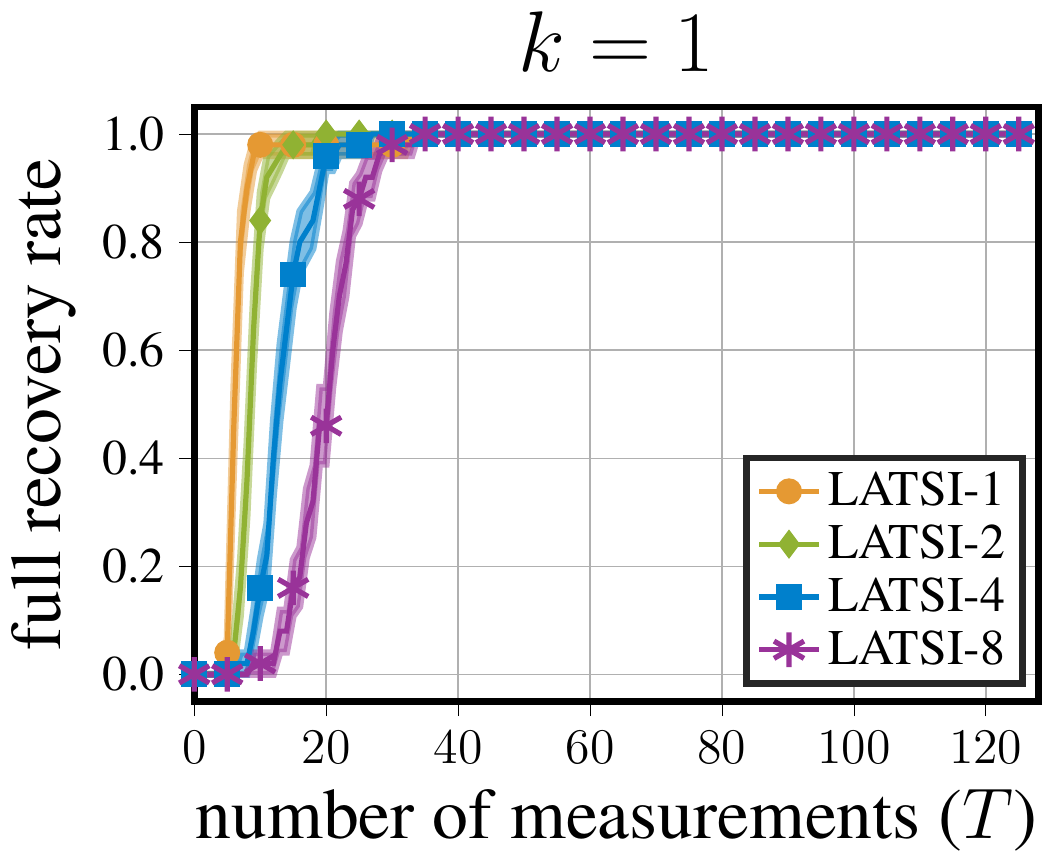}
	\end{subfigure}
	\begin{subfigure}{0.245\linewidth}
		\includegraphics[width=\linewidth]{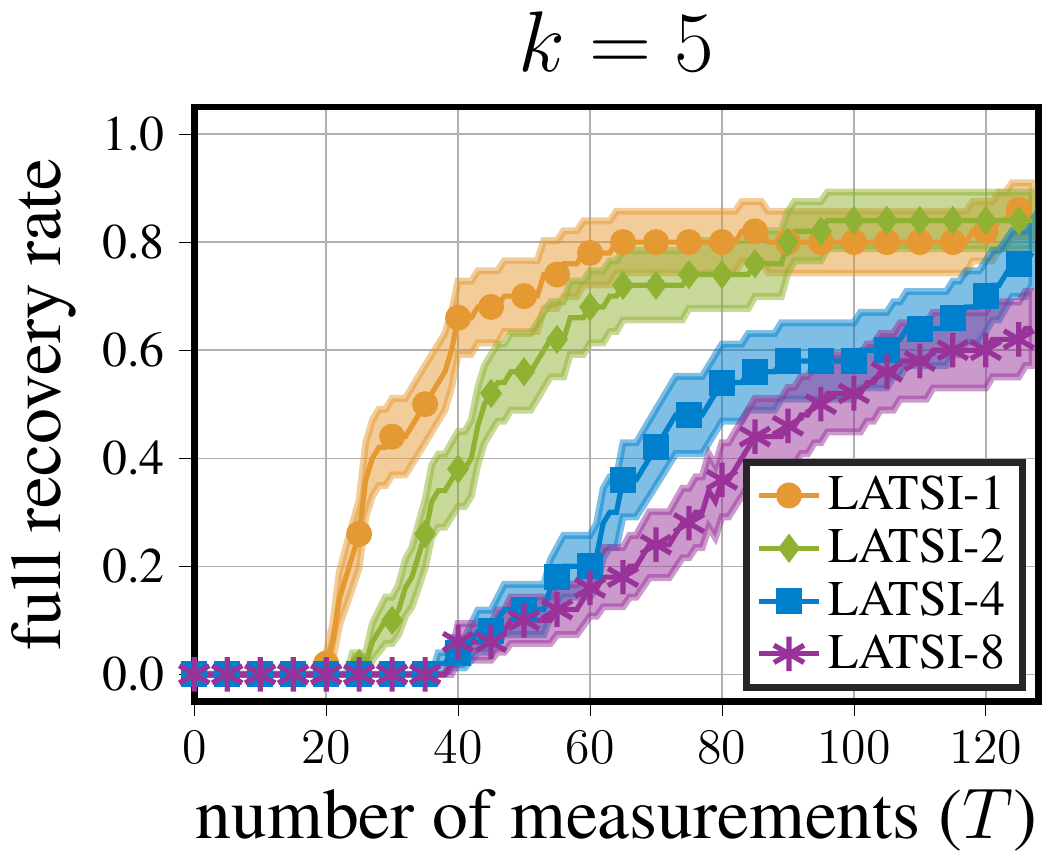}
	\end{subfigure}
	\begin{subfigure}{0.245\linewidth}
		\includegraphics[width=\linewidth]{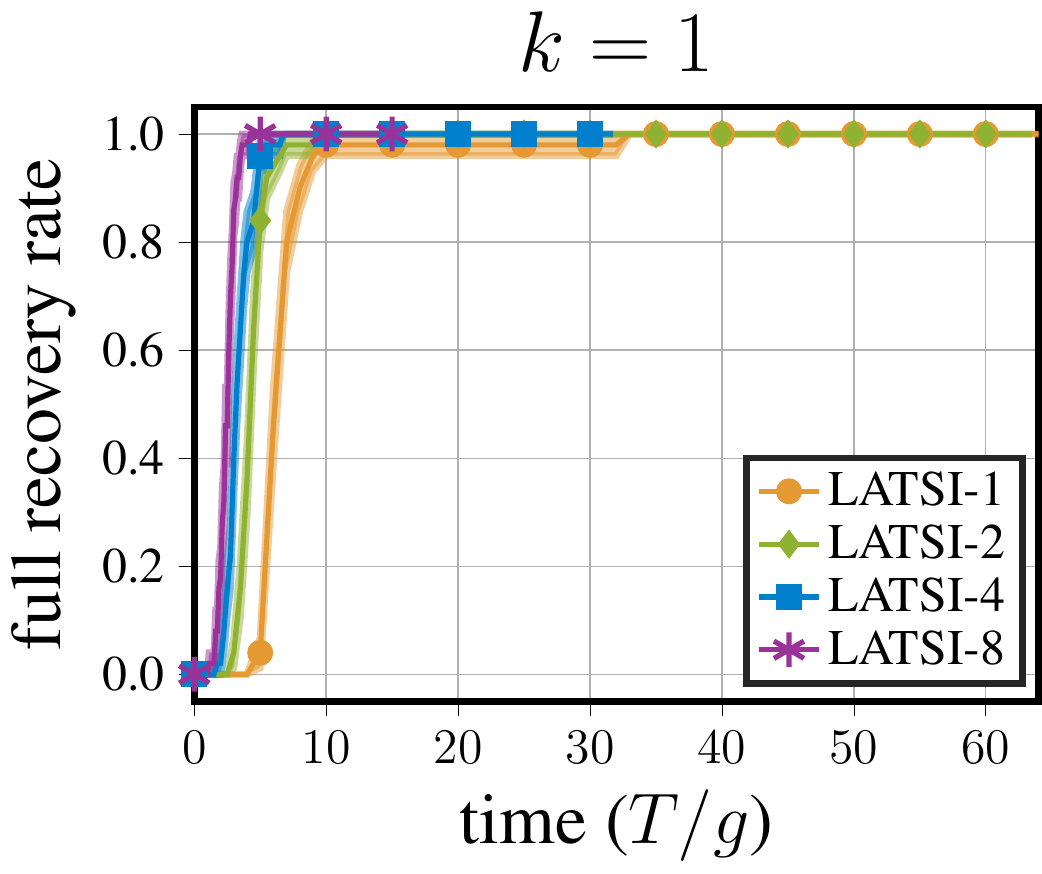}
	\end{subfigure}
	\begin{subfigure}{0.245\linewidth}
		\includegraphics[width=\linewidth]{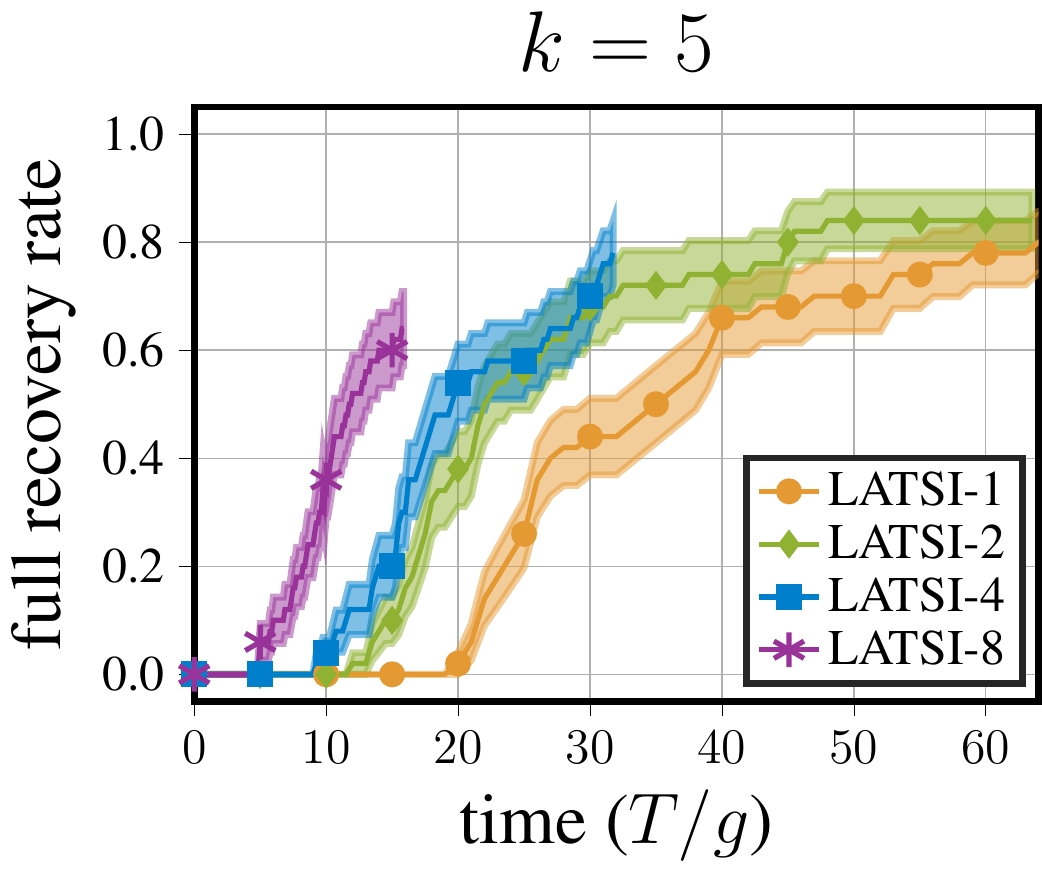}
	\end{subfigure}
	\caption{Full recovery rate of LATSI with 1, 2, 4 and 8 agents for $n = 128$, $k=1, 5$}
	\label{fig:LATSI_n128}
\end{figure}

In this section we provide additional results for a 1-dimensional search space ($d = 1$), where we estimate a $k$-sparse signal $\beta$ of length $n = 128$ with two sparsity rates of $k = 1, 5$.  Here, we use the same set of parameters as those outlined in \fref{sec:results}. \fref{fig:TS_n128}, \ref{fig:RSI_n128} and \ref{fig:LATSI_n128}  show the performance of SPATS, RSI and LATSI respectively. Our observations overall are similar to those in \fref{sec:multi-agent}. Concretely, we observe that both SPATS and LATSI show an improvement in performance with time as more agents become available. The efficiency of this asynchronous search by SPATS and LATSI becomes more pronounced with higher sparsity rate $k$ where there are more targets in the search space. On the other hand, multi agent RSI's performance worsens as we increase the number of agents (especially with larger sparsity rate $k$). This observation is similar to that in \fref{sec:multi-agent} and is explained by the lack of randomness in the information theoretic reward function of RSI. 


\subsection*{2-dimensional Search Space }
\begin{figure}
	\centering
	\begin{subfigure}{0.245\linewidth}
		\includegraphics[width=\linewidth]{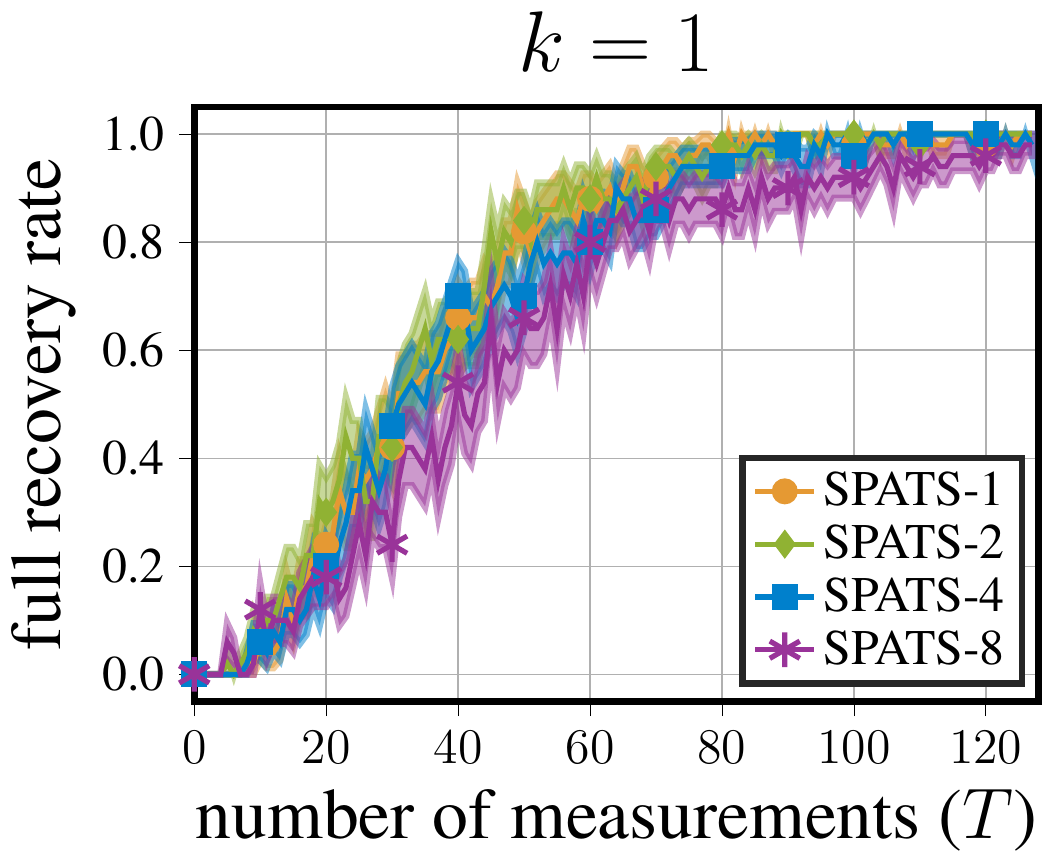}
	\end{subfigure}
	\begin{subfigure}{0.245\linewidth}
		\includegraphics[width=\linewidth]{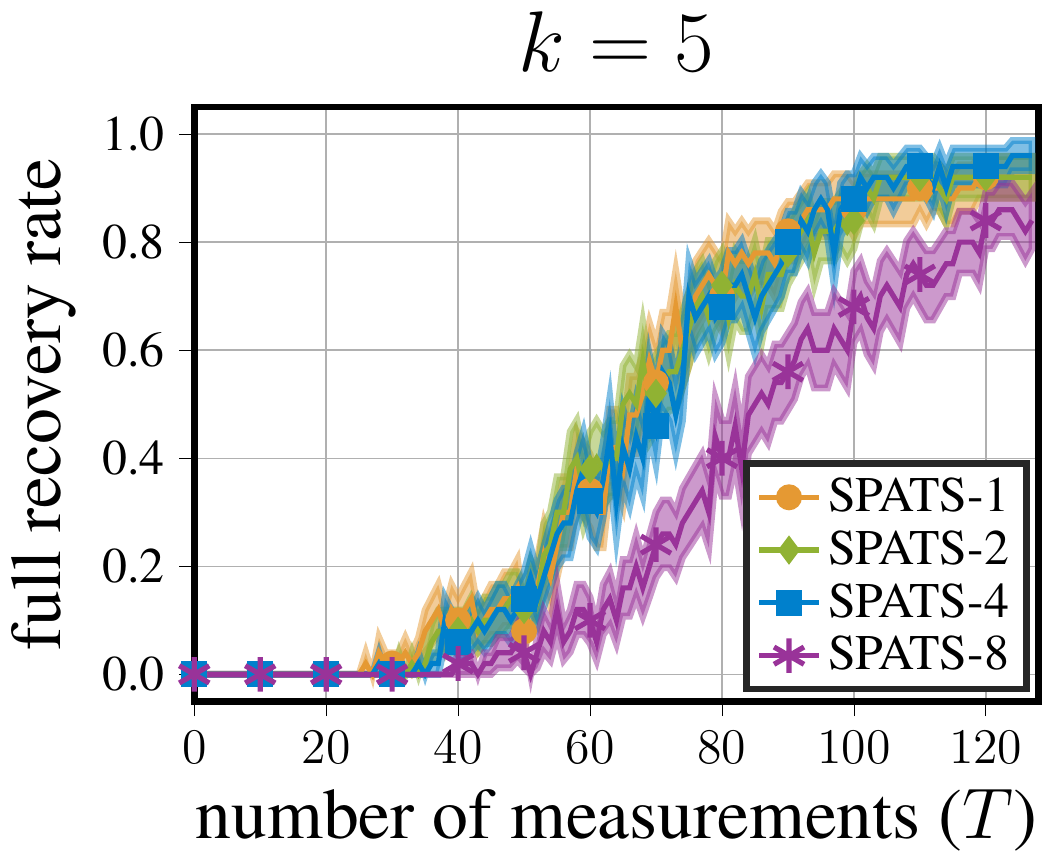}
	\end{subfigure}
	\begin{subfigure}{0.245\linewidth}
		\includegraphics[width=\linewidth]{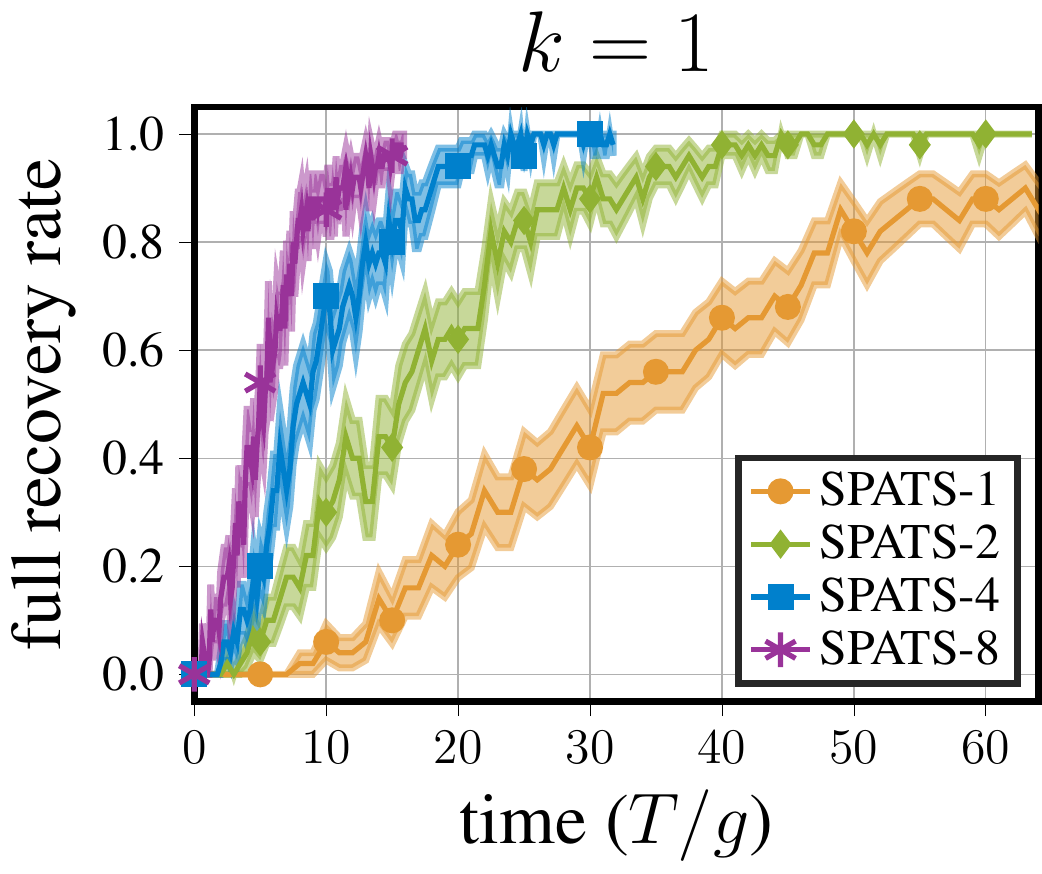}
	\end{subfigure}
	\begin{subfigure}{0.245\linewidth}
		\includegraphics[width=\linewidth]{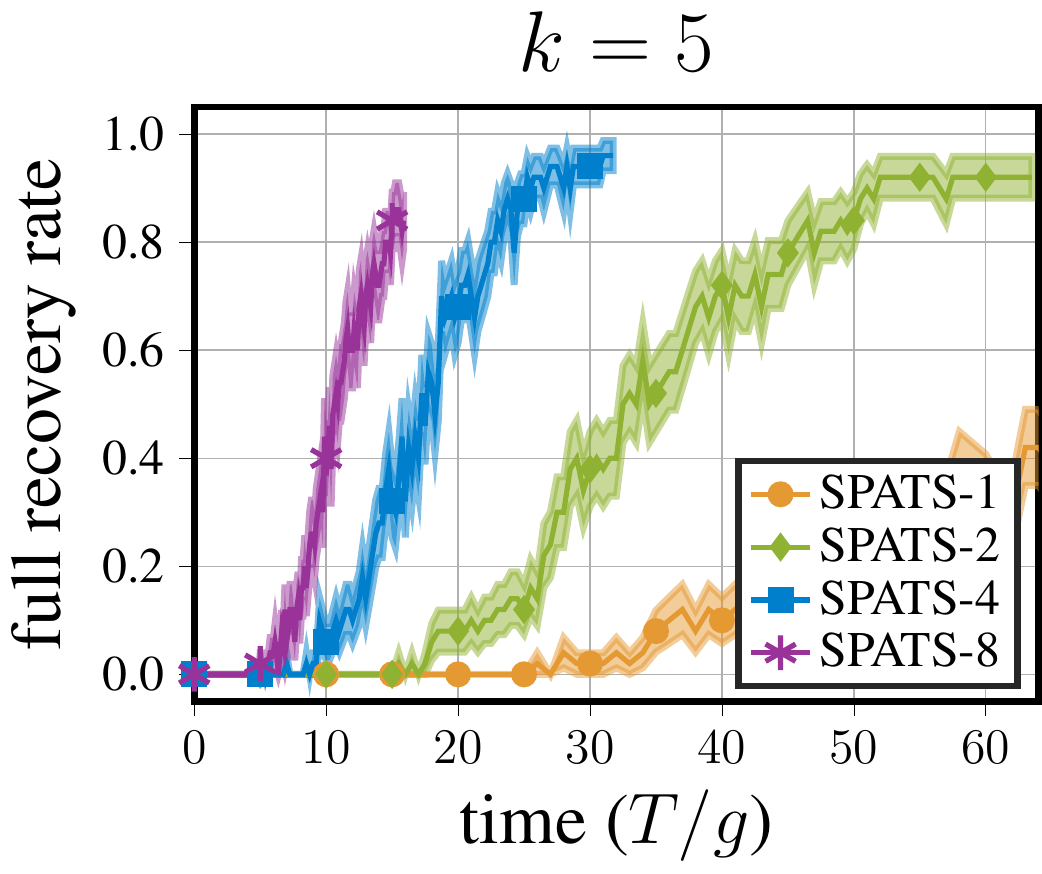}
	\end{subfigure}
	\caption{Full recovery rate of SPATS with 1, 2, 4 and 8 agents for $n = 16\times16$, $k=1, 5$}
	\label{fig:TS_n256}
\end{figure}

\begin{figure}
	\centering
	\begin{subfigure}{0.245\linewidth}
		\includegraphics[width=\linewidth]{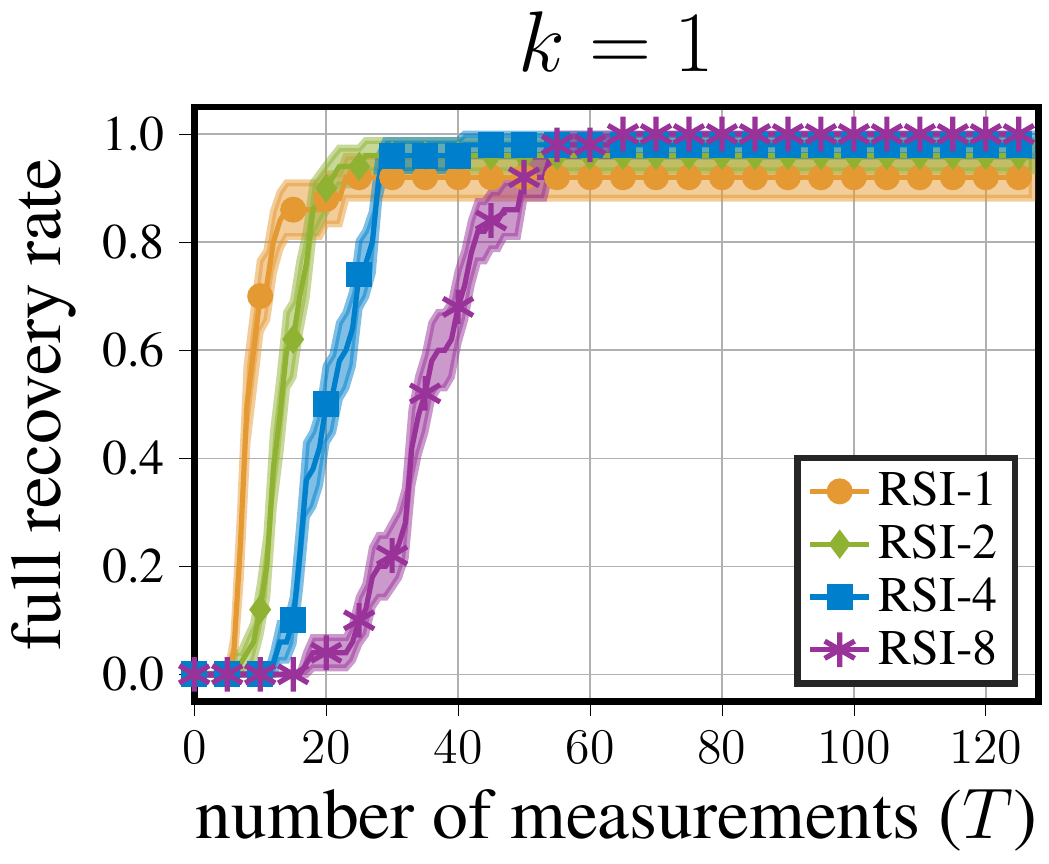}
	\end{subfigure}
	\begin{subfigure}{0.245\linewidth}
		\includegraphics[width=\linewidth]{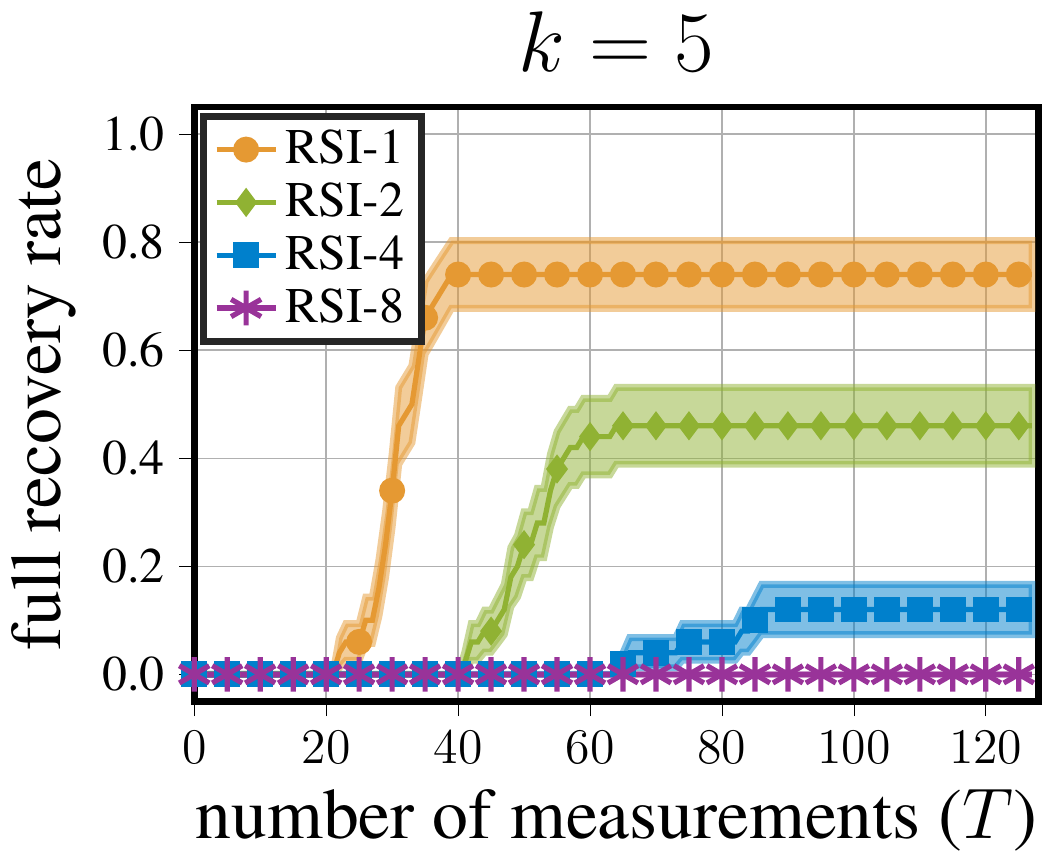}
	\end{subfigure}
	\begin{subfigure}{0.245\linewidth}
		\includegraphics[width=\linewidth]{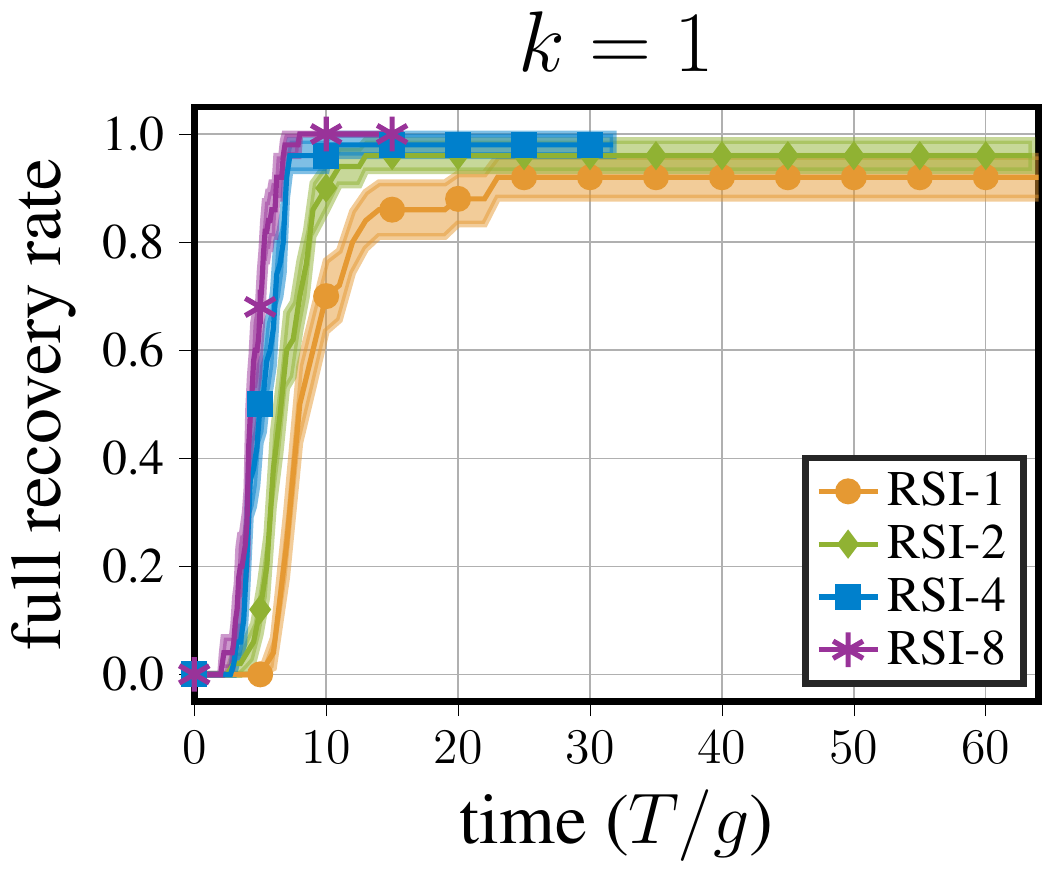}
	\end{subfigure}
	\begin{subfigure}{0.245\linewidth}
		\includegraphics[width=\linewidth]{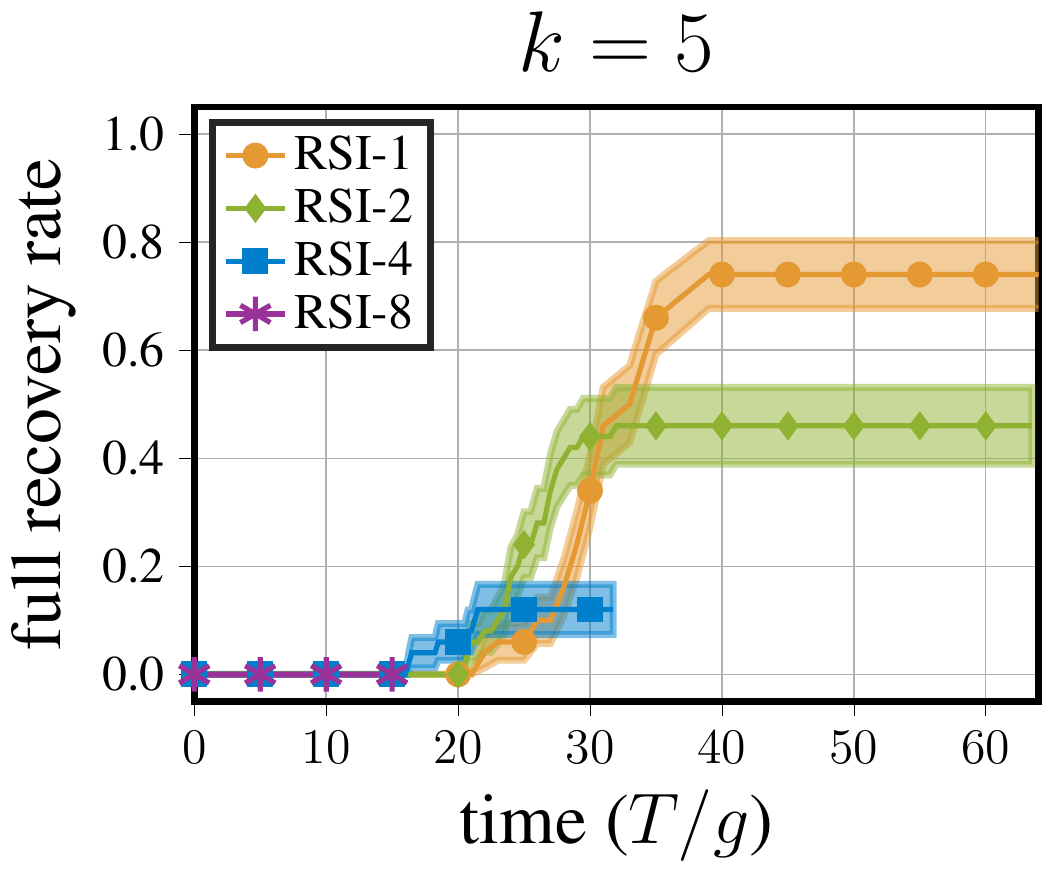}
	\end{subfigure}
	\caption{Full recovery rate of RSI with 1, 2, 4 and 8 agents for $n = 16\times16$, $k=1, 5$}
	\label{fig:RSI_n256}
\end{figure}

\begin{figure}
	\centering
	\begin{subfigure}{0.245\linewidth}
		\includegraphics[width=\linewidth]{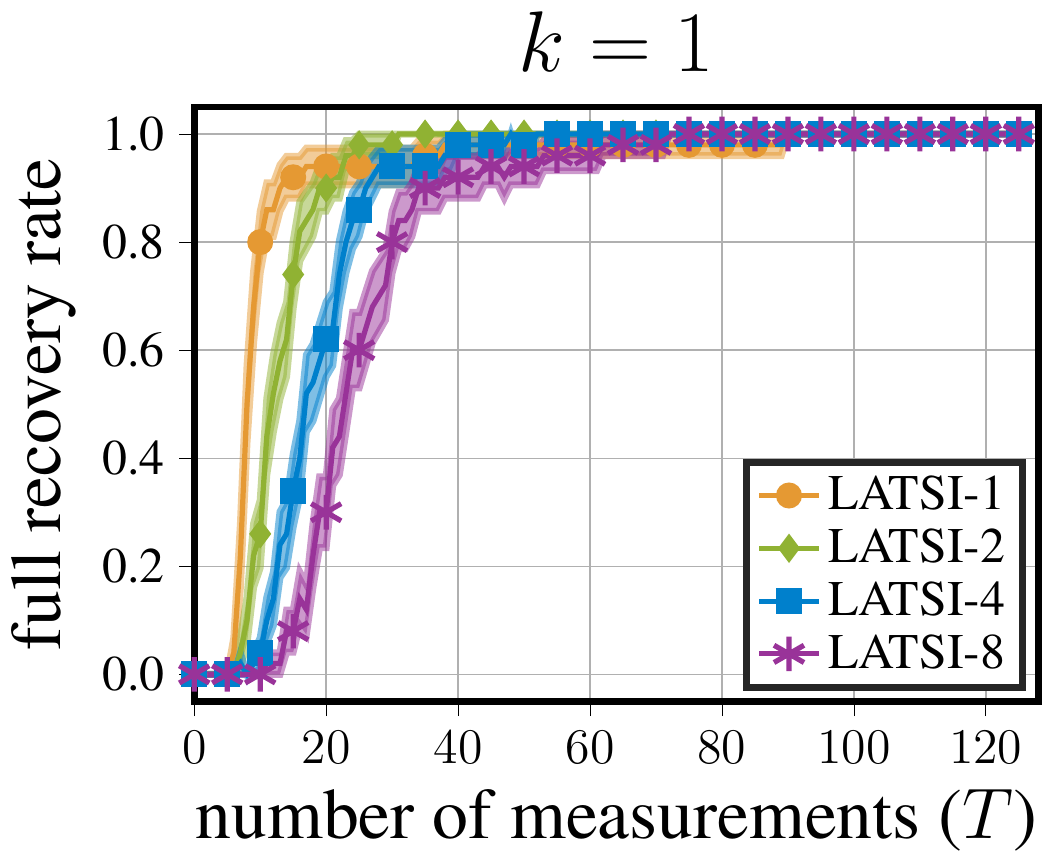}
	\end{subfigure}
	\begin{subfigure}{0.245\linewidth}
		\includegraphics[width=\linewidth]{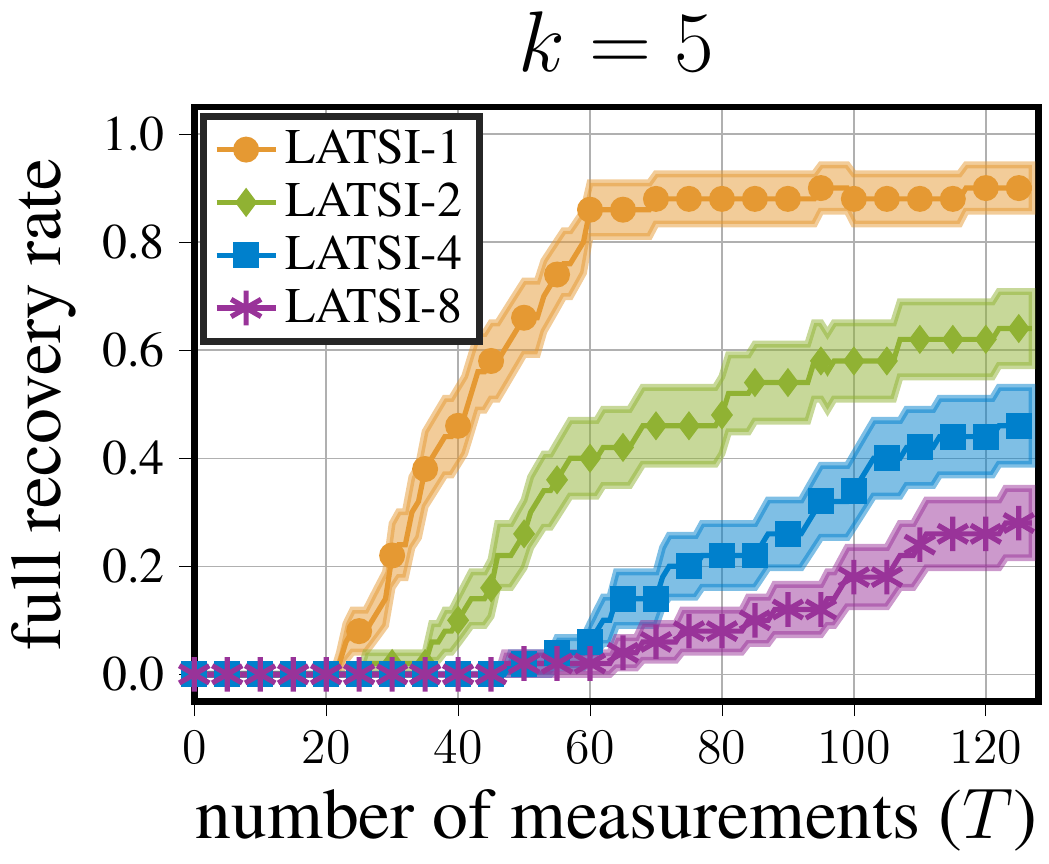}
	\end{subfigure}
	\begin{subfigure}{0.245\linewidth}
		\includegraphics[width=\linewidth]{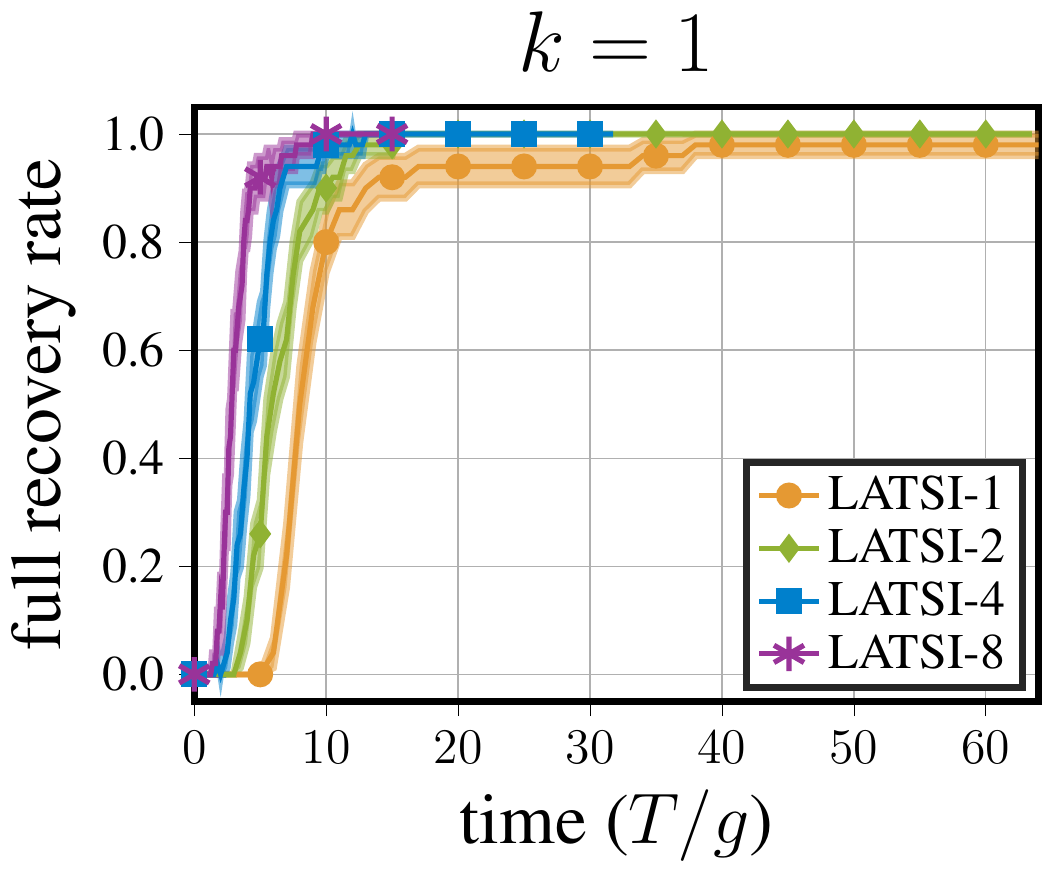}
	\end{subfigure}
	\begin{subfigure}{0.245\linewidth}
		\includegraphics[width=\linewidth]{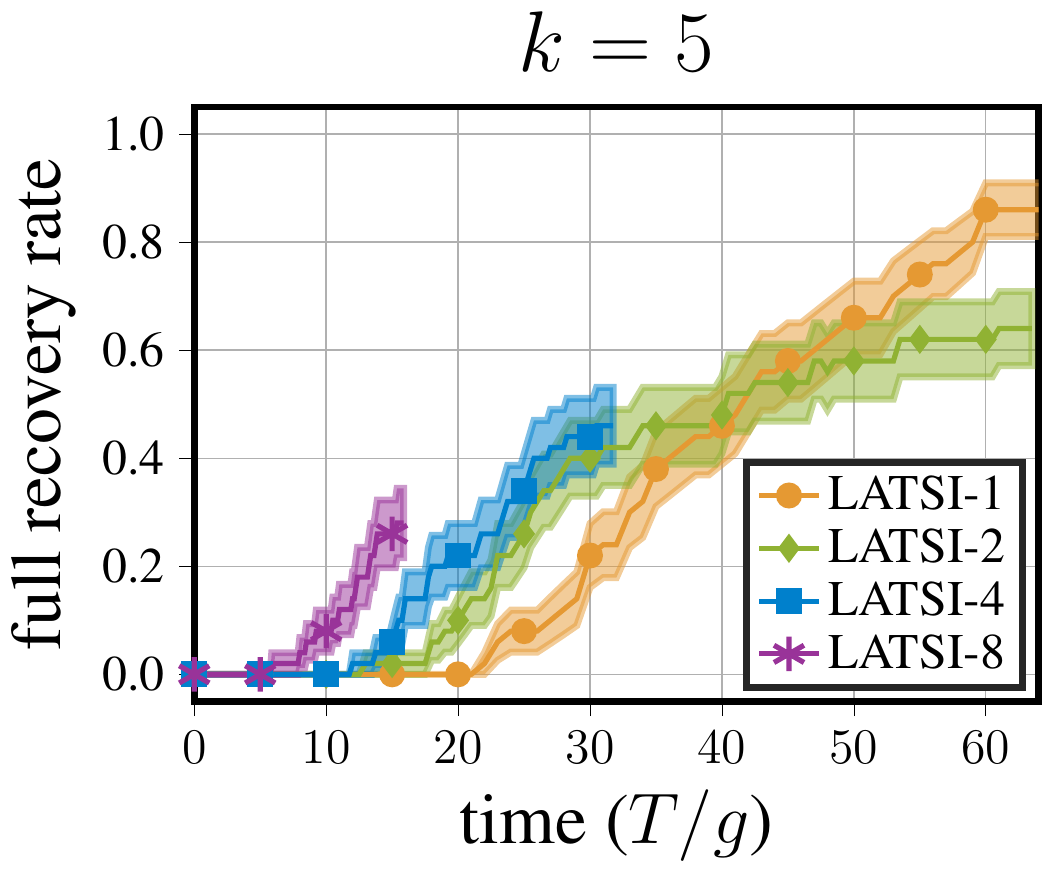}
	\end{subfigure}
	\caption{Full recovery rate of LATSI with 1, 2, 4 and 8 agents for $n = 16\times16$, $k=1, 5$}
	\label{fig:LATSI_n256}
\end{figure}

In this section we provide additional results for a 2-dimensional search space ($d = 2$), where we estimate a $k$-sparse signal $\beta$ of length $n = 16\times16$ and two sparsity rates of $k = 1, 5$.  The same set of parameters outlined in \fref{sec:results} are followed. \fref{fig:TS_n256}, \ref{fig:RSI_n256} and \ref{fig:LATSI_n256} show the performance of SPATS, RSI and LATSI respectively, in the multi agent setting. Our observations overall are similar to those in \fref{sec:multi-agent}. The probabilistic decision-making nature of SPATS ensures that its performance multiplies by its number of agents.
RSI relies on information theoretic decision making (no randomness). Hence, this lack of randomness in its reward function together with the poor approximation of mutual information for $k>1$, results in a poor performance generally worsening with more agents.
The multi agent performance of LATSI shows similar trends as SPATS given its partly probabilistic reward function. However, LATSI is not as efficient as SPATS given its poor approximation of mutual information on the info-greedy share of the reward function. These results further support the analysis in \fref{sec:results}.

\subsection*{1-dimensional vs 2-dimensional Search Space}
\begin{figure}
	\centering
	\begin{subfigure}{0.5\linewidth}
		\begin{subfigure}{0.5\linewidth}
			\includegraphics[width=\linewidth]{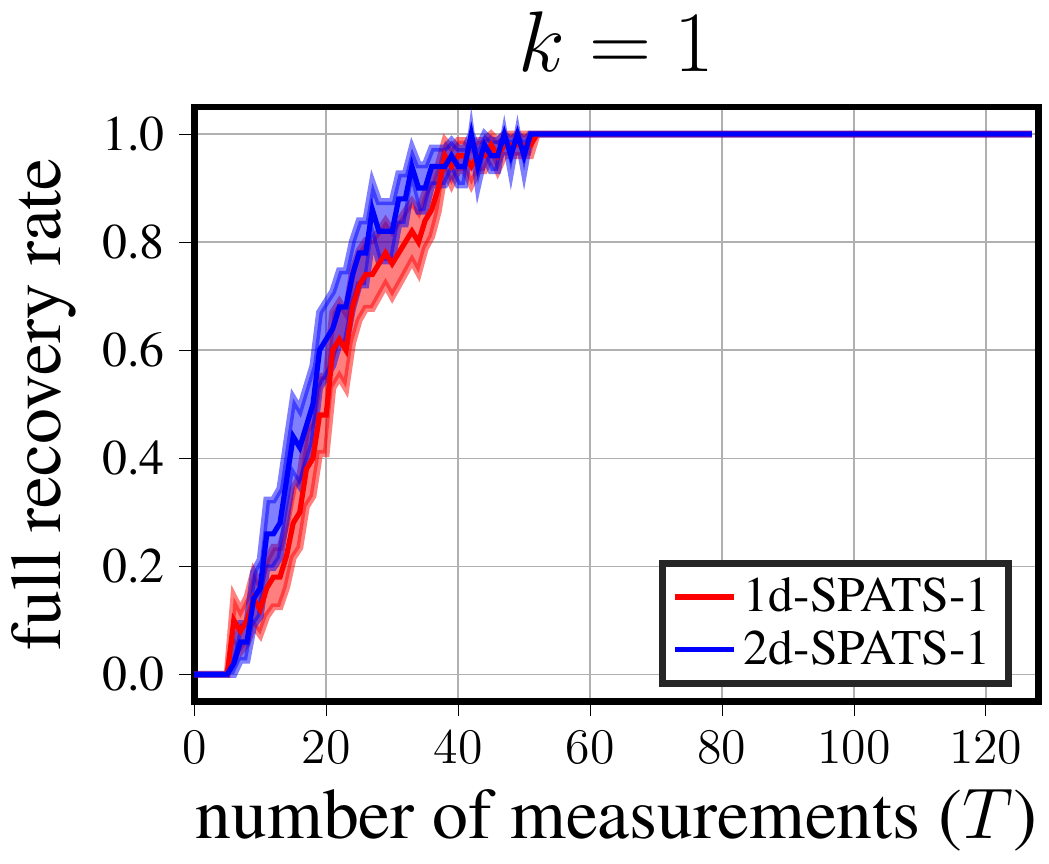}
		\end{subfigure}%
		\begin{subfigure}{0.5\linewidth}
			\includegraphics[width=\linewidth]{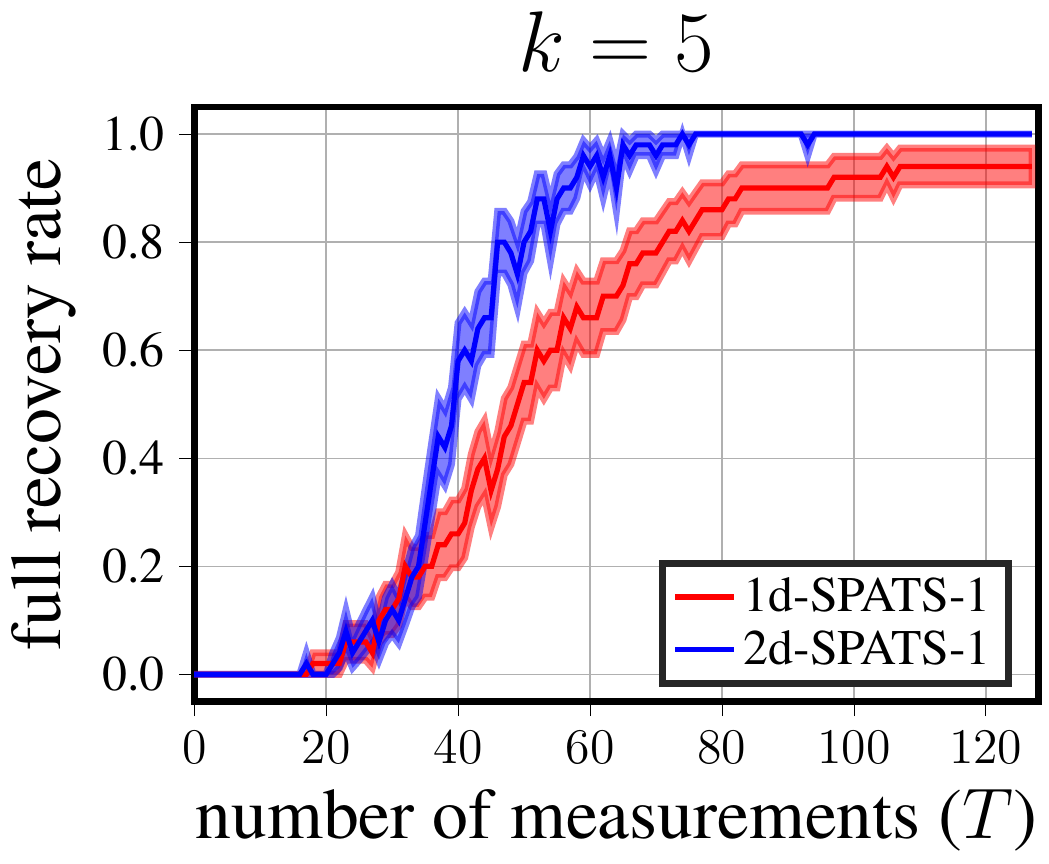}
		\end{subfigure}
		\caption{SPATS}
	\end{subfigure}%
	\begin{subfigure}{0.5\linewidth}
		\begin{subfigure}{0.5\linewidth}
			\includegraphics[width=\linewidth]{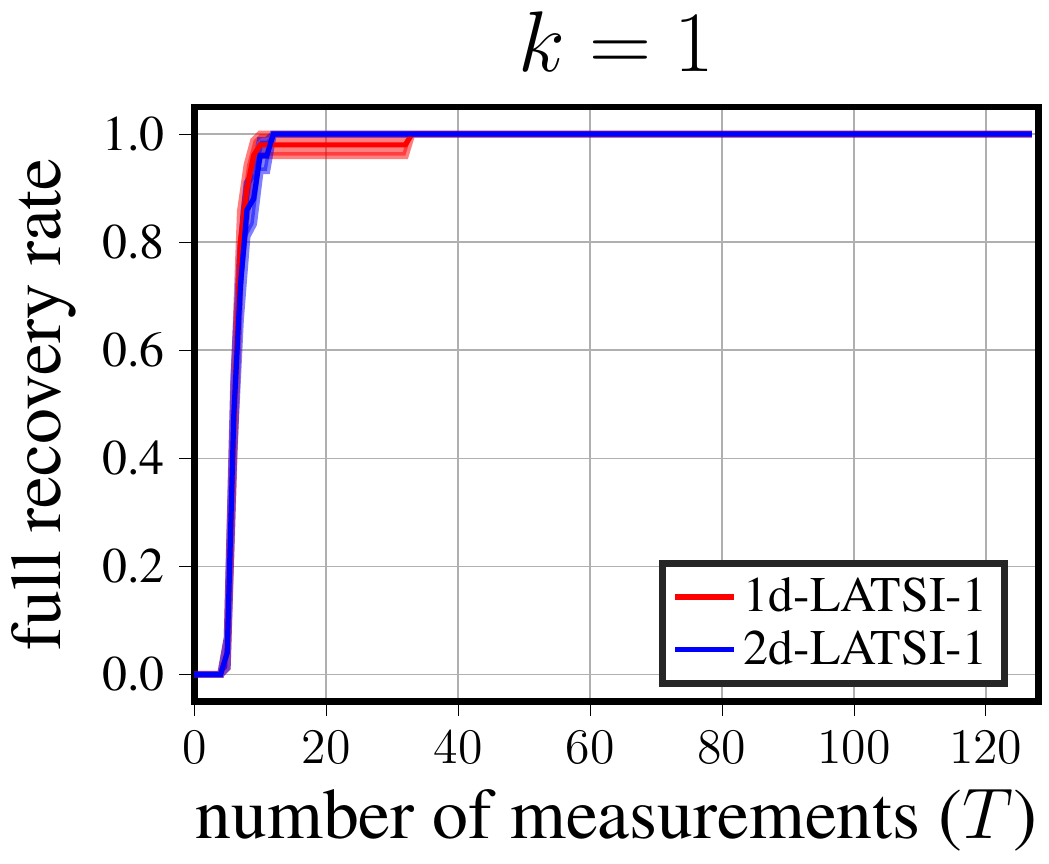}
		\end{subfigure}%
		\begin{subfigure}{0.5\linewidth}
			\includegraphics[width=\linewidth]{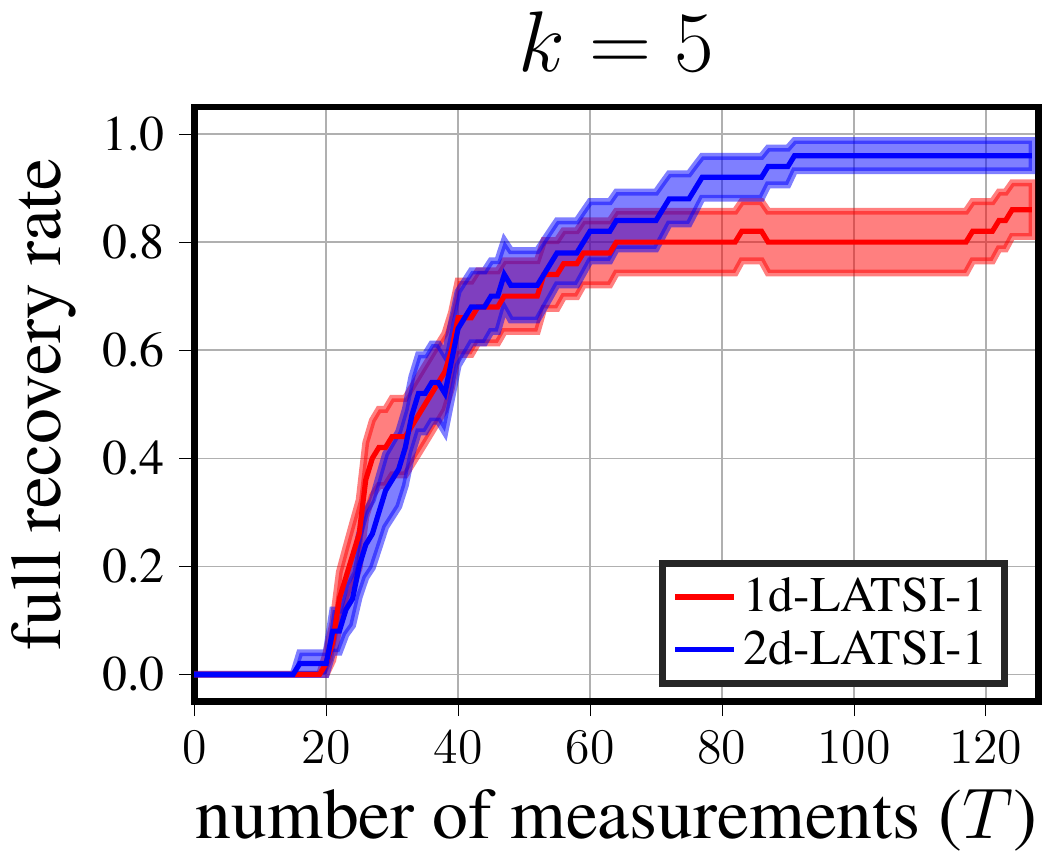}
		\end{subfigure}
		\caption{LATSI}
	\end{subfigure}
	\caption{Full recovery rate in 1-dimensional ($n=128$) vs 2-dimensional ($n = 8\times16$) search space with 1 agent for sparsity rates $k = 1, 5$}
	\label{fig:1Dvs2D}
\end{figure}

In \fref{fig:1Dvs2D}, we compare the performance of SPATS and LATSI for 1-dimensional (1d) and 2-dimensional (2d) search spaces with the same length $n=128$. We compare a 1d search space of length $n = 128$ with a 2d search space of length $n = 8 \times 16$ (i.e. flattened length $n = 128$) for two sparsity rates $k = 1, 5$. The same set of parameters outlined in \fref{sec:results} are followed. As evident in this figure, while 1-dimensional SPATS and LATSI have a similar performance for $k=1$, with larger $k$ they perform better in a 2d space rather than in 1d. The reason for this behavior is our region sensing assumption. Specifically, region sensing in the 2-dimensional grid expands the available action space and allows for a larger feasible action set compared to the 1-dimensional grid. Mainly, the larger action set helps our algorithms to choose better sensing actions in the 2-dimensional space, particularly when multiple targets are present.

\begin{figure}
	\centering
	\begin{subfigure}{0.49\linewidth}
		\centering
		\includegraphics[width=0.75\linewidth]{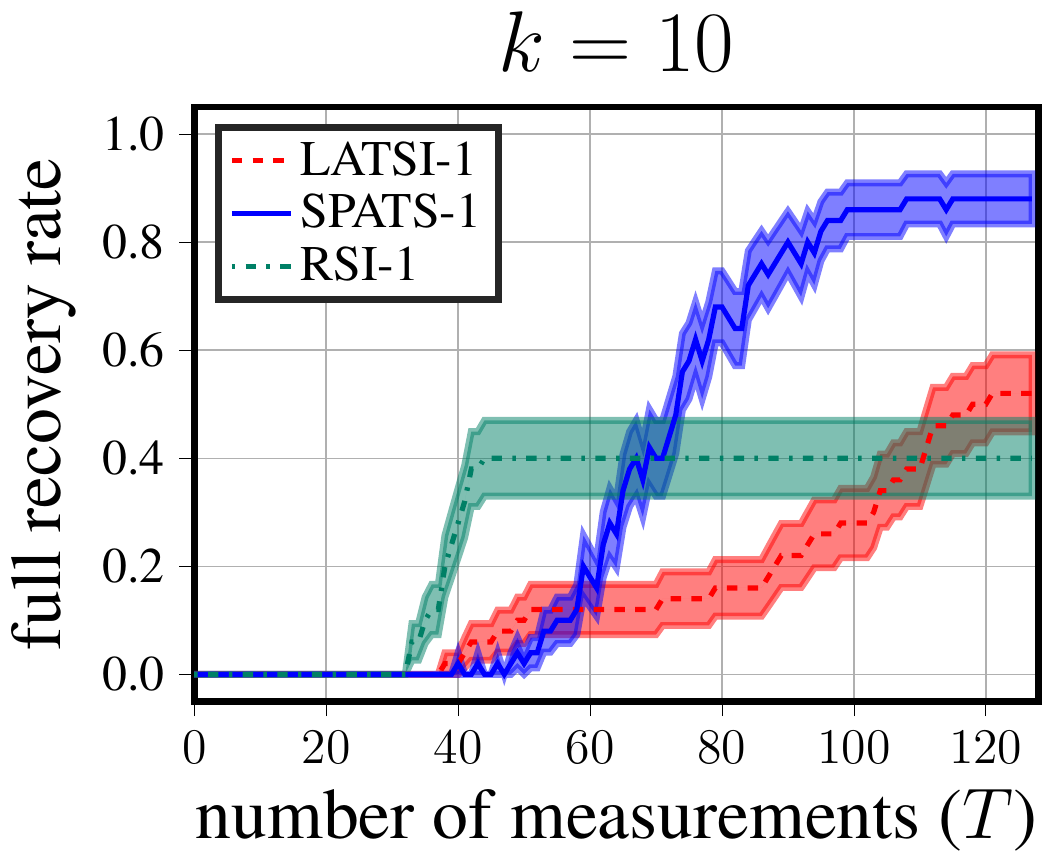}
		\caption{single agent}
		\label{fig:k10_1_agents}
	\end{subfigure}
	\begin{subfigure}{0.49\linewidth}
		\centering
		\includegraphics[width=0.75\linewidth]{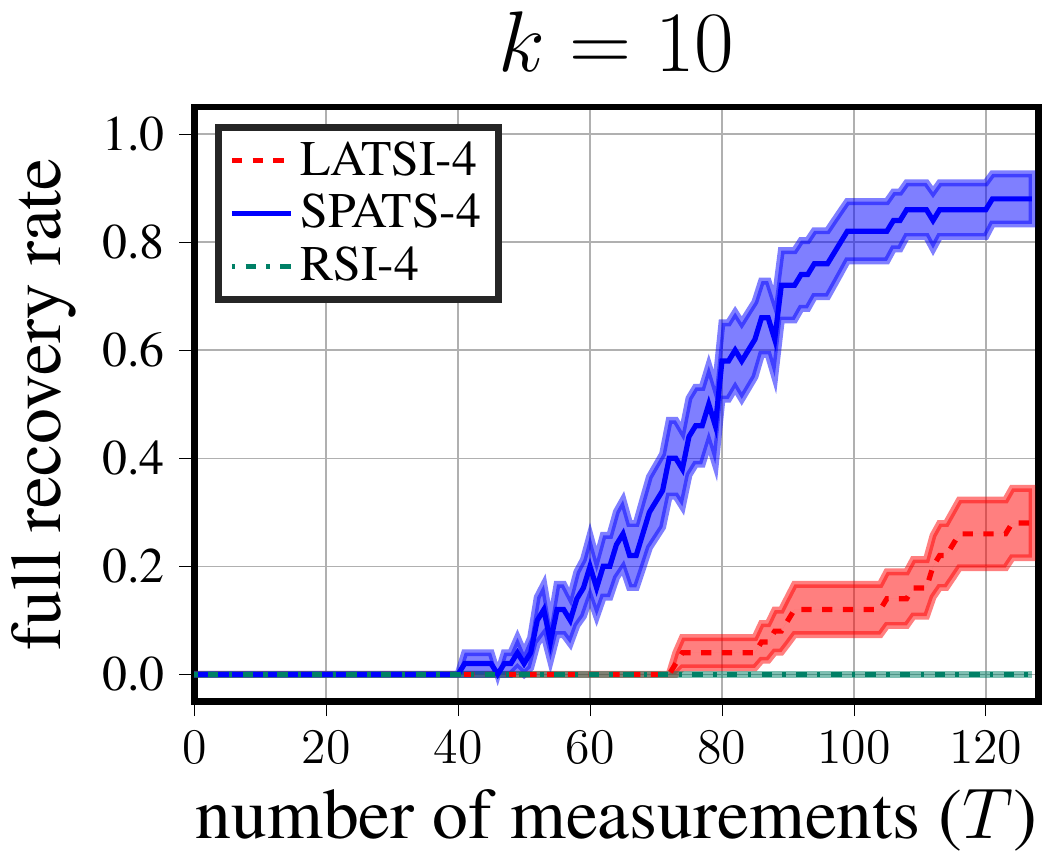}
		\caption{four agents}
		\label{fig:k10_4_agents}
	\end{subfigure}
	\caption{Full recovery rate of SPATS, LATSI and RSI for 1 and 4 agents for sparsity rate $k=10$}
	\label{fig:k10_1and4agents}
\end{figure}

\begin{figure}
	\centering
	\includegraphics[scale=0.5]{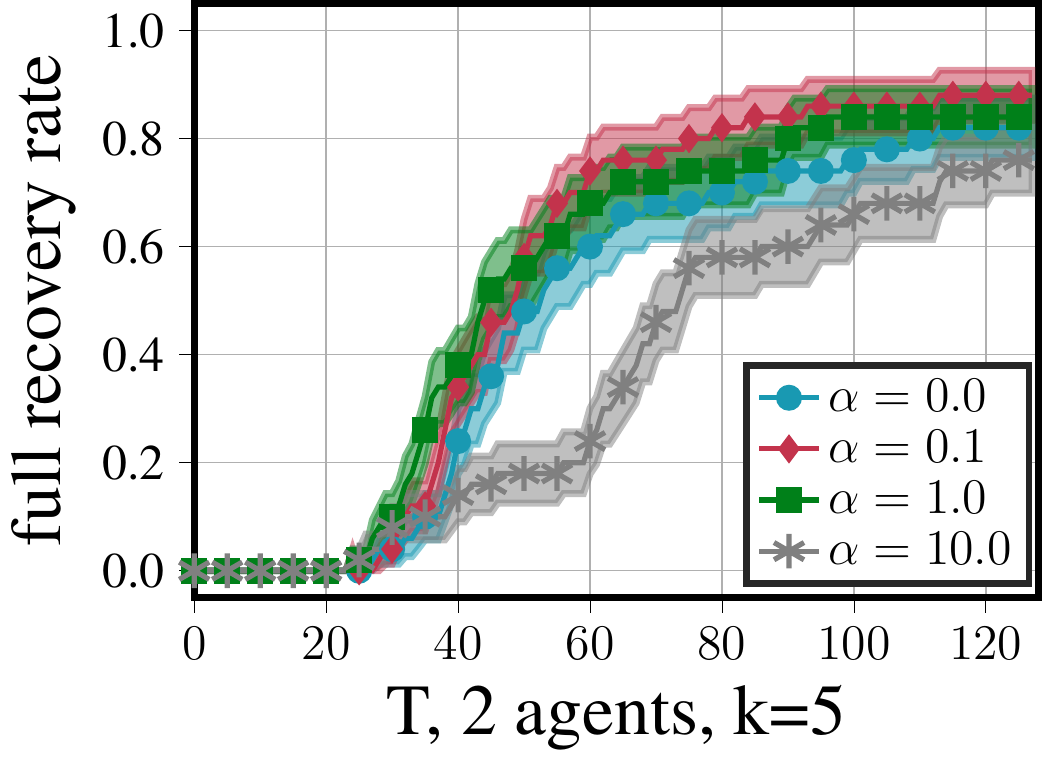}
	\caption{Effect of $\alpha$ in full recovery performance of LATSI with multiple agents and multiple targets}
	\label{fig:LTR_vary_alpha}
\end{figure}
\subsection*{Sparsity Rate}
In this section, we provide an additional set of results for sparsity rate of $k=10$ in  a 1-dimensional search space of length $n = 128$. \fref{fig:k10_1and4agents} illustrates the full recovery rate of SPATS, LATSI and RSI for one and four agents, respectively. Overall, we see that increasing the number of targets from $k=5$ to $k=10$ reduces the performance of all algorithms. Both figures confirm our previous observations that for sparsity rates of $k>1$, SPATS algorithm outperforms LATSI and RSI. In the case of single agent, the poor performance of LATSI and RSI can be traced back to poor approximation of mutual information for $k>1$ (see \fref{sec:single-agent}). In the four-agent scenario, RSI and LATSI are significantly worse than SPATS due to lack of randomness in information-greedy approaches (see \fref{sec:multi-agent} for details).

\subsection*{Sensitivity Analysis for LATSI}\label{app:sensitivity}

Recall tuning parameter $\alpha$ as the scaling factor in \fref{eq:LATSIreward} which is used to combine Laplace-TS and RSI's reward functions in LATSI's reward function $\bm{R}^+$. As evident by \fref{eq:LATSIreward}, increasing the value of $\alpha$ in $\bm{R}^+$ increases the weight of the reward computed by Laplace-TS.
We experiment with different values of $\alpha \in \{0, 0.1, 1, 10\}$ to determine it's effect on the full recovery rate performance. Since it is our aim to exploit parallelization, we perform the simulations with 2 agents and $k = 5$ in  a 1-dimensional search space of length $n = 128$. Plotting the results in \fref{fig:LTR_vary_alpha}, we observe that the algorithm's performance is robust for a wide range of $\alpha$. Based on this observation, we chose $\alpha = 1$ for all experiments with LATSI.

\end{document}